\documentclass{article}

\usepackage[backref=page]{hyperref}       
\renewcommand*{\backref}[1]{}
\hypersetup{
    colorlinks=true,
    linkcolor=blue,
    filecolor=magenta,      
    urlcolor=cyan,
    citecolor=cyan,
}

\usepackage{url}            
\usepackage{booktabs}       
\usepackage{amsfonts}       
\usepackage{nicefrac}       
\usepackage{microtype}      

\usepackage{float}
\usepackage{mathtools}
\usepackage{amsthm}
\usepackage{amssymb}
\usepackage{amsfonts}
\usepackage{graphicx}
\usepackage{booktabs}
\usepackage{siunitx}
\usepackage{arydshln}
\usepackage{caption}
\usepackage{subcaption}
\usepackage{soul}
\usepackage[dvipsnames]{xcolor}
\usepackage{epstopdf}
\usepackage{fancyvrb}
\usepackage{cancel}
\usepackage{makecell}

\newtheorem{statement}{Statement}[section]

\newcommand{\tr}{\text{tr}}
\newcommand{\Var}{\text{Var}}

\usepackage{pifont}
\def\num#1{%
        \raisebox{.9pt}{\textcircled{\raisebox{-.9pt}{#1}}}%
}

\newcommand{\pd}[2]{\frac{\partial #1}{\partial #2}}

\newcommand{\pp}[1]{\left( #1 \right)}
\newcommand{\mb}{\mathbf}

\newcommand{\norm}[1]{\left|\left| #1 \right|\right|}
\newcommand{\expect}[2]{\mathbb{E}_{#1}\left[ #2 \right]}

\newcommand{\ovec}[1]{\operatorname{vec}\pp{#1}}

\makeatletter
\newcommand{\longdash}[1][2em]{%
  \makebox[#1]{$\m@th\smash-\mkern-7mu\cleaders\hbox{$\mkern-2mu\smash-\mkern-2mu$}\hfill\mkern-7mu\smash-$}}
\makeatother
\newcommand{\omitskip}{\kern-\arraycolsep}

\usepackage{algorithmicx}
\usepackage{forloop}
\usepackage{algpseudocode}

\usepackage{overpic}

\usepackage{amssymb}
\usepackage{pifont}
\newcommand{\cmark}{\ding{51}}%
\newcommand{\xmark}{\ding{55}}%

\definecolor{lightgraysquare}{rgb}{0.7,0.7,0.7}
\newcommand*{\greysquare}{\textcolor{lightgraysquare}{\blacksquare}}

\usepackage{listings}

\definecolor{dkgreen}{rgb}{0,0.6,0}
\definecolor{gray}{rgb}{0.5,0.5,0.5}
\definecolor{mauve}{rgb}{0.58,0,0.82}
\definecolor{lightgray}{HTML}{EEEEEE}

\usepackage{wrapfig}
\usepackage{tabularx}
\usepackage{arydshln}
\usepackage{array}
\usepackage{multirow}
\usepackage[accepted]{icml2021}

\newcommand{\boldtheta}{{\boldsymbol{\theta}}}
\newcommand{\boldepsilon}{\boldsymbol{\epsilon}}
\newcommand{\boldxi}{\boldsymbol{\xi}}

\newcommand{\bolds}{\boldsymbol{s}}
\newcommand{\boldx}{\boldsymbol{x}}

\newcommand{\boldh}{\boldsymbol{h}}
\newcommand{\boldg}{\boldsymbol{g}}
\newcommand{\boldp}{\boldsymbol{p}}

\newcommand{\boldv}{\boldsymbol{v}}

\newcommand{\boldzero}{\boldsymbol{0}}
\newcommand{\boldone}{\boldsymbol{1}}

\newcommand{\ges}{\hat{\boldg}^{\text{ES}}}
\newcommand{\gpes}{\hat{\boldg}^\text{PES}}

\newcommand{\gesanti}{\hat{\boldg}^{\text{ES-A}}}

\makeatletter
\def\gpesanti{%
  \@ifnextchar^%
    {\@gpesanti}
    {\@gpesanti^{}}%
}
\def\@gpesanti^#1{%
  \hat{\boldg}^{\text{PES-A} #1}%
}
\makeatother

\newcommand{\veps}{\operatorname{vec}(\boldepsilon)}
\newcommand{\vTheta}{\operatorname{vec}(\Theta)}

\icmltitlerunning{Unbiased Gradient Estimation in Unrolled Computation Graphs with Persistent Evolution Strategies}

\begin{document}

\twocolumn[
\icmltitle{Unbiased Gradient Estimation in Unrolled Computation Graphs \\with Persistent Evolution Strategies}

\vspace{-0.3cm}
\begin{icmlauthorlist}
\icmlauthor{Paul Vicol}{toronto}
\icmlauthor{Luke Metz}{google}
\icmlauthor{Jascha Sohl-Dickstein}{google}
\end{icmlauthorlist}

\icmlaffiliation{toronto}{University of Toronto; work done while on internship at Google.}
\icmlaffiliation{google}{Google Brain}

\icmlcorrespondingauthor{Paul Vicol}{pvicol@cs.toronto.edu}

\icmlkeywords{evolution strategies, unbiased gradient estimation, unrolled computation graph, real-time recurrent learning, hyperparameter optimization, learned optimizers}

\vskip 0.5cm
]

\printAffiliationsAndNotice{}  

\begin{abstract}
Unrolled computation graphs arise in many scenarios, including training RNNs, tuning hyperparameters through unrolled optimization, and training learned optimizers. Current approaches to optimizing parameters in such computation graphs suffer from high variance gradients, bias, slow updates, or large memory usage. We introduce a method called Persistent Evolution Strategies (PES), which divides the computation graph into a series of truncated unrolls, and performs an evolution strategies-based update step after each unroll. PES eliminates bias from these truncations by accumulating correction terms over the entire sequence of unrolls. PES allows for rapid parameter updates, has low memory usage, is unbiased, and has reasonable variance characteristics. We experimentally demonstrate the advantages of PES compared to several other methods for gradient estimation on synthetic tasks, and show its applicability to training learned optimizers and tuning hyperparameters.
\end{abstract}

\vspace{-0.7cm}
\section{Introduction}
\vspace{-0.1cm}
Unrolled computation graphs arise in many scenarios in machine learning, including when training RNNs~\citep{williams1990efficient}, tuning hyperparameters through unrolled computation graphs~\citep{baydin2017online,domke2012generic,maclaurin2015gradient,wu2018understanding,franceschi2017forward,donini2019scheduling,franceschi2018bilevel,liu2018darts,shaban2019truncated}, and training learned optimizers~\citep{li2016learning,li2017learning,andrychowicz2016learning,wichrowska2017learned,metz2018meta, metz2019understanding,metz2020using,metz2020tasks}.
Many methods exist for computing gradients in such computation graphs, including ones based on reverse-mode~\citep{williams1990efficient,tallec2017unbiasing,aicher2019adaptively,grefenstette2019generalized} and forward-mode~\citep{williams1989learning,tallec2017unbiased,mujika2018approximating,benzing2019optimal,marschall2019unified,menick2020practical} gradient accumulation.
These methods have different tradeoffs with respect to compute, memory, and gradient variance.

Backpropagation through time involves backpropagating through a full unrolled sequence (e.g. of length $T$) for each parameter update.
Unrolling a model over full sequences faces several difficulties: 
1) the memory cost scales linearly with the unroll length, because we need to store intermediate activations for backprop (though this can be reduced at the cost of additional compute \citep{dauvergne2006data, chen2016training}); 
2) we only perform a single parameter update after each full unroll, which is computationally expensive and introduces large latency between parameter updates; 
3) long unrolls can lead to exploding or vanishing gradients \citep{pascanu2013difficulty}, 
and chaotic and poorly conditioned loss landscapes~\citep{pearlmutter1996investigation, maclaurin2015gradient, parmas2018pipps, metz2019understanding}.
This is especially true in meta-learning \citep{metz2019understanding}.

The most commonly-used technique to alleviate these issues is truncated backprop through time (TBPTT)~\citep{werbos1990backpropagation,tallec2017unbiasing}, which splits the full sequence into shorter sub-sequences and performs a backprop update after processing each sub-sequence.
However, a critical drawback of TBPTT is that it yields biased gradients, that can severely impact training (e.g. only taking into account short-term dependencies). 
To address the poorly conditioned loss surfaces that often result from sequential computation, it can additionally be useful to minimize a smoothed version of the loss.
Evolution strategies (ES) is a family of algorithms that estimate gradients using stochastic finite-differences, 
and which provide an unbiased estimate of the gradient of the objective smoothed with a Gaussian.
ES works well on pathological meta-optimization loss surfaces \citep{metz2019understanding}; however, due to the computational expense of running full unrolls, ES can only practically be applied in a truncated fashion, introducing bias.

An alternative to BPTT is real-time recurrent learning (RTRL), which performs forward gradient accumulation~\citep{williams1989learning}.
RTRL enables online parameter updates (after each partial unroll) and does not suffer from truncation bias; however, its memory and compute requirements render it intractable for large-scale problems.
Many approximations to RTRL have been proposed~\citep{tallec2017unbiased,mujika2018approximating,benzing2019optimal}, but most have high variance, are complicated to implement, or are only applicable to a restricted class of models.

We introduce an approach to unbiased gradient estimation using short, truncated unrolls, called Persistent Evolution Strategies (PES).
In PES, we accumulate the perturbations experienced by the \textit{outer parameters} in each partial unroll---rather than starting perturbations from scratch as in vanilla ES---which yields an unbiased estimate of the gradient even when using truncated sequences.
PES is simple to implement, and because it is an evolution strategies-based approach, it retains desirable characteristics such as being trivially parallelizable, memory efficient, and broadly applicable to many different types of problems, including to non-differentiable target functions.

\paragraph{Contributions}
\vspace{-0.2cm}
\begin{itemize}
    \vspace{-0.2cm}
    \item We introduce a method called Persistent Evolution Strategies (PES) to obtain unbiased gradient estimates for the parameters of an unrolled system from \textit{partial unrolls} of the system.
    \item We prove that PES is an unbiased gradient estimate for a smoothed version of the loss, and an unbiased estimate of the true gradient for quadratic losses.
    \item We provide theoretical and empirical analyses of its variance. 
    In addition, we describe a variance reduction technique for PES, that incorporates the analytic gradient (computed with standard backprop) of the most recent unroll of the dynamical system.
    \item We demonstrate the applicability of PES in several illustrative scenarios: 1) we apply PES to tune hyperparameters including learning rates and momentums, by estimating hypergradients through partial unrolls of optimization algorithms; 2) we use PES to meta-train a learned optimizer; 3) we use PES to learn policy parameters for a continuous control task.
\end{itemize}
\vspace{-0.1cm}
We provide a \href{https://colab.research.google.com/drive/1qD65uz2zq6UgqVayzaNGxMHPKD1y_XP1?usp=sharing}{Colab notebook} implementation of PES.

\vspace{-0.2cm}
\section{Background}
\label{sec:background}

We provide an overview of notation in Appendix~\ref{app:notation}.

\vspace{-0.1cm}
\paragraph{Problem Setup.}
\vspace{-0.1cm}
\begin{figure*}
    \centering
    \includegraphics[width=0.8\linewidth]{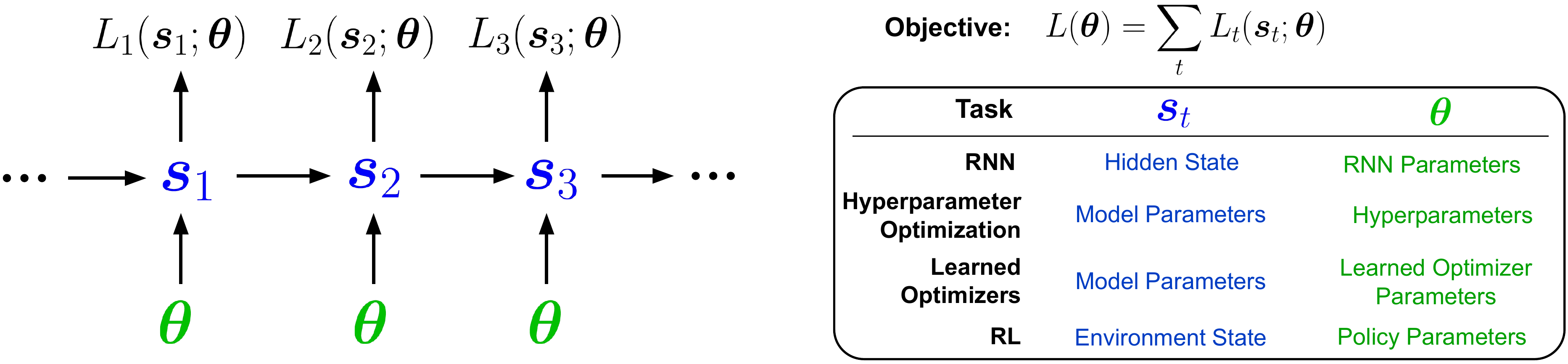}
    \caption{
    \textbf{An unrolled computation graph}, illustrating how both RNNs and unrolled optimization can be described using Equations \ref{eq s recurrence} and \ref{eq L}.
    In RNN training, $\bolds_t$ is the hidden state of the RNN, $L_t(\cdot)$ is the prediction cross entropy at each timestep, $\boldtheta$ refers to the RNN parameters, and $f$ corresponds to the forward pass of the RNN, that takes an input and the previous hidden state ($\bolds_t$) and returns a new hidden state ($\bolds_{t+1})$.   In unrolled optimization, $\bolds_t$ contains the parameters of the base model and optimizer accumulators (e.g. momentum), $L_t(\cdot)$ is a meta-objective such as validation performance, $\boldtheta$ contains hyperparameters (e.g. the learning rate, weight decay, etc.) that govern the optimization, and $f$ corresponds to the update step of an optimization algorithm such as SGD, RMSprop~\citep{Tieleman2012}, or Adam~\citep{kingma2014adam}.
    }
    \vspace{-0.3cm}
    \label{fig:unrolled-graph}
\end{figure*}
We consider unrolled computation graphs with state $\bolds_t$ updated based on parameters $\boldtheta$ via the recurrence:
\begin{equation}
    \bolds_t = f(\bolds_{t-1}, \boldx_t; \boldtheta)
    \label{eq s recurrence}
\end{equation}
where $\boldx_t$ is an optional input at step $t$.
The objective function for optimizing $\boldtheta$ is the sum of per-timestep losses $L_t(\bolds_t; \boldtheta)$:
\begin{equation}
    L(\boldtheta) = \sum_{t=1}^T L_t(\bolds_t ; \boldtheta)
    \label{eq L}
\end{equation}
This setup is general, even encompassing situations where we want to consider only the final loss at step $T$, which can be expressed using a telescoping sum of loss differences between successive steps~\citep{beatson2019efficient}.
\footnote{For details on telescoping sums, see Appendix~\ref{app:telescoping}.}
Instances of this problem setup include training RNNs, training learned optimizers, learning policies for control tasks, and unrolled optimization, as illustrated in Figure~\ref{fig:unrolled-graph}.
\vspace{-0.1cm}
\paragraph{Unrolled Optimization.}
Optimization algorithms can be unrolled to yield computation graphs, in which the nodes are the model parameters at successive optimization steps.
Estimating gradients through unrolled optimization has been used to tune hyperparameters~\citep{domke2012generic,maclaurin2015gradient, baydin2017online,donini2019scheduling,franceschi2017forward} and train learned optimizers~\citep{li2016learning,li2017learning,andrychowicz2016learning,wichrowska2017learned,metz2019understanding,metz2020using,metz2020tasks,metz2018meta}.
\vspace{-0.1cm}
\paragraph{Truncation Bias.}
Truncation, or short horizon, bias poses a major challenge when unrolled optimization is decomposed into a sequence of short sequential unrolls of length $K \ll T$.
These challenges have been demonstrated in gradient-based hyperparameter optimization~\citep{wu2018understanding} and in the training of learned optimizers~\citep{metz2019understanding}.
Approaches to mitigating short horizon bias are an area of active research~\citep{micaelli2020non}.
\vspace{-0.1cm}
\paragraph{Smoothing.}
Unrolling optimization for many steps can lead to pathological meta-loss surfaces that exhibit near-discontinuities and chaotic structure \citep{parmas2018pipps, metz2019understanding}.
Optimization on such non-smooth landscapes fails due to exploding gradients or gets stuck in poor local minima.
One effective method to address these pathologies is to smooth the meta-loss surface, e.g. descend the Gaussian-blurred objective $L(\boldtheta) = \mathbb{E}_{\tilde{\boldtheta} \sim \mathcal{N}(\boldtheta, \sigma^2 I)}[L(\tilde{\boldtheta})]$  \cite{staines2012variational,metz2019understanding}.
Conveniently, ES provides an unbiased estimate of the gradient of this smoothed objective.
However, the ES estimate remains biased when computed on truncations.
\vspace{-0.1cm}
\paragraph{Evolution Strategies.}
Evolution Strategies (ES)~\citep{rechenberg1973,nesterov2017random} refers to a family of methods for estimating a descent direction for arbitrary black-box functions using stochastic finite differences.
Since ES only requires function evaluations and not gradients, it is a zeroth-order optimization method.
The vanilla ES estimator is defined as:
\begin{equation} \label{eq:vanilla-es}
    \ges = \frac{1}{N \sigma^2} \sum_{i=1}^N \boldepsilon^{(i)} L(\boldtheta + \boldepsilon^{(i)})
\end{equation}
where $\boldepsilon^{(i)} \sim \mathcal{N}(0, \sigma^2 I)$.
ES is trivially parallelizable, and thus highly scalable---it has seen renewed interest in recent years as a viable optimization algorithm for reinforcement learning among other black-box problems~\citep{salimans2017evolution,mania2018simple,ha2018world,houthooft2018evolved,cui2018evolutionary, ha2020neuroevolution}.
The estimator in Eq.~\ref{eq:vanilla-es} has high variance, and thus many variance reduction techniques have been proposed, including control variates~\citep{tang2020variance} and antithetic sampling~\citep{mcbook}.
Antithetic sampling involves using pairs of function evaluations $\boldtheta + \boldepsilon$ and $\boldtheta - \boldepsilon$, yielding the following estimator:
\begin{equation*} \label{eq:es-antithetic}
    \gesanti = \frac{1}{N \sigma^2} \sum_{i=1}^{N/2} \boldepsilon^{(i)} (L(\boldtheta + \boldepsilon^{(i)}) - L(\boldtheta - \boldepsilon^{(i)}))
\end{equation*}
where $N$ is even, and $\boldepsilon^{(i)} \sim \mathcal{N}(0, \sigma^2 I)$.
Several methods have been proposed to improve the search space for ES, including covariance matrix adaptation ES (CMA-ES)~\citep{hansen2016cma} and Guided ES~\citep{maheswaranathan2018guided}.
A limitation of ES is that applying it to full unrolls is often computationally costly (as we only make one update to the system parameters every full unroll), while applying ES to partial unrolls suffers from truncation bias similarly to TBPTT.
In contrast, PES allows computation of gradients from partial updates without incurring truncation bias.

\paragraph{Hysteresis.}
Any approach that performs online parameter updates, including RTRL and its approximations, will suffer from hysteresis, which refers to the dependence of the state of a system on its history. 
This is due to the fact that if we update $\boldtheta$, then any accumulated state (e.g. in the case of RTRL, the accumulated Jacobian $\frac{d \bolds_t}{d \boldtheta}$) will be incorrect because it is computed from previous values of $\boldtheta$.
To eliminate hysteresis completely, one would need to run the full sequence for a given problem for each parameter update, which is often prohibitively expensive.
In practice, hysteresis can be mitigated by using sufficiently small learning rates; this introduces a tradeoff between training stability and training speed.

\vspace{-0.2cm}
\section{Related Work}
\label{sec:related-work}
\vspace{-0.2cm}

\begin{table*}[t]
\vspace{-0.3cm}
\centering
\label{table:comparison}
\caption{\textbf{Comparison of approaches for learning parameters in unrolled computation graphs}. $S$ is the size of the system state (e.g. the RNN hidden state dimension, or in the case of hyperparameter optimization the inner-problem's weight dimensionality and potentially the optimizer state; $P$ is the dimensionality of $\boldtheta$; $T$ is the total number of steps in a sequence/unroll; $K$ is the truncation length; and $N$ is the number of samples (also called \textit{particles}) used for the reparameterization gradient and in ES-based algorithms; $F$ and $B$ are the costs of a forward and backward pass, respectively; terms in {\color{purple} purple} denote computation/memory that can be split across parallel workers.
See Appendix~\ref{app:deriv-RTRLmemory} for details.
}

\resizebox{\textwidth}{!}{%
\begin{tabular}{@{}cccccccc@{}}
\toprule
\thead{\textbf{Method}}             & \thead{\textbf{Compute}}    & \thead{\textbf{Memory}}    & \thead{\textbf{Parallel}}      & \thead{\textbf{Unbiased}} & \thead{\textbf{Optimize}\\\textbf{ Non-Diff.}} & \thead{\textbf{Smoothed}} \\ \midrule
BPTT{\footnotesize~\citep{rumelhart1985learning}}           & $T(F+B)$               & $TS$           & \xmark  & \cmark & \xmark & \xmark  \\
TBPTT{\footnotesize~\citep{williams1990efficient}}          & $K(F+B)$               & $KS$                       & \xmark  & \xmark & \xmark & \xmark  \\
ARTBP{\footnotesize~\citep{tallec2017unbiasing}}            & $K(F+B)$               & $KS$                       & \xmark  & \cmark & \xmark & \xmark  \\
RTRL{\footnotesize~\citep{williams1989learning}}            & $PS^2 + S(F+B)$                & $SP + S^2$  & \xmark  & \cmark & \xmark & \xmark  \\
UORO{\footnotesize~\citep{tallec2017unbiased}}              & $F + B + S^2 + P$                  & $S + P$                      & \xmark  & \cmark & \xmark & \xmark  \\
Reparam.{\footnotesize~\citep{metz2019understanding}}       & ${\color{purple}N} T(F+B)$               & ${\color{purple}N} TS$                       & \cmark  & \cmark & \xmark & \cmark \\
ES{\footnotesize~\citep{rechenberg1973}}             & ${\color{purple}N}TF$                  & ${\color{purple}N}S$     & \cmark  & \cmark & \cmark & \cmark  \\
Trunc. ES{\footnotesize~\citep{metz2019understanding}}      & ${\color{purple}N}KF$                  & ${\color{purple}N}S$     & \cmark  & \xmark & \cmark & \cmark  \\
\midrule
PES (Ours)                                                  & ${\color{purple}N}$$KF$                  & ${\color{purple}N}(S+P)$ & \cmark  & \cmark & \cmark & \cmark  \\
PES + Analytic (Ours)                                                  & ${\color{purple}N}$$KF + K(F+B)$                   & ${\color{purple}N}(S+P) + (K+1)S$  & \cmark  & \cmark & \xmark & \cmark  \\
\bottomrule
\end{tabular}}
\vspace{-0.4cm}
\label{table:computation-comparison}
\end{table*}

In this section, we discuss additional related work on online learning algorithms, and on one special class of unrolled optimization problems: hyperparameter optimization (HO).
Table~\ref{table:computation-comparison} compares several approaches to gradient estimation in unrolled computation graphs, with respect to compute, memory, parallelization, unbiasedness, and smoothing.
In addition, Table~\ref{table:hparam-comparison} in Appendix~\ref{app:ho} provides a comparison of the HO algorithms mentioned in this section.
\vspace{-0.2cm}
\paragraph{Online Learning Algorithms.}
Real-time recurrent learning (RTRL) performs forward-mode gradient accumulation: it does not require storage of past states, but requires matrix-matrix products and storage of a matrix $G_t$ of size $\text{dim}(\bolds_t) \times \text{dim}(\boldtheta)$.
When $\text{dim}(\boldtheta)$ is large, as in RNN training, the cost of storing $G_t$ and the cost of computing the required matrix-matrix products is prohibitive. Several approaches propose efficient variants of RTRL based on cheaper, noisy approximations of $G_t$. Unbiased Online Recurrent Optimization (UORO)~\citep{tallec2017unbiased} uses an unbiased rank-1 approximation to the full matrix; Kronecker-Factored RTRL (KF-RTRL)~\citep{mujika2018approximating} uses a Kronecker product decomposition to approximate the RTRL update for a class of RNNs; and Optimal Kronecker Sum Approximation (OK)~\citep{benzing2019optimal} uses a similar approximation but with the lowest possible variance among methods within an approximation family.
\citet{cooijmans2019variance} also draw a connection between UORO and REINFORCE applied to estimate the gradient of an RNN by injecting noise into the hidden states.
In contrast, PES injects noise into the parameters.

\vspace{-0.2cm}
\paragraph{Hyperparameter Optimization (HO)}
There are three main approaches that can be categorized based on the types of problem-specific information used:
1) \textit{black-box} approaches that do not consider the internal structure of the objective $L$;
2) \textit{gray-box} approaches that make use of the fact that the objective is the result of an iterative optimization procedure (e.g. by using the validation performance of a model); and
3) \textit{gradient-based} approaches that require access to the exact functional form of the objective $L$, and that require the objective to be differentiable in the hyperparameters.
Black-box approaches include grid search, random search~\citep{bergstra2012random}, Bayesian optimization (BO)~\citep{snoek2012practical}, and ES~\citep{salimans2017evolution,metz2019understanding}.
Gray-box approaches include Freeze-Thaw BO~\citep{swersky2014freeze}, successive halving~\citep{jamieson2016non}, Hyperband~\citep{li2017hyperband}, Population-Based Training~\citep{jaderberg2017population}, and hypernetwork-based approaches to HO~\citep{lorraine2018stochastic,mackay2019self}.

A key advantage of gradient-based approaches is that they scale to high-dimensional hyperparameters (e.g. millions of hyperparameters)~\citep{lorraine2020optimizing}.
~\citet{maclaurin2015gradient} differentiate through unrolled optimization to tune many hyperparameters including learning rates and weight decay coefficients. These methods can perform poorly, however, when the underlying meta-loss is not smooth. Additionally they cannot optimize non-differentiable objectives, for example accuracy rather than loss.

PES can be considered a gray-box approach as it does not require the objective to be differentiable like gradient-based approaches, but it does take into account the iterative optimization of the inner problem.

\vspace{-0.1cm}
\section{Persistent Evolution Strategies}
\label{sec:pes}
\vspace{-0.1cm}

In this section, we introduce a method to obtain unbiased gradient estimates from partial unrolls of a computation graph, called Persistent Evolution Strategies (PES).
First, we derive the PES gradient estimator, prove that it is unbiased, and present a practical algorithm (Algorithm~\ref{alg:pes}).
Then we discuss the variance characteristics of PES, both theoretically and empirically.

\begin{figure*}[t]
\vspace{-0.3cm}
\label{fig-algs}
\begin{minipage}{0.48\textwidth}
\begin{algorithm}[H]
  \caption{Truncated Evolution Strategies (ES) applied to partial unrolls of a computation graph.}
  \label{alg:es}
\begin{algorithmic}
    \State \textbf{Input:} $\bolds_0$, initial state
    \State \hspace{2.7em} $K$, truncation length for partial unrolls
    \State \hspace{2.7em} $N$, number of particles
    \State \hspace{2.9em} $\sigma$, standard deviation of perturbations
    \State \hspace{2.9em} $\alpha$, learning rate for ES optimization
    \State Initialize $\bolds = \bolds_0$ ${\color{white}\bolds^{(i)} = \bolds_0}$
    \State {\color{white} Initialize $\boldxi^{(i)} \gets \boldzero$ for $i \in \{ 1, \dots, N \}$}
    \While {true}
        \State $\ges \gets \boldzero$
        \For{$i=1,\dots, N$}
            \State $\boldepsilon^{(i)} = 
                \left\{\begin{array}{lcl}
                	\text{draw from } \mathcal{N}(0, \sigma^2 I) &  & i \text{ odd} \\
                	-\boldepsilon^{(i-1)} &  & i \text{ even}
                \end{array}\right.$            
            \State $\hat{L}_K^{(i)} \gets \text{unroll}(\bolds, \boldtheta + \boldepsilon^{(i)}, K)$
            \State {\color{white} $\boldxi^{(i)} \gets \boldxi^{(i)} + \boldepsilon^{(i)}$}
            \State $\ges \gets \ges + \boldepsilon^{(i)} \hat{L}_K^{(i)}$
        \EndFor
        \State $\ges \gets \frac{1}{N \sigma^2} \ges$
        \State $\bolds \gets \text{unroll}(\bolds, \boldtheta, K)$
        \State $\boldtheta \gets \boldtheta - \alpha \ges$
    \EndWhile
\end{algorithmic}
\end{algorithm}
\end{minipage}
\hfill
\begin{minipage}{0.48\textwidth}
\begin{algorithm}[H]
  \caption{Persistent evolution strategies (PES). Differences from ES are {\color{purple} highlighted in purple.}}
  \label{alg:pes}
\begin{algorithmic}
    \State \textbf{Input:} $\bolds_0$, initial state
    \State \hspace{2.7em} $K$, truncation length for partial unrolls
    \State \hspace{2.7em} $N$, number of particles
    \State \hspace{2.9em} $\sigma$, standard deviation of perturbations
    \State \hspace{2.9em} $\alpha$, learning rate for PES optimization
    \State Initialize ${\color{purple}\bolds^{(i)}} = \bolds_0$ for {\color{purple} $i \in \{1, \dots, N\}$}
    \State {\color{purple} Initialize $\boldxi^{(i)} \gets \boldzero$ for $i \in \{ 1, \dots, N \}$}
    \While {true}
        \State $\gpes \gets \boldzero$
        \For{$i=1,\dots, N$}
            \State $\boldepsilon^{(i)} = 
                \left\{\begin{array}{lcl}
                	\text{draw from } \mathcal{N}(0, \sigma^2 I) &  & i \text{ odd} \\
                	-\boldepsilon^{(i-1)} &  & i \text{ even}
                \end{array}\right.$            
            \State {\color{purple} $\bolds^{(i)}$}, $\hat{L}_K^{(i)} \gets \text{unroll}({\color{purple} \bolds^{(i)}}, \boldtheta + \boldepsilon^{(i)}, K)$  
            \State {\color{purple} $\boldxi^{(i)} \gets \boldxi^{(i)} + \boldepsilon^{(i)}$}
            \State $\gpes \gets \gpes + {\color{purple} \boldxi^{(i)}} \hat{L}_K^{(i)}$
        \EndFor
        \State $\gpes \gets \frac{1}{N \sigma^2} \gpes$
        \State {\color{white}$s \gets \text{unroll}(\bolds, \theta, K)$}
        \State $\boldtheta \gets \boldtheta - \alpha \gpes$
    \EndWhile
\end{algorithmic}
\end{algorithm}
\end{minipage}
\vspace{-0.2cm}
\caption{\textbf{A comparison of vanilla ES and PES gradient estimators}, applied to partial unrolls of a computation graph.
The conditional statement for $\boldepsilon^{(i)}$ is used to implement antithetic sampling.
For clarity, we describe the meta-optimization updates to $\boldtheta$ using SGD, but we typically use Adam in practice.
See Appendix~\ref{app:diagram-alg} for diagrammatic representations of these algorithms.
}
\label{fig:both-algorithms}
\vspace{-0.2cm}
\end{figure*}

\vspace{-0.2cm}
\paragraph{Derivation.\footnote{See Appendix~\ref{app:pes-derivation} for an expanded derivation, and Appendix~\ref{app:pes-stochastic-computation-graph} for an alternate derivation using stochastic computation graphs~\citep{schulman2015gradient}.}}
Unrolled computation graphs (as illustrated in Figure~\ref{fig:unrolled-graph}) depend on shared parameters $\boldtheta$ at every timestep; in order to account for how these contribute to the overall gradient $\nabla_\boldtheta L(\boldtheta)$, we use subscripts $\boldtheta_t$ to distinguish between applications of $\boldtheta$ at different steps, where $\boldtheta_t = \boldtheta, \forall t$.
We further define $\Theta = (\boldtheta_1, \dots, \boldtheta_T)^\top$, which is a matrix with the per-timestep $\boldtheta_t$ as its rows. 
For notational simplicity in the following derivation, we drop the dependence on $\bolds_t$ and explicitly include the dependence on each $\boldtheta_t$, writing $L_t(\bolds_t ; \boldtheta)$ as either $L_t(\boldtheta_1, \dots, \boldtheta_t)$ or simply $L_t(\Theta)$.
We wish to compute the gradient $\nabla_{\boldtheta} L(\boldtheta)$ of the total loss over all unrolls. We begin by writing this gradient in terms of the full gradient $\pd{L(\Theta)}{\ovec{\Theta}} \in \mathbb{R}^{PT \times 1}$, and then using ES to approximate $\pd{L(\Theta)}{\ovec{\Theta}}$,
\begin{align*}
\frac{d L(\boldtheta)}{d \boldtheta} &= \sum_{\tau=1}^T \pd{L\pp{\Theta}}{\boldtheta_\tau}
= \pp{\mb I \otimes \boldone^\top} \pd{L(\Theta)}{\ovec{\Theta}}
, \\
    \boldg^\text{PES} &= \pp{\mb I \otimes \boldone^\top} \expect{\boldepsilon}{
        \frac{1}{\sigma^2} 
        \ovec{\boldepsilon} L\pp{\Theta + \boldepsilon}
        } \\
    &= \frac{1}{\sigma^2} \expect{\boldepsilon}{
        \left( \sum_{\tau=1}^T \boldepsilon_\tau \right) L\pp{\Theta + \boldepsilon}
        },
\end{align*}
where $\otimes$ denotes the Kronecker product, $\boldepsilon = \pp{\boldepsilon_1, \dots, \boldepsilon_T}^\top$ is a matrix of perturbations $\boldepsilon_t$ to be added to the $\boldtheta_t$ at each timestep, and the expectation is over entries in $\boldepsilon$ drawn from an i.i.d. Gaussian with variance $\sigma^2$. 
This ES approximation is an unbiased estimator of the gradient of the Gaussian-smoothed objective $\mathbb{E}_{\boldepsilon}[ L(\Theta + \boldepsilon)]$.
We next show that $\boldg^\text{PES}$ decomposes into a sum of sequential gradient estimates,
\begin{align}
    \vspace{-0.2cm}
    \boldg^\text{PES} &= 
        \frac{1}{\sigma^2} \expect{\boldepsilon}{
        \left( \sum_{\tau=1}^T \boldepsilon_\tau \right) \sum_{t=1}^T L_t\pp{\Theta + \boldepsilon}
        } \nonumber \\
        &= 
        \frac{1}{\sigma^2} \expect{\boldepsilon}{
        \sum_{t=1}^T \pp{\sum_{\tau=1}^t \boldepsilon_\tau}  L_t\pp{\Theta + \boldepsilon}
        } \label{eq T to t} \\
    &= \expect{\boldepsilon}{\sum_{t=1}^T \hat{\boldg}^\text{PES}_{t, \boldepsilon} }, \label{eq PES expectation} \\
    \hat{\boldg}^\text{PES}_{t, \boldepsilon} &= \frac{1}{\sigma^2} \boldxi_t L_t\pp{\boldtheta_1 + \boldepsilon_1, \dots, \boldtheta_t + \boldepsilon_t}. \label{eq pes step contribution}
    \vspace{-0.2cm}
\end{align}
where $\boldxi_t = \sum_{\tau=1}^t \boldepsilon_\tau$, Equation \ref{eq T to t} relies on $L_t\pp{\cdot}$ being independent of $\boldepsilon_\tau$ for $\tau > t$, and Equation \ref{eq pes step contribution} similarly relies on $L_t\pp{\cdot}$ only being a function of $\boldtheta_\tau$ for $\tau \leq t$.
The PES estimator consists of Monte Carlo estimates of Equation \ref{eq PES expectation},
\begin{align}
    \vspace{-0.2cm}
    \hat{\boldg}^\text{PES} &= \frac{1}{N} \sum_{i=1}^N \sum_{t=1}^T \hat{\boldg}^\text{PES}_{t, \boldepsilon^{(i)}}
    \vspace{-0.2cm}
    \label{eq:pes}
\end{align} 
where $\boldepsilon^{(i)}$ are samples of $\boldepsilon$, and $N$ is the number of Monte Carlo samples. Gradient estimates at each time step can be evaluated sequentially, and used to perform SGD.

\paragraph{PES with Antithetic Sampling.}
In practice, we use antithetic sampling to reduce variance. The PES estimator with antithetic sampling, which we denote $\gpesanti$, is given by:
\begin{align*}
    \gpesanti &= (\mb{I} \otimes \boldone^\top) \expect{\boldepsilon}{\frac{1}{2\sigma^2} \ovec{\boldepsilon} \pp{L(\Theta + \boldepsilon) - L(\Theta - \boldepsilon)}} \\
    &\approx \frac{1}{2 \sigma^2 N} \sum_{i=1}^N \sum_{t=1}^T \boldxi^{(i)}_t \pp{L_t(\Theta + \boldepsilon^{(i)}) - L_t(\Theta - \boldepsilon^{(i)})}
\end{align*}

\paragraph{PES is Unbiased for Quadratic Losses.}
See Appendix~\ref{app:pes-bias} for a proof of the following 
Statement~\ref{statement:unbiased}.
\begin{statement}[PES is unbiased] \label{statement:unbiased}
    Let $\boldtheta \in \mathbb{R}^P$ and $L(\boldtheta) = \sum_{t=1}^T L_t(\boldtheta)$.
    Suppose that $\nabla_{\boldtheta} L(\boldtheta)$ exists, and assume that $L$ is quadratic, so that it is equivalent to its second-order Taylor series expansion:
    $
    L(\Theta + \boldepsilon) = L(\Theta) + \veps^\top \nabla_{\vTheta} L(\Theta) + \frac{1}{2} \veps^\top \nabla^2_{\vTheta} L(\Theta) \veps
    $.
    Then, $\text{\emph{bias}}(\gpesanti) = \mathbb{E}_{\boldepsilon}[\gpesanti] - \nabla_{\boldtheta} L(\boldtheta) = \boldzero$.
\end{statement}

\vspace{-0.2cm}

\paragraph{Algorithm.}
Based on Eq.~\ref{eq:pes}, we see that we can obtain unbiased gradient estimates from partial unrolls by: 1) \textit{not resetting the particles} between unrolls, and 2) \textit{accumulating the perturbations $\boldxi_t$} each particle has experienced over multiple unrolls.
The resulting algorithm is simple to implement, requiring only minor modifications from vanilla ES.
Algorithm~\ref{alg:es} describes truncated ES applied to partial unrolls, where it suffers from short horizon bias. 
Algorithm~\ref{alg:pes} shows PES applied to the same problem, where it provides unbiased gradient estimates. 
Both algorithms (Fig.~\ref{fig:both-algorithms}) are shown with antithetic sampling (perturbations are paired with their negations), which drastically reduces variance.
\subsection{Variance Analysis}
We use the total variance, $\tr(\Var(\gpesanti))$, to quantify the variance of the estimator.
We provide a full derivation of the variance in Appendix~\ref{app:pes-variance}, and here we present some takeaways.
The variance depends on the gradients of each loss term $L_t$ with respect to each of the per-timestep parameters $\boldtheta_\tau$.
To gain insight into the structure of these gradients, we can arrange them in a matrix:
\begin{align}
M
=
\begin{bmatrix}
\nabla_{\boldtheta_1} L_1 & \nabla_{\boldtheta_1} L_2 & \nabla_{\boldtheta_1} L_3 & \cdots & \nabla_{\boldtheta_1} L_T \\
0 & \nabla_{\boldtheta_2} L_2 & \nabla_{\boldtheta_2} L_3 & \cdots & \nabla_{\boldtheta_2} L_T \\
0 & 0 & \nabla_{\boldtheta_3} L_3 & \cdots & \nabla_{\boldtheta_3} L_T \\
\vdots & \vdots & \vdots & \ddots & \vdots \\
0 & 0 & 0 & \cdots & \nabla_{\boldtheta_T} L_T \\
\end{bmatrix}
\label{eq:grad-matrix}
\end{align}
$M$ is upper-triangular due to the fact that $\nabla_{\boldtheta_\tau} L_t = 0$ for all $\tau > t$.
The variance of the PES estimator depends on the \textit{covariance between the gradients} $\nabla_{\boldtheta_\tau} L_t$ in this matrix.
\begin{table}[t]
\centering
\begin{tabular}{@{}cc@{}}
\toprule
\thead{\textbf{Scenario}} & \thead{\textbf{$\tr \pp{\Var \pp{\gpesanti}}$}} \\
\midrule
Diagonal $M$, i.i.d. grads   &  $\norm{\boldg} \pp{\frac{1}{2} PT + \frac{1}{2} P + T}$ \\[6pt]
Diagonal $M$, identical grads & $\norm{\boldg} \pp{\frac{1}{2T} P + \frac{1}{2} P + 1}$  \\[6pt]
Upper-tri $M$, i.i.d. grads & $\norm{\boldg} \mathcal{O}(T^2 + PT)$ \\[6pt]
Upper-tri $M$, identical grads & $\norm{\boldg} \mathcal{O}(\frac{P}{T})$  \\
\bottomrule
\end{tabular}
\vspace{-0.1cm}
\caption{\textbf{The variance of the PES estimator depends on the covariance of gradients across timesteps.} The variance of the gradient estimate is given as a function of the number of parameters $P$, unrolls $T$, and true gradient norm $\norm{\boldg}$. Each row corresponds to different structure in the gradient matrix $M$ (Equation \ref{eq:grad-matrix}). In the best case, subdividing a sequence into more PES unrolls $T$ reduces the variance by a factor of $1 \over T$. See Appendix~\ref{app:pes-variance} for details, and Figure \ref{fig:variance} for empirical variance scaling on an RNN task.
}
\label{table:variance-scenarios}
\vspace{-0.2cm}
\end{table}
\begin{figure}[t]
    \centering
    \includegraphics[width=0.75\linewidth]{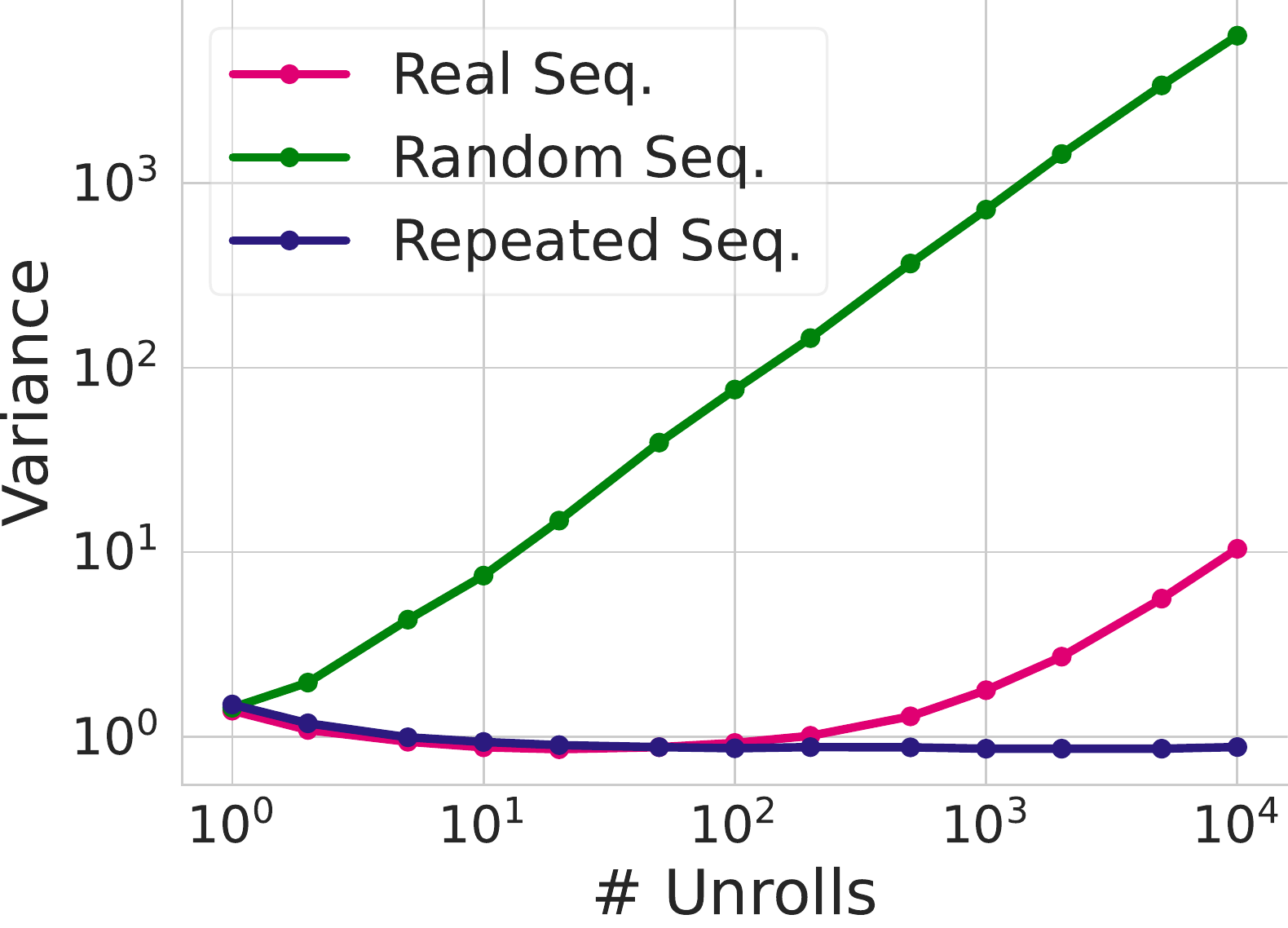}
    \vspace{-0.3cm}
    \caption{\textbf{The empirical variance of the PES estimator for a simple task initially {\em decreases}, before increasing linearly, as a sequence is subdivided into a larger number of unrolls $T$.} The task is training an RNN on the Penn TreeBank character-level language modeling dataset (``Real Seq.''). Also shown is the scaling behavior when the RNN is trained on either i.i.d. tokens (``Random Seq.'') or identical tokens (``Repeated Seq.''). The variance of the gradient estimate for a real sequence lies between these extremes.
    }
    \vspace{-0.5cm}
    \label{fig:variance}
\end{figure}
\paragraph{Variance Scenarios.}
We consider two structures for $M$: 1) a \textit{diagonal} structure, where the gradients $\nabla_{\boldtheta_i} L_j = 0, \forall i \neq j$; and 2) an \textit{upper-triangular} structure.
For each of these two matrix structures, we consider two possibilities for the covariance between gradients: a) all gradients $\nabla_{\boldtheta_i} L_j$ are identical; b) all gradients are i.i.d.
The total variance for each of the four resulting scenarios is shown in Table~\ref{table:variance-scenarios}.
It is possible for the $\boldg_t$ to have variance larger than  any of these scenarios, though we do not observe this in practice. 
\paragraph{Empirical Variance Measurements.}
To investigate the variance characteristics of PES empirically, we computed the variance in a toy setting.
We used an LSTM with 5 hidden units and 5-dimensional embeddings, for character-level language modeling on the Penn Treebank corpus~\citep{marcus1993building} (with a vocabulary consisting of 50 unique tokens).
We measured the variance of the $\gpesanti$ gradient estimate on a fixed sequence of $10^4$ characters.
The ground-truth gradient of the smoothed objective was computed using vanilla ES with 5000 particles over the full sequence (without truncation).
Figure~\ref{fig:variance} shows the variance of the PES gradient estimate using different numbers of unrolls, ranging from 1 (a single unroll for the full sequence) to $10^4$ (one unroll per input token).
Note that we do not update the parameters of the RNN after each unroll; we simply accumulate the gradient estimates over all partial unrolls.
We plot the variance normalized by the squared norm of the ground-truth gradient.
We observe an initial {\em drop} in variance, and then a linear growth.
Additional empirical variance measurements are presented in Figure~\ref{fig:lstm-variance-scenarios} (Appendix~\ref{app:pes-variance}).
\paragraph{Reducing Variance by Incorporating the Analytic Gradient.}
For functions $L$ that are differentiable, we can use the analytic gradient from the most recent partial unroll (e.g., backpropagating through the last $K$-step unroll) to reduce the variance of the PES gradient estimates.
In Appendix~\ref{app:analytic-gradient}, we show how we can incorporate the analytic gradient in the ES estimate for $\frac{\partial L_t(\Theta)}{\partial \boldtheta}$, deriving the following estimator:
\begin{align}
    \frac{\partial L_t(\Theta)}{\partial \boldtheta}
    &\approx \frac{1}{\sigma^2} \mathbb{E}_{\boldepsilon} \left[ \left( \sum_{\tau < t} \boldepsilon_\tau \right) (L_t(\Theta + \boldepsilon) - \boldepsilon_t^\top \boldp_t) \right] + \boldp_t
\end{align}
where $\boldp_t = \frac{\partial L_t(\Theta)}{\partial \boldtheta_t}$.
We call the resulting estimator PES+Analytic.
In Appendix~\ref{app:analytic-gradient} we describe the implementation of this estimator (Algorithm~\ref{alg:pes-analytic}), which requires a few simple changes from the standard PES estimator.
We also provide empirical variance measurements for PES+Analytic, using the same setup as was used for Figure~\ref{fig:variance}; we found that it can reduce variance by 1-2 orders of magnitude, given the same number of particles as PES.

\vspace{-0.2cm}
\section{Experiments}
\label{sec:experiments}
\vspace{-0.2cm}
First, we demonstrate via a toy experiment that PES does not suffer from truncation bias, allowing it to converge to correct solutions that are not found by TBPTT or truncated ES.
Then, we apply PES to several illustrative scenarios: we use PES to meta-train a learned optimizer, learn a policy for continuous control, and optimize hyperparameters.
All experiments used JAX~\citep{jax2018github}.
A simplified code snippet implementing PES is provided in Appendix~\ref{app:code}.

\vspace{-0.1cm}
\subsection{Influence Balancing}
\label{sec:influence-balancing}

\begin{figure}[t]
    \centering
    \includegraphics[width=0.75\linewidth]{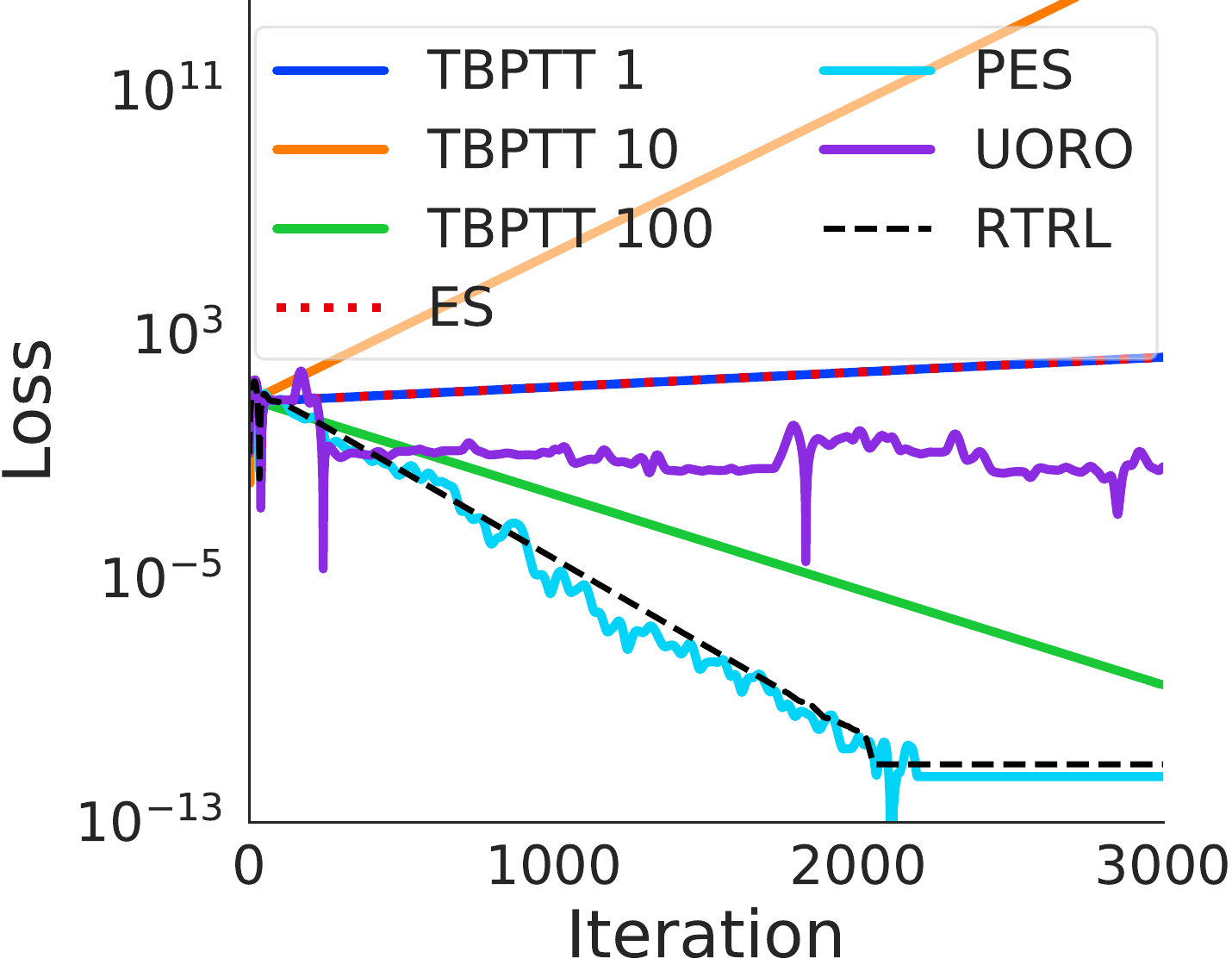}
    \vspace{-0.3cm}
    \caption{\textbf{Loss curves for the influence balancing task}. TBPTT with short truncations diverges, while PES performs nearly identically to exact RTRL. See Section~\ref{sec:influence-balancing} for experiment details.}
    \label{fig:influence-bal-fig}
    \vspace{-1cm}
\end{figure}

To demonstrate the lack of truncation bias of PES in a toy setting, we use the influence balancing task introduced by~\citet{tallec2017unbiased}. This simple task is particularly sensitive to short-horizon bias, as the gradient for the single parameter $\theta \in \mathbb R$ has the wrong sign when estimated from short unrolls. See Appendix~\ref{app:ib-details} for more details.
We use vanilla SGD to update $\theta$, with gradient estimates derived from TBPTT with different truncation horizons, exact RTRL, UORO, and PES.
TBPTT does not converge with short truncations $K \in \{ 1, 10 \}$; it requires much longer truncations ($K=100$) to move in the right direction.
In Figure~\ref{fig:influence-bal-fig}, we show that PES achieves nearly identical performance to exact RTRL.
UORO reaches the same performance as RTRL after approximately $30k$ iterations.
Note that the purpose of this experiment is to demonstrate that PES is unbiased, and is able to match the performance of exact RTRL given sufficiently many particles ($N=10^3$) to reduce variance; it is not intended as a comparison of total compute.

\subsection{Learned Optimizer Meta-Optimization}
\vspace{-0.1cm}
\label{sec:learnedopt}

In this section we demonstrate PES's applicability for learned optimizer training. We meta-train an MLP-based learned optimizer as described in \citet{metz2019understanding}.
This optimizer is used to train a two hidden-layer, 128 unit, MLP on CIFAR-10 with a batch size of 128.
Our meta-objective is the average training loss.
We train with a total number of inner-steps of $T=1000$ and a truncation length of $K=4$, using both PES and truncated ES.
\begin{figure}[t]
    \centering
    \includegraphics[width=0.86\linewidth]{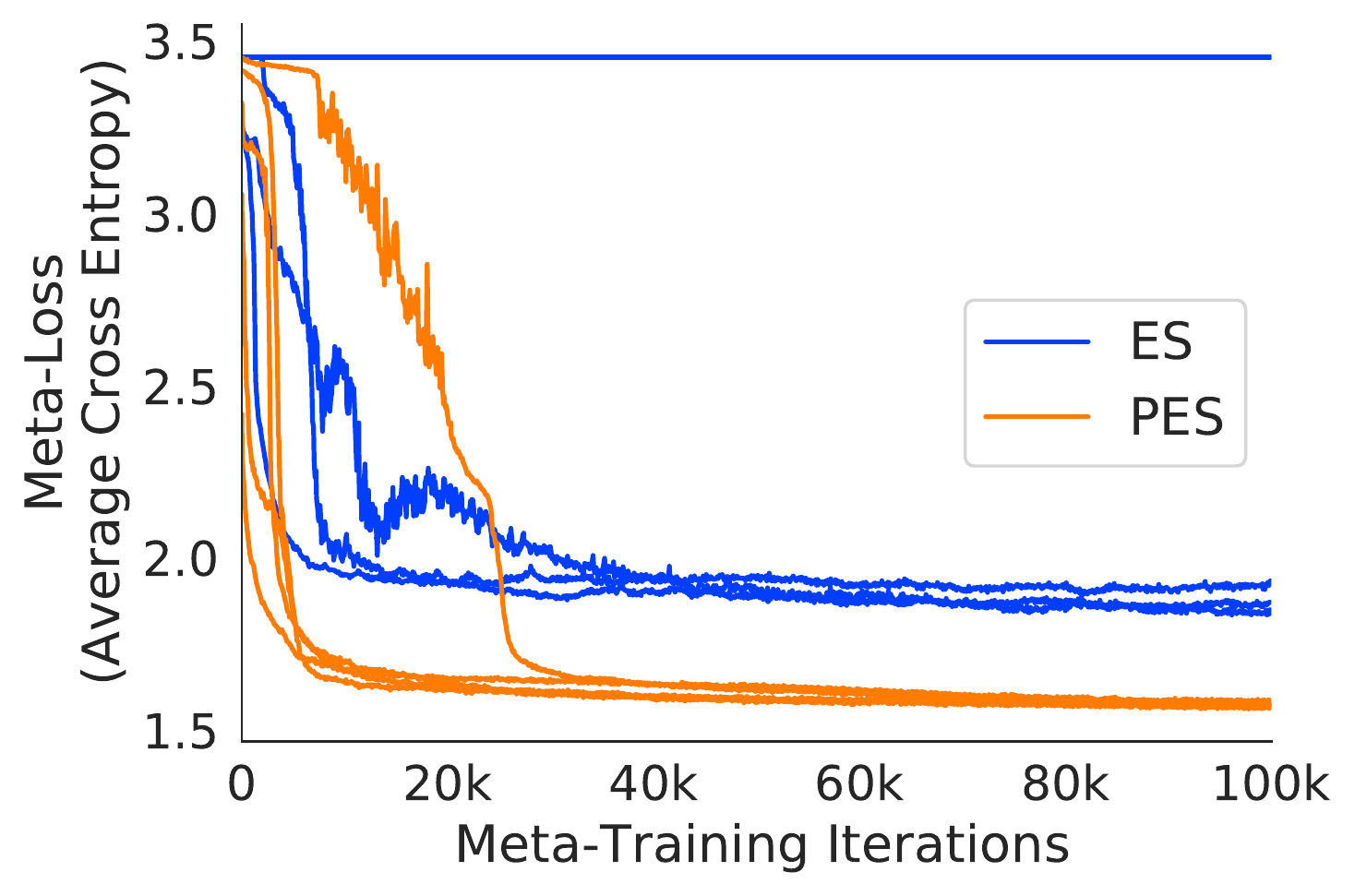}
    \vspace{-0.3cm}
    \caption{\textbf{Training learned optimizers}. We find that PES achieves better performance compared to truncated ES.
    Curves of the same color denote different initializations of the learned optimizer.
    See Section~\ref{sec:learnedopt} for details.
    }
    \vspace{-0.5cm}
    \label{fig:learned_opt}
\end{figure}
We outer-train with Adam, using a learning rate of $10^{-4}$ selected via grid search over half-orders of magnitude for each method independently. We use gradient clipping of $3$ applied to each gradient coordinate. We outer-train on 8 TPUv2 cores with asynchronous, batched updates of size $16$. To evaluate, we compute the meta-loss averaged over 20 inner initializations over the course of meta-training. Results can be found in Figure~\ref{fig:learned_opt}. Due to PES's unbiased nature, PES achieves both lower losses, and is more consistent across random initializations of the learned optimizer.

\begin{figure*}[t]
    \centering
    \includegraphics[width=0.36\linewidth]{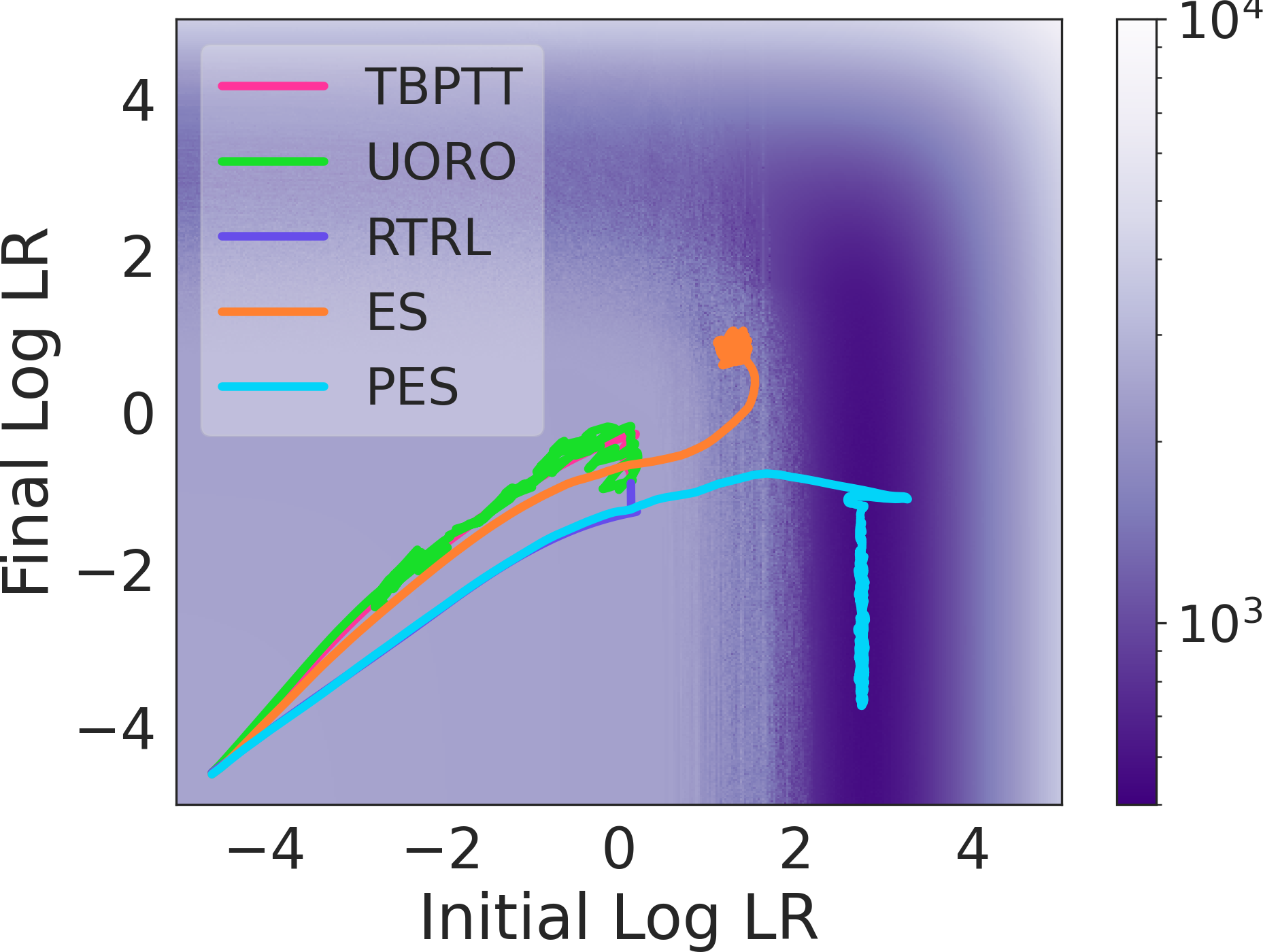}
    \qquad \qquad
    \includegraphics[width=0.36\linewidth]{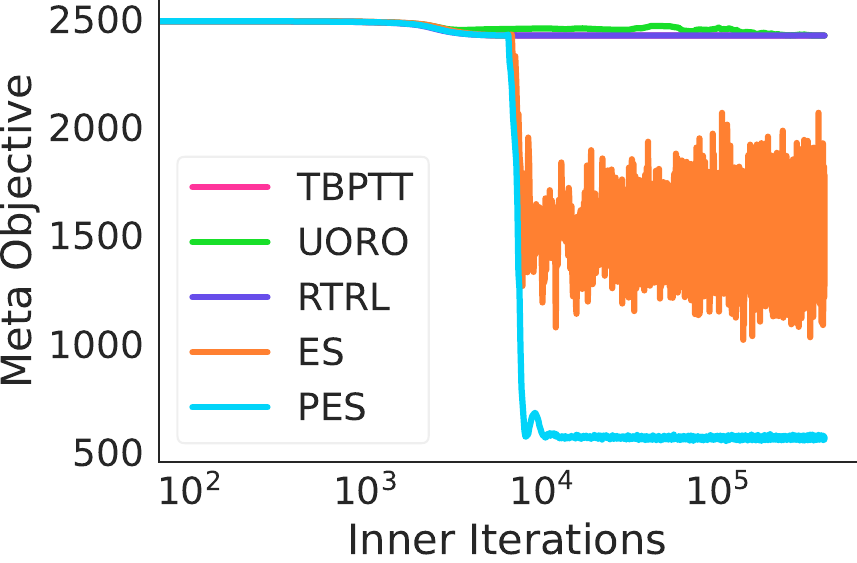}
    \vspace{-0.3cm}
    \caption{\textbf{The meta-objective surface (left), and meta-objective vs inner problem timesteps (right)}, for the 2D regression problem in Section \ref{sec:2d-regression}. We plot meta-optimization trajectories for TBPTT, UORO, RTRL, ES, and PES starting from the same initialization, $(-4.5, -4.5)$ in log-space.
    All techniques except PES either suffer from truncation bias, or become stuck due to high-frequency structure in the meta-objective surface. PES is both unbiased, and smooths the outer-objective removing high-frequency structure.
    Ablations over the truncation length and number of particles for this task are provided in Appendix~\ref{app:sensitivity}.
    }
    \vspace{-0.3cm}
    \label{fig:toy2d}
\end{figure*}

\vspace{-0.1cm}
\subsection{Learning a Continuous Control Policy}
\vspace{-0.1cm}
\label{sec:policies}

\begin{figure}[h]
    \centering
    \includegraphics[width=0.82\linewidth]{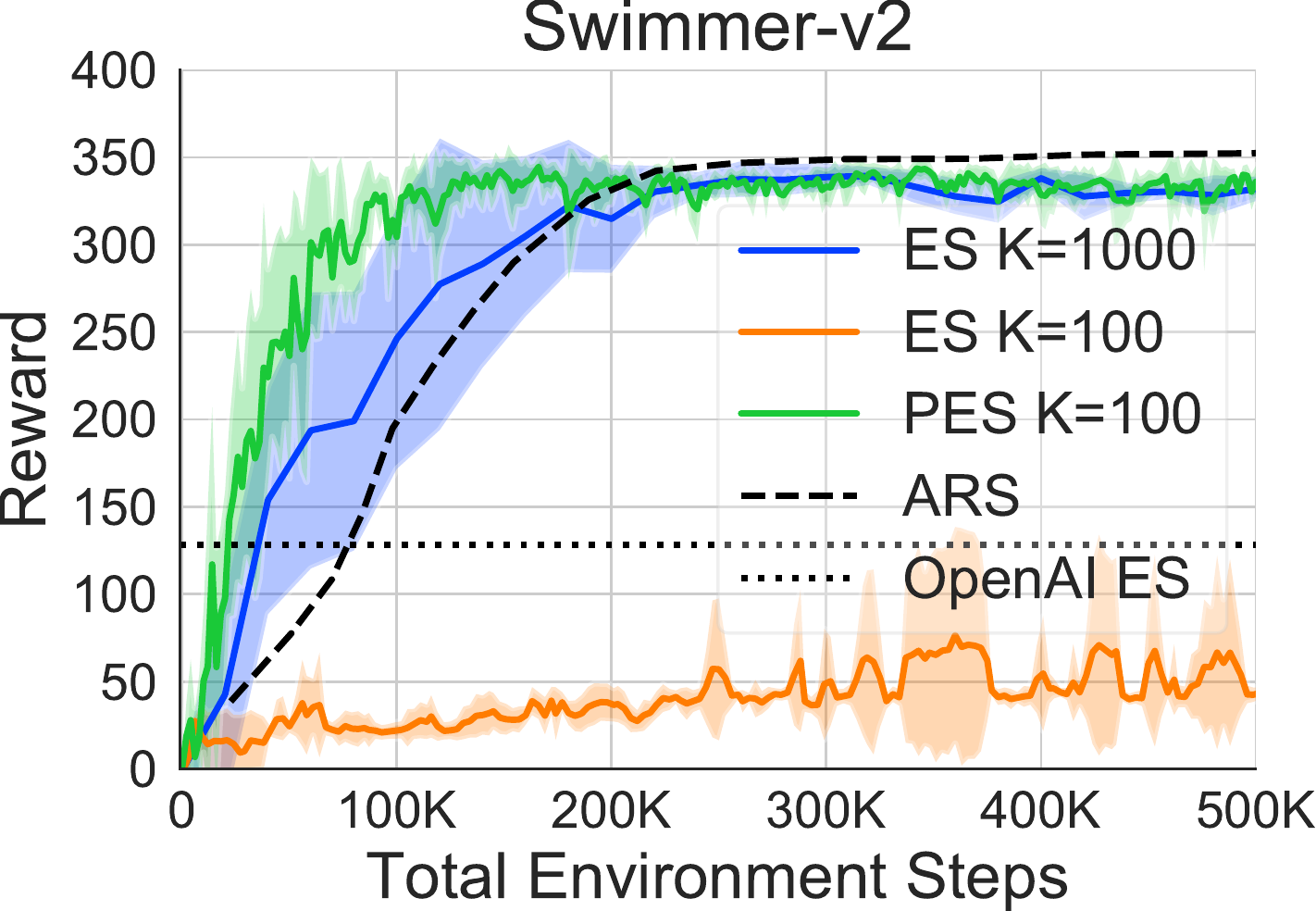}
    \vspace{-0.2cm}
    \caption{\textbf{Learning a policy for continuous control}.
    We find that PES applied to truncated unrolls performs similarly to ES applied to full episodes, while truncated ES fails due to bias.
    We plot the ARS V1 result from~\citet{mania2018simple} (dashed curve) to show that our full-unroll baseline is comparable to theirs.
    The dotted line shows the maximum reward reported for the ES approach in~\citet{salimans2017evolution}, which does not solve the Swimmer task.
    See Section~\ref{sec:policies} for details.
    }
    \label{fig:mujoco}
\end{figure}

Recent work~\citep{salimans2017evolution,mania2018simple} has shown that ES-based algorithms can be a viable alternative to more complex RL algorithms.
ES optimizes the parameters of a policy directly, by sampling parameters from a distribution, running an episode, and estimating the gradient; this is in contrast to standard RL algorithms that sample actions from a distribution output by a policy.
Here, we demonstrate that PES can be used to train a policy for a continuous control problem using partial unrolls, improving on the efficiency of vanilla ES typically applied to full unrolls.
We train a linear policy on the Swimmer-v2 MuJoCo environment, following~\citet{mania2018simple}.
For PES, the objective for each partial unroll is the sum of rewards over that unroll.
We also applied vanilla ES to the partial unrolls to demonstrate that this na\"ive strategy does not work---truncation bias occurs for these control problems as well.
Figure~\ref{fig:mujoco} compares vanilla ES applied to full episodes, ES applied to partial episodes, PES applied to partial episodes, and variants of ES from \citet{mania2018simple} and \citet{salimans2017evolution}.
To evaluate policies, we computed the average full-episode reward over 50 random environment seeds.
In Figure~\ref{fig:mujoco}, we show the mean performance of each algorithm over 6 random seeds, with standard deviation shown by the shaded region.
We see that PES reaches the same performance as full-unroll ES in slightly fewer total environment steps.

\vspace{-0.1cm}
\subsection{Hyperparameter Optimization}
\vspace{-0.1cm}
\label{sec:hyperopt}

In this section we demonstrate that PES can be used for hyperparameter optimization across four different problems. We show that PES performs well when the meta-loss has many local minima, does not suffer from truncation bias, can be applied to non-differentiable objectives, and can be used to optimize many hyperparameters (both continuous and discrete) simultaneously.

\vspace{-0.1cm}
\paragraph{Toy 2D Regression.}
\label{sec:2d-regression}
First, we used PES to meta-optimize a learning rate schedule for a toy 2D regression problem that has one global minimum, but many local minima to which truncated gradient methods could converge.
The inner optimization trajectories for different values of the outer-parameters are shown in Appendix~\ref{app:exp-details}.
We tuned a linear learning rate schedule parameterized by the initial and final log-learning rates, $\theta_0$ and $\theta_1$, respectively: $\alpha_t = \left(1 - \frac{t}{T}\right) e^{\theta_0} + \frac{t}{T} e^{\theta_1}$.
In Figure~\ref{fig:toy2d} we compare TBPTT, UORO, RTRL, ES, and PES applied to this meta-optimization task.
We found that the gradient-based methods (TBPTT, UORO, and RTRL) got stuck in suboptimal regions due to high-frequency structure in the meta-loss landscape.
ES makes more progress due to smoothing, but still suffers from truncation bias.
PES smooths the meta-objective surface and is unbiased, converging to a substantially better solution.

\begin{figure*}[ht]
    \centering
      \begin{overpic}[width=0.34\linewidth]{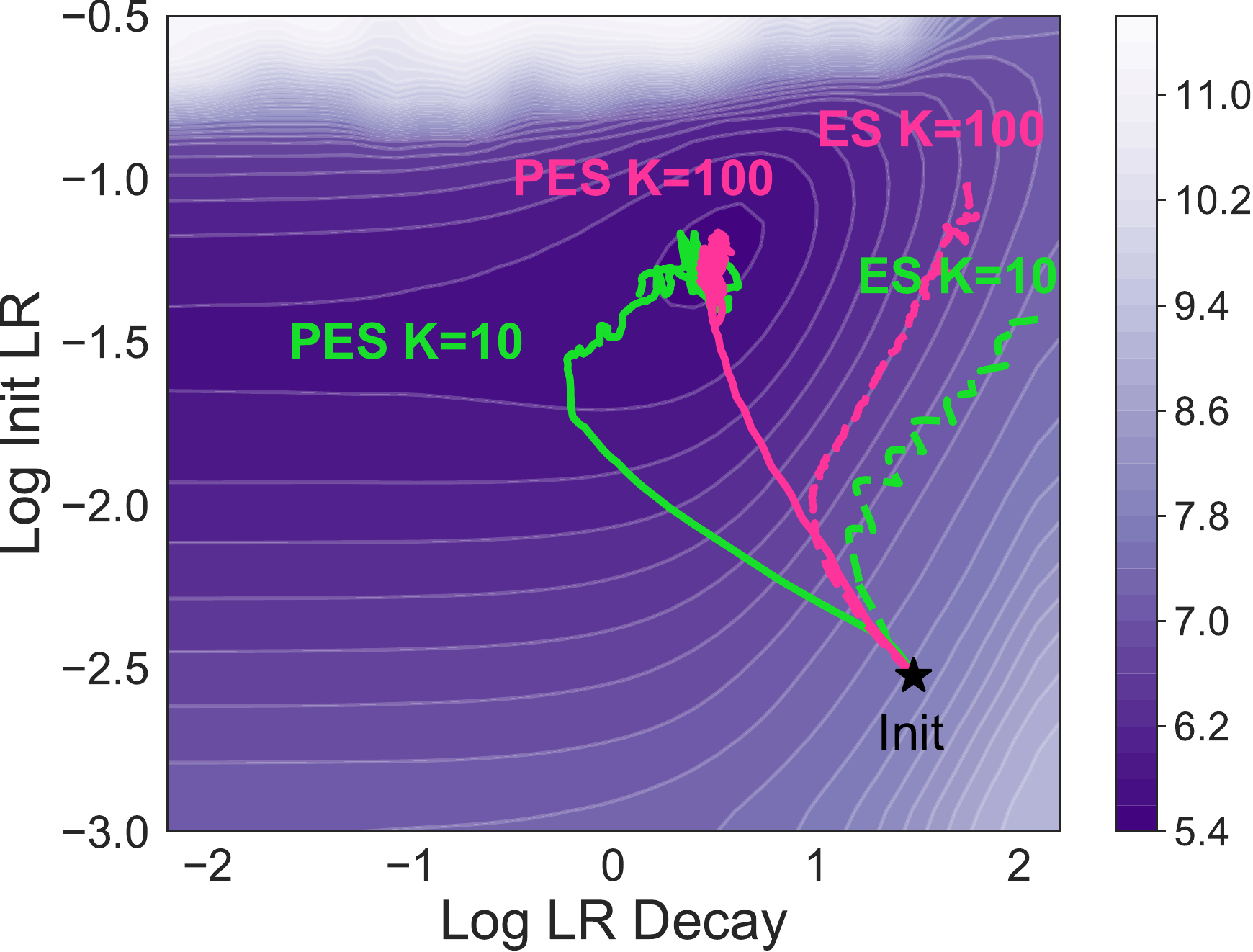}
     \put (0,1) {\textbf{\small(a)}}
    \end{overpic}    
    \qquad
    \begin{overpic}[width=0.34\linewidth]{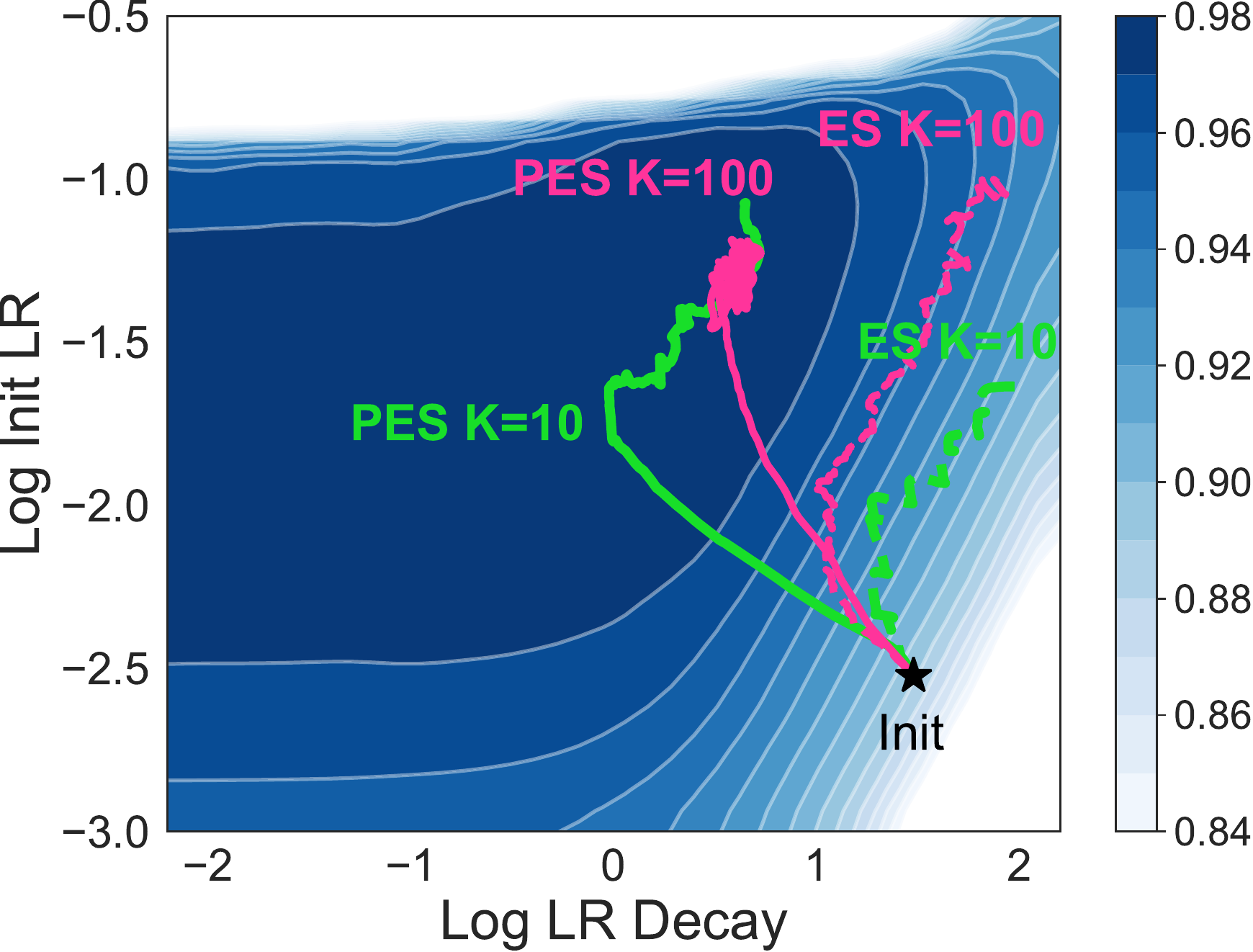}
     \put (0,1) {\textbf{\small(b)}}
    \end{overpic}   
    \vspace{-0.2cm}
    \caption{\textbf{Meta-optimization of a learning rate schedule for an MLP on MNIST.}
    We tune the initial learning rate and decay factor, both parameterized in log-space.
    Here we show the meta-loss landscape, and the optimization trajectories taken by ES and PES, for unroll lengths $K$ of 10 and 100.
    The meta-objective in \textbf{(a)} is the training loss while the meta-objective in \textbf{(b)} is the validation accuracy.
    In both visualizations, darker colors denote better values.
    Note that most gradient-based approaches are unable to target accuracy.
    }
    \vspace{-0.5cm}
    \label{fig:mnist-metaopt}
\end{figure*}

\vspace{-0.2cm}
\paragraph{MNIST MLP.}
\label{sec:mnist}
Next, we used PES to meta-learn a learning rate schedule for an MLP classifier on MNIST.
Following~\citet{wu2018understanding}, we used a two-layer MLP with 100 hidden units per layer and ReLU activations and the learning rate schedule parameterization $\alpha_t = \frac{\theta_0}{\left( 1 + \frac{t}{Q} \right)^{\theta_1}}$, where $\alpha_t$ is the learning rate at step $t$, $\theta_0$ is the initial learning rate, $\theta_1$ is the decay factor, and $Q$ is a constant fixed to 5000.
This schedule is used for SGD with fixed momentum 0.9.
The full unrolled inner problem consists of $T=5000$ optimization steps, and we apply vanilla ES and PES with truncation lengths $K \in \{10, 100\}$, yielding 500 and 50 unrolls per inner problem, respectively.
The meta-objective is the sum of training losses over the inner optimization trajectory.
In Figure~\ref{fig:mnist-metaopt}(a) we see that ES converges to a suboptimal region of the hyperparameter space due to truncation bias, while PES finds the correct solution.

\vspace{-0.2cm}
\paragraph{Targeting Validation Accuracy.} \label{sec:valid_acc}
Because PES only requires function evaluations and not gradients, it can optimize non-differentiable objectives such as accuracy rather than loss.
We demonstrate this by tuning the same parameterization of learning rate schedule as before, but using the accuracy on the MNIST validation set as the meta-objective.
Figure~\ref{fig:mnist-metaopt}(b) compares the meta-optimization trajectories of ES and PES on the validation accuracy meta-objective; again ES is biased and fails to converge to the right solution, while PES works well.

\vspace{-0.2cm}
\paragraph{Tuning Many Hyperparameters.}
\label{sec:many-hparams}
Here, we show that PES can tune several hyperparameters simultaneously, and achieves better performance than random search with an uninformative search space, using less compute.
We tuned both continuous and discrete hyperparameters: the number of units per hidden layer (discrete architectural hyperparameters) and per-parameter-block learning rates and momentum coefficients (continuous hyperparameters).
We trained a 5-hidden-layer MLP (6 layers including the output layer mapping to logits) on FashionMNIST, yielding 29 hyperparameters in total.
We set the maximum number of hidden units per layer to 100, and tuned sigmoid-transformed hyperparameters representing the fraction of hidden units to use.
The meta-objective was the sum of validation losses over the inner optimization trajectory.
\begin{figure}[h]
    \centering
    \vspace{-0.2cm}
    \includegraphics[width=0.8\linewidth]{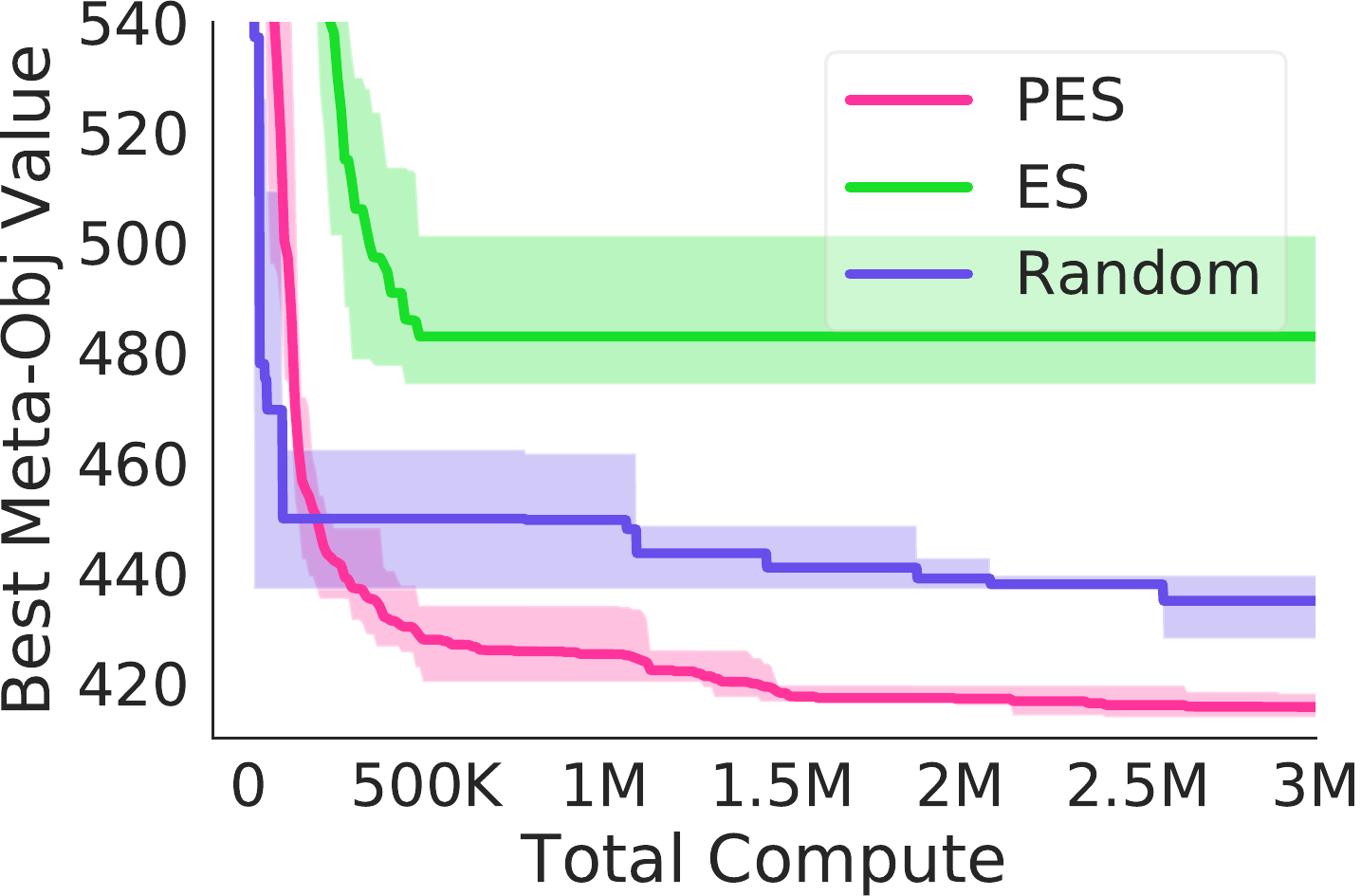}
    \vspace{-0.3cm}
    \caption{\textbf{Meta-optimization of per-parameter-block learning rates and momentum coefficients (29 hyperparameters total).}
    }
    \label{fig:many-hparams}
    \vspace{-0.3cm}
\end{figure}

Figure~\ref{fig:many-hparams} compares the best meta-objective values achieved by random search, vanilla ES, and PES, expressed in terms of the total number of inner iterations used (which accounts for the particles used in ES and PES).
We ran each method with four random seeds, and plot the mean (solid lines) and the min/max (shaded region) performance.
Each evaluation computes the mean meta-objective over 10 full inner problems using different random seeds for model initialization and data sampling.
PES outperforms ES and random search, achieving lower loss using less compute.

\vspace{-0.3cm}
\section{Conclusion}
\vspace{-0.2cm}
\label{sec:conclusion}
We introduced a method for unbiased gradient estimation in unrolled computation graphs, called Persistent Evolution Strategies (PES).
PES obtains gradients from truncated unrolls---which speeds up optimization by allowing for frequent parameter updates---while not suffering from truncation bias that affects many competing approaches.
We show that PES is broadly applicable, with experiments demonstrating its application to an RNN-like task, hyperparameter optimization, reinforcement learning, and meta-training of learned optimizers.

\section*{Acknowledgements}
\label{sec:acknowledgements}
We thank Sergey Ioffe and Niru Maheswaranathan for very helpful discussions and feedback on the paper.

\bibliography{icml2021}
\bibliographystyle{icml2021}

\appendix

\onecolumn

\icmltitle{Unbiased Gradient Estimation in Unrolled Computation Graphs \\with Persistent Evolution Strategies \\ \ \\ Supplementary Material}

\begin{icmlauthorlist}
\icmlauthor{Paul Vicol}{}
\icmlauthor{Luke Metz}{}
\icmlauthor{Jascha Sohl-Dickstein}{}
\end{icmlauthorlist}

\icmlaffiliation{toronto}{University of Toronto}
\icmlaffiliation{google}{Google Brain}

This appendix is structured as follows:
\begin{itemize}
    \item In Section~\ref{app:notation} we give an overview of the notation used in this paper.
    \item In Section~\ref{app:ho} we provide a table comparing several hyperparameter optimization approaches.
    \item In Section~\ref{app:exp-details} we provide experimental details.
    \item In Section~\ref{app:telescoping} we discuss telescoping sums as a way to target the final loss rather than the sum of losses as the meta-objective.
    \item In Section~\ref{app:pes-derivation} we provide a derivation of the PES estimator.
    \item In Section~\ref{app:pes-bias} we prove that PES is unbiased.
    \item In Section~\ref{app:pes-variance} we derive the variance of the PES estimator.
    \item In Section~\ref{app:analytic-gradient} we derive a variant of the PES estimator that incorporates the analytic gradient from the most recent partial unroll to reduce variance.
    \item In Section~\ref{app:pes-stochastic-computation-graph} we show the connection between PES and the framework for gradient estimation in stochastic computation graphs introduced in~\citet{schulman2015gradient}.
    \item In Section~\ref{app:deriv-RTRLmemory} we show derivations and compute/memory costs of the methods in Table~\ref{table:computation-comparison}.
    \item In Section~\ref{app:diagram-alg} we provide diagrammatic representations of the ES and PES algorithms.
    \item In Section~\ref{app:sensitivity} we provide an ablation study over the truncation length and number of particles for PES.
    \item In Section~\ref{app:code} we provide simplified code to implement PES in JAX~\citep{jax2018github}.
\end{itemize}

\clearpage

\section{Notation}
\label{app:notation}

Table~\ref{table:notation} summarizes the notation used in this paper.
\begin{table}[htbp]
\centering
\begin{tabular}{cc}
\toprule
\textbf{Symbol} & \textbf{Meaning} \\
\midrule
 ES        & Evolution strategies \\[3pt]
 PES       & Persistent evolution strategies \\[3pt]
 (T)BPTT   & (Truncated) backpropagation through time \\[3pt]
 RTRL      & Real time recurrent learning \\[3pt]
 UORO      & Unbiased online recurrent optimization \\[3pt]
 $T$        & The total sequence length / total unroll length of the inner problem \\[3pt]
 $K$          & The truncation length for subsequences / partial unrolls \\[3pt]
 $S$          & The dimensionality of the state of the unrolled system, $\text{dim}(\bolds)$  \\[3pt]
 $P$          & The dimensionality of the parameters of the unrolled system, $\text{dim}(\boldtheta)$  \\[3pt]
 $\boldtheta$ & The parameters of the unrolled system                    \\[3pt]
 $\boldtheta_t$ & The parameters of the unrolled system at time $t$, where $\boldtheta_t = \boldtheta, \forall t$   \\[3pt]
 $\Theta$     & A matrix whose rows are the parameters at each timestep, $\Theta = (\boldtheta_1, \dots, \boldtheta_T)^\top$   \\[3pt]
 $\bolds_t$        & The state of the unrolled system at time $t$             \\[3pt]
 $\boldx_t$        & The (optional) external input to the unrolled system at time $t$    \\[3pt]
 $f$            & The update function that evolves the unrolled system   \\[3pt]
 $N$          & The number of particles for ES and PES                   \\[3pt]
 $\sigma^2$   & The variance of the ES/PES perturbations                 \\[3pt]
 $\boldepsilon_t$ & A perturbation applied to the parameters $\boldtheta$ at timestep $t$\\[3pt]
 $\boldepsilon$ & A matrix whose rows are the perturbations at each timestep, $\boldepsilon = (\boldepsilon_1, \dots, \boldepsilon_T)^\top$ \\[3pt]
 $\boldxi_t$    & The sum of PES perturbations up to time $t$, $\boldxi_t = \boldepsilon_1 + \cdots + \boldepsilon_t$ \\[3pt]
 $L_t(\Theta)$   & The loss at timestep $t$, $L_t(\Theta) = L_t(\boldtheta_1, \dots, \boldtheta_t)$ \\[3pt]
 $L(\boldtheta)$, $L(\Theta)$   & The total loss, $L(\boldtheta) = L(\Theta) = \sum_{t=1}^T L_t(\Theta) = \sum_{t=1}^T L_t(\boldtheta_1, \dots, \boldtheta_t)$ \\[3pt]
 $\boldg_t$        & The true gradient at step $t$: $\nabla_{\boldtheta} L_t(\boldtheta)$     \\[3pt]
 $\ges$        & The vanilla ES gradient estimate (with Monte-Carlo sampling)     \\[3pt]
 $\gpes$       & The PES gradient estimate (with Monte-Carlo sampling)     \\[3pt]
 $\gpesanti$   & The antithetic PES gradient estimate (with Monte-Carlo sampling)     \\[3pt]
 $g\pp{t, \tau}$ & Shorthand for $\pd{L_t(\Theta)}{\boldtheta_{\tau}}$, used in variance expressions \\[3pt]
 $\otimes$ & Kronecker product \\[3pt]
 $\alpha$   & The learning rate for the parameters $\boldtheta$          \\[3pt]
 \multirow{3}{*}{$\text{unroll}(\bolds, \boldtheta, K)$} & A function that unrolls the system for $K$ steps \\
            & starting with state $\bolds$, using parameters $\boldtheta$. \\
            & Returns the updated state and loss resulting from the unroll \\
\bottomrule
\end{tabular}
\vspace{-0.2cm}
\caption{\textbf{Table of notation, defining the terms we use in this paper.}}
\label{table:notation}
\vspace{-0.2cm}
\end{table}

\pagebreak

\section{Hyperparameter Optimization Methods}
\label{app:ho}

Table~\ref{table:hparam-comparison} presents a comparison of several hyperparameter optimization approaches.
We distinguish between black-box, gray-box, and gradient-based approaches, and focus our comparison on whether each method can tune optimization hyperparameters, regularization hyperparameters, and discrete hyperparameters, as well as whether the method requires multiple runs through the inner problem or is online (operating within the timespan of a single inner problem), and whether the method is unbiased, meaning that it will eventually converge to the optimal hyperparameters.

\begin{table*}[h]
\centering
\begin{tabular}{@{}cccccccc@{}}
\toprule
\textbf{Method}             & \textbf{Type}  & \textbf{Parallel}     & \textbf{\begin{tabular}[c]{@{}c@{}}Tune\\ Opt\end{tabular}} & \textbf{\begin{tabular}[c]{@{}c@{}}Tune\\ Reg\end{tabular}} & \textbf{\begin{tabular}[c]{@{}c@{}}Tune\\ Discrete\end{tabular}} & \textbf{\begin{tabular}[c]{@{}c@{}}Online\\ (1 Run)\end{tabular}} & \textbf{Unbiased}     \\ \midrule
Grid/Random{\tiny ~\citep{bergstra2012random}}        & $\blacksquare$ & \cmark & \cmark & \cmark & \cmark & \xmark & \cmark \\
BayesOpt {\tiny ~\citep{snoek2012practical}}           & $\blacksquare$ & \cmark & \cmark & \cmark & \cmark & \xmark & \cmark \\
SHAC {\tiny ~\citep{kumar2018parallel}}               & $\blacksquare$ & \cmark & \cmark & \cmark & \cmark & \xmark & \xmark \\
Freeze-Thaw BO {\tiny ~\citep{swersky2014freeze}}     & $\greysquare$  & \cmark & \cmark & \cmark & \cmark & \xmark & \cmark \\
Full ES            & $\blacksquare$ & \cmark & \cmark & \cmark & \cmark & \xmark & \cmark \\
PBT{\tiny ~\citep{jaderberg2017population}}                & $\greysquare$  & \cmark & \cmark & \cmark & \cmark & \cmark & \xmark \\
Succ. Halving{\tiny ~\citep{jamieson2016non}} & $\greysquare$  & \cmark & \cmark & \cmark & \cmark & \xmark & \xmark \\
Hyperband{\tiny ~\citep{li2017hyperband}} & $\greysquare$  & \cmark & \cmark & \cmark & \cmark & \xmark & \xmark \\
TBPTT{\tiny ~\citep{domke2012generic}}   & $\nabla$       & \xmark & \cmark & \cmark & \xmark & \xmark &  \xmark \\
Full BPTT{\tiny ~\citep{maclaurin2015gradient}}   & $\nabla$       & \xmark & \cmark & \cmark & \xmark & \xmark &  \cmark \\
STN{\tiny ~\citep{mackay2019self}}                & $\nabla$       & \xmark & \xmark & \cmark & \cmark & \cmark & \xmark \\
IFT{\tiny ~\citep{lorraine2020optimizing}}                & $\nabla$       & \xmark & \xmark & \cmark & \xmark & \cmark & \cmark \\
HD{\tiny ~\citep{baydin2017online}}                 & $\nabla$       & \xmark & \cmark & \xmark & \xmark & \cmark & \xmark \\
MARTHE{\tiny ~\citep{donini2019scheduling}}             & $\nabla$       & \xmark & \cmark & \cmark      & \xmark & \cmark      & \xmark  \\
RTHO{\tiny ~\citep{franceschi2017forward}}               & $\nabla$       & \xmark & \cmark & \cmark     & \xmark  & \cmark & \cmark \\
\midrule
PES (Ours)               & $\greysquare$  & \cmark & \cmark & \cmark & \cmark & \xmark & \cmark \\
PES+Analytic (Ours)               & $\nabla$  & \cmark & \cmark & \cmark & \xmark & \xmark & \cmark \\
\bottomrule
\end{tabular}
\vspace{-0.2cm}
\caption{\textbf{Comparison between hyperparameter optimization approaches.} $\blacksquare$ denotes \textit{black-box}, $\greysquare$ denotes \textit{gray-box}, and $\nabla$ denotes \textit{gradient-based} approaches.
}
\label{table:hparam-comparison}
\end{table*}

\section{Experiment Details}
\label{app:exp-details}

In this section, we provide details for the experiments from Section~\ref{sec:experiments}.

\paragraph{Computing Infrastructure.}
All experiments except for learned optimizer training were run on NVIDIA P100 GPUs (using only a single GPU per experiment).
The learned optimizer experiment in Section~\ref{sec:learnedopt} was trained on 8 TPUv2 cores; we used asynchronous multi-TPU training for convenience, not necessity (these experiments could be run on a single GPU if desired).

\subsection{2D Toy Regression}
\label{app:2d-regression-details}

\begin{figure}[H]
    {
    \centering
    \includegraphics[width=0.3\linewidth]{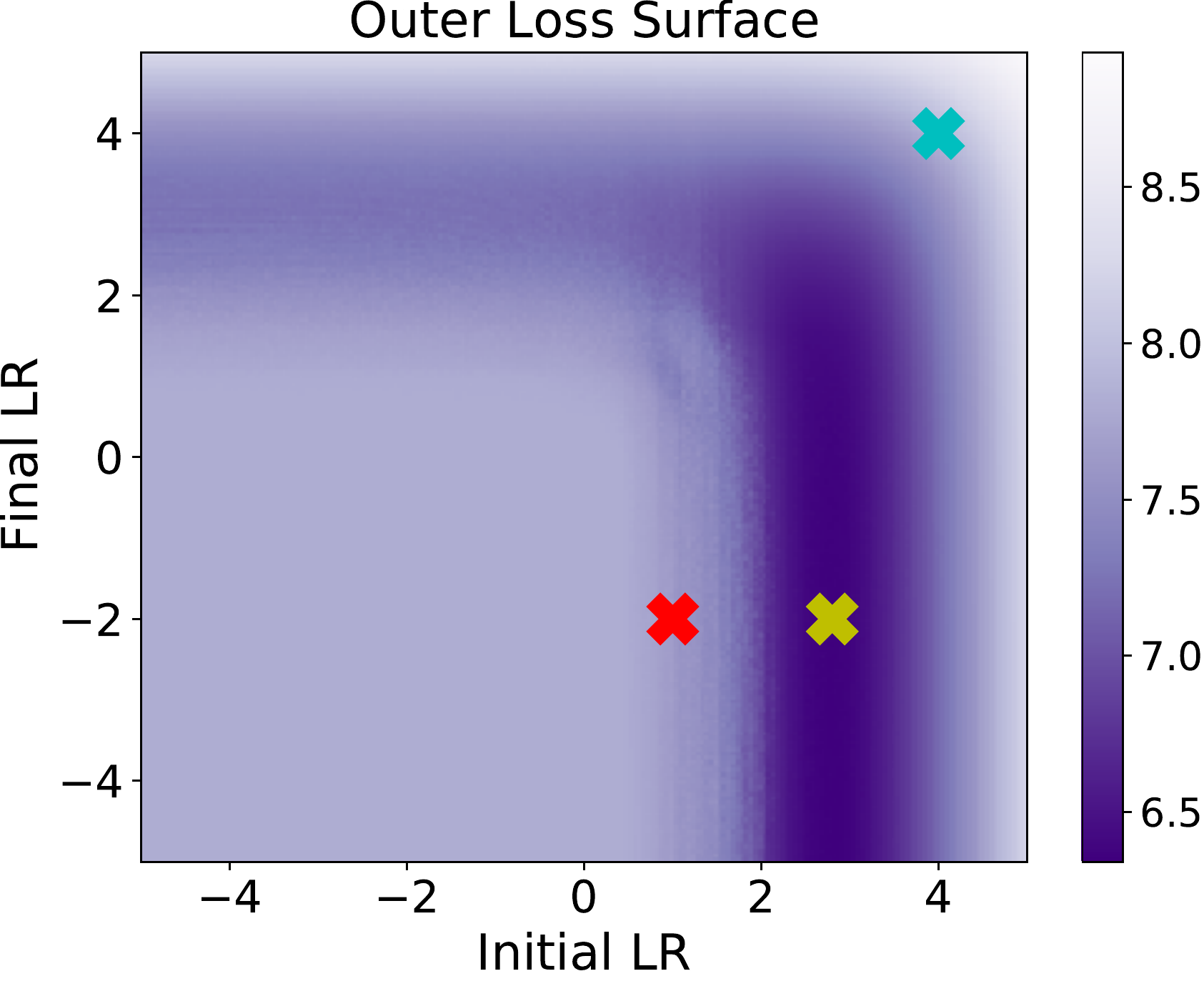}
    \hspace{0.2cm}
    \includegraphics[width=0.32\linewidth]{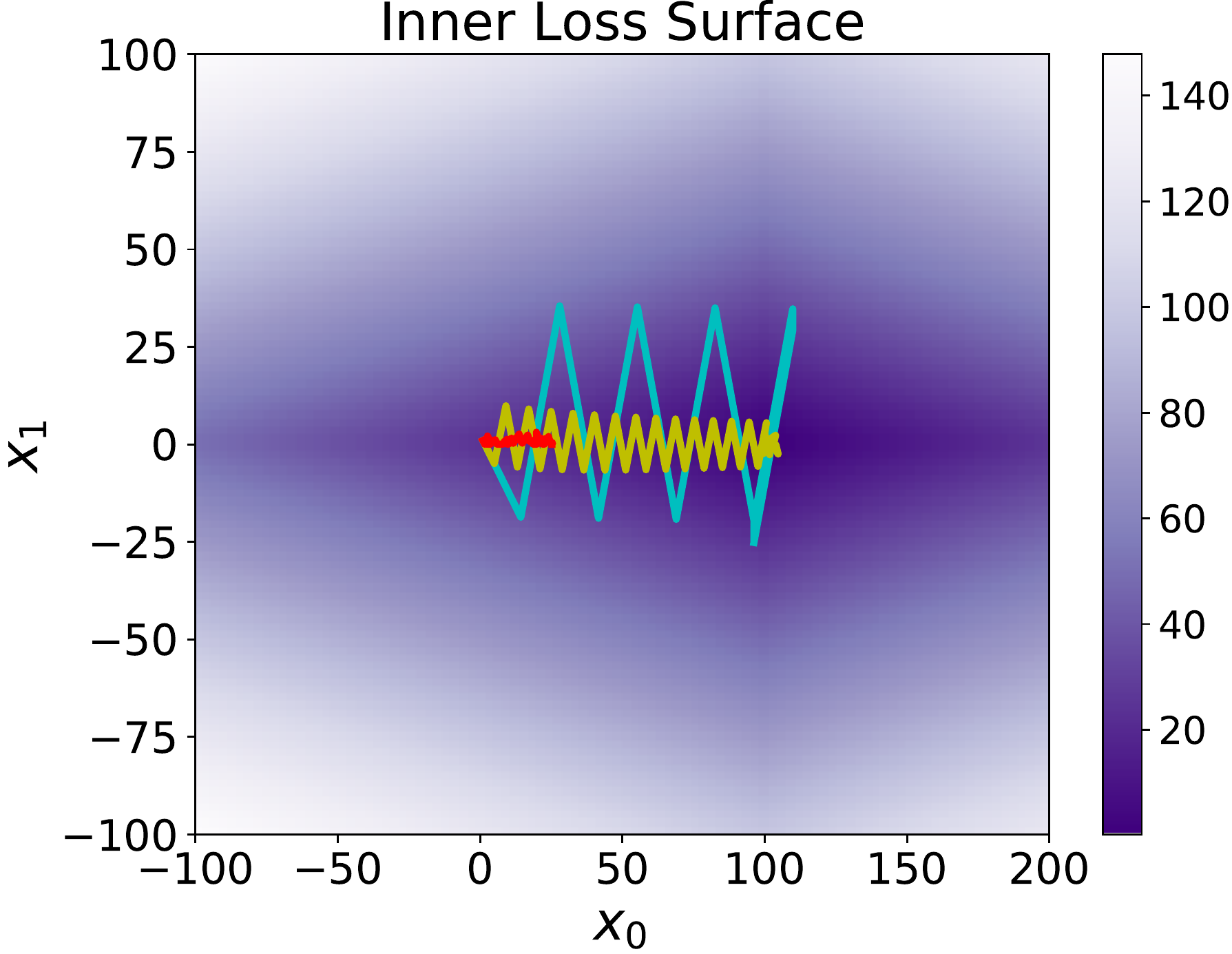}
    \hspace{0.2cm}
    \includegraphics[width=0.31\linewidth]{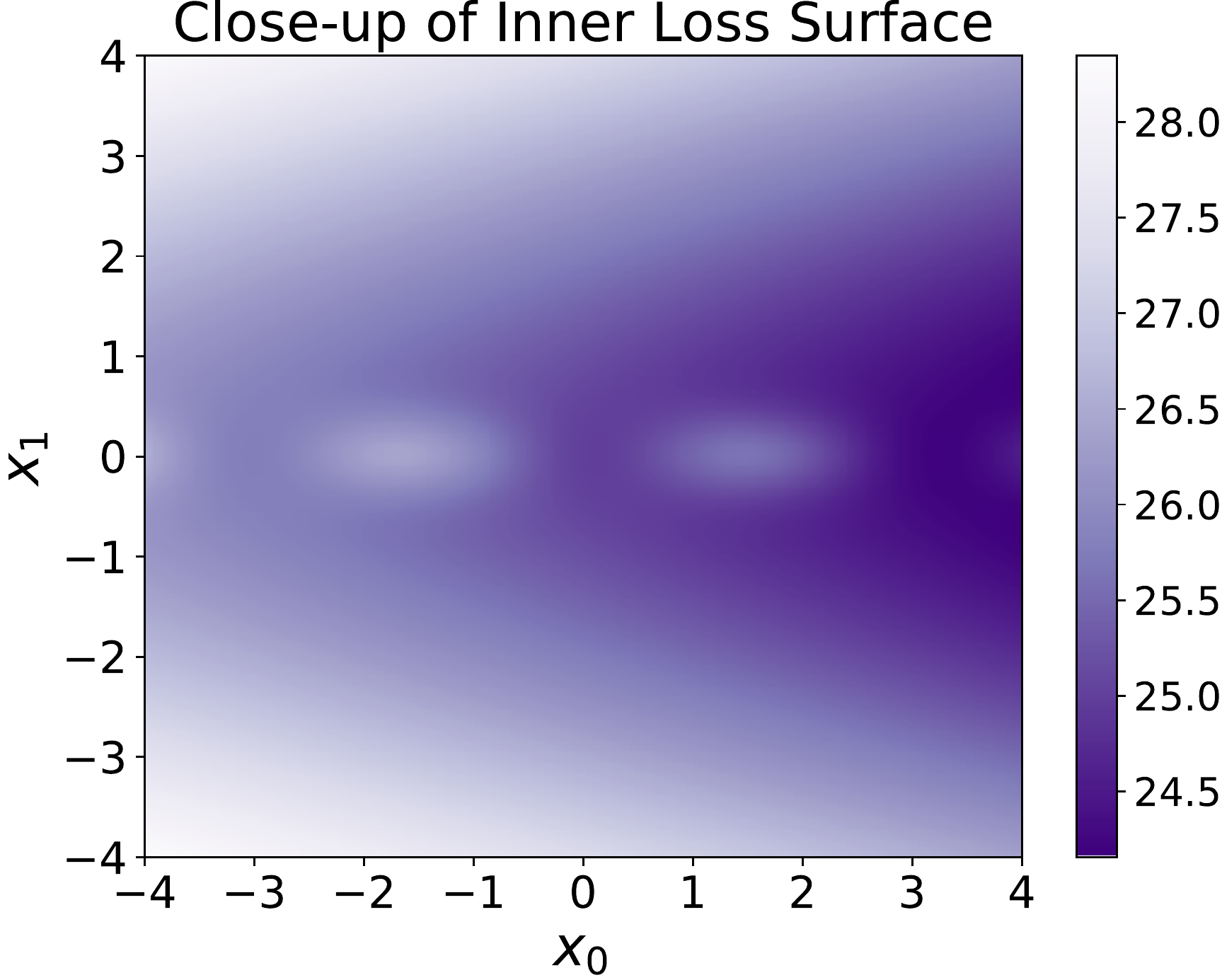}
    } \\
    \hspace*{2.1cm} \textbf{(a)} \hspace{5.1cm} \textbf{(b)} \hspace{5cm} \textbf{(c)}
    \caption{\textbf{Optimization landscape for the toy 2D regression problem.} \textbf{(a)} The outer loss (e.g. meta-loss) surface, showing the meta-objective values (e.g. the sum of losses over the inner optimization trajectory) for different settings of the two hyperparameters controlling the initial and final (log) learning rates of a linear decay schedule; \textbf{(b)} the inner loss surface, showing color-coded optimization trajectories corresponding to the hyperparameters highlighted in (a); \textbf{(c)} a close-up of the inner loss surface in the region where the parameters $(x_0, x_1)$ are initialized at the start of the inner problem.}
    \label{fig:toy2d-inner-trajectories}
    \vspace{-0.2cm}
\end{figure}

The inner objective is a toy 2D function defined as:
\begin{equation}
    f(x_0, x_1) = \sqrt{x_0^2 + 5} - \sqrt{5} + \sin^2(x_1) \exp(-5x_0^2) + 0.25|x_1 - 100|
\end{equation}
This was manually designed to be a challenging problem for any meta-optimization method that suffers from truncation bias.
In Figure~\ref{fig:toy2d-inner-trajectories} we visualize the outer loss surface (aka the meta-loss surface) and the inner loss surface for this task; we show the optimization trajectories on the inner loss surface corresponding to three different choices of optimization hyperparameters (shown by color-coded markers).

In our experiments, the total inner problem length was $T=100$, and we used truncated unrolls of length $K=10$.
For ES and PES, we used perturbation variance $\sigma^2 = 1$, and 100 particles (50 antithetic pairs).
We used Adam with learning rate 1e-2 as the outer optimizer for all methods (TBPTT, RTRL, UORO, ES, and PES).

\subsection{Influence Balancing}
\label{app:ib-details}

\begin{wrapfigure}[13]{r}{0.35\linewidth}
    \vspace{-0.5cm}
    \centering
    \includegraphics[width=\linewidth]{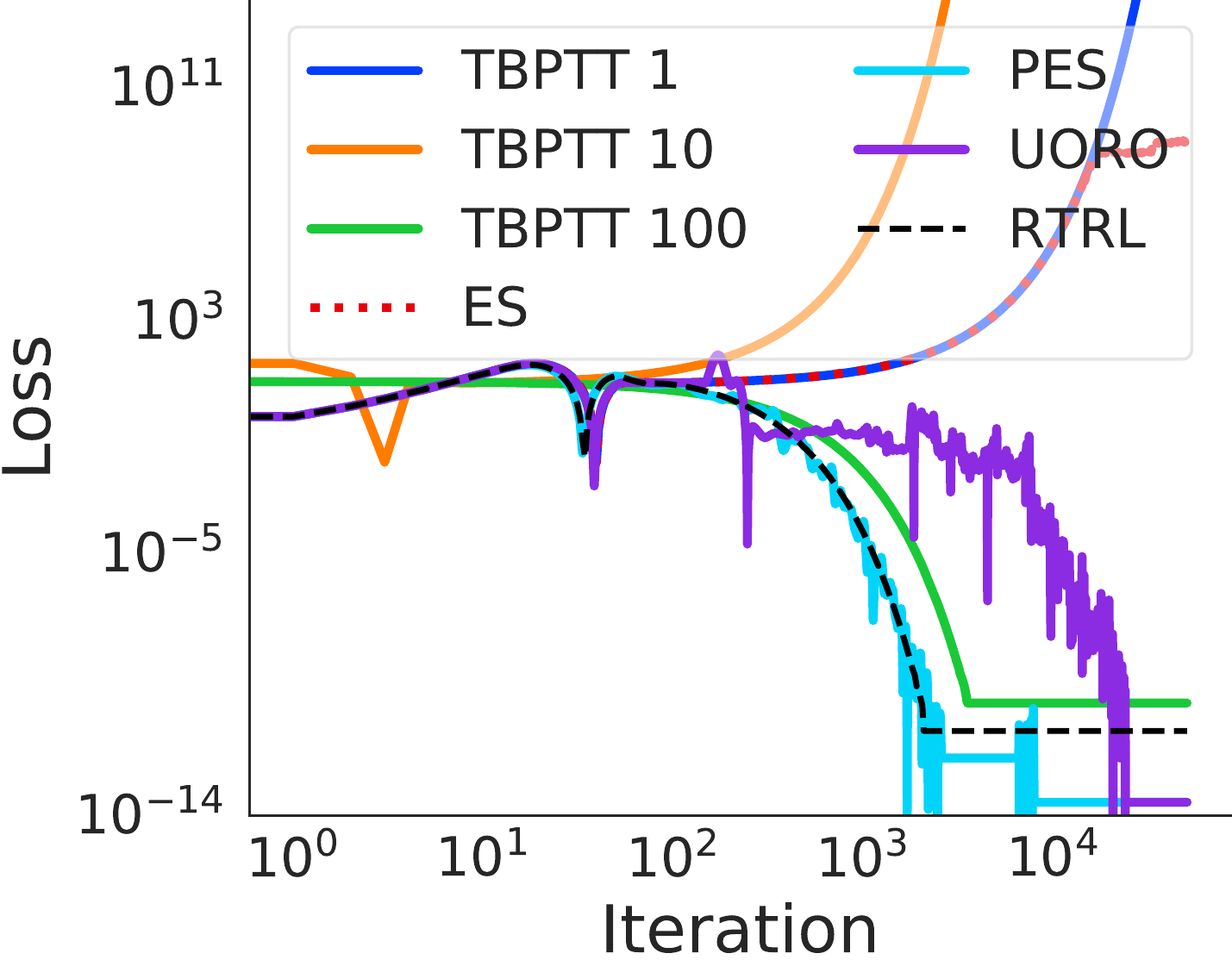}
    \caption{Longer run of influence balancing, with log-scaled x-axis.}
    \label{fig:influence-long-run}
\end{wrapfigure}
The influence balancing task considers learning a scalar parameter $\theta \in \mathbb{R}$ that governs the evolution of the following unrolled system:
\begin{equation}
    \bolds_{t+1} = A \bolds_t + (\underbrace{\theta, \dots, \theta}_{\text{$p$ positive}}, \underbrace{-\theta, \dots, -\theta}_{\text{$n-p$ negative}})^\top
\end{equation}
where $A$ is a fixed $n \times n$ matrix with $A_{i,i} = 0.5$, $A_{i,i+1} = 0.5$ and 0 everywhere else.
The vector on the right hand side consists of $\theta$ tiled $n$ times, with $p$ positive and $n-p$ negative copies.
In our experiments, we used $n=23$ and $p=10$.
The loss at each step is regression on the first index in the state vector $\bolds_t$:
\begin{equation}
    L_t = \frac{1}{2} (\bolds_t^0 - 1)^2
\end{equation}
For the influence balancing experiment, we used $n = 23$ with 10 positive and 13 negative $\theta$'s.
The state was initialized to a vector of ones, $\bolds_0 = \mathbf{1}$, and $\theta$ was initialized to 0.5.
We used gradient descent for optimization, with learning rate 1e-4.
We did not use learning rate decay as was used in~\citep{tallec2017unbiased}, as we did not find this to be necessary for convergence.
For ES and PES we used perturbation scale $\sigma = 0.1$ and $10^3$ particles.

\subsection{MNIST Experiments}
\label{app:mnist-details}

\paragraph{MNIST Meta-Optimization.}
Following~\citet{wu2018understanding}, we used a two-layer MLP with 100 hidden units per layer and ReLU activations and the learning rate schedule parameterization $\alpha_t = \frac{\theta_0}{\left( 1 + \frac{t}{Q} \right)^{\theta_1}}$, where $\alpha_t$ is the learning rate at step $t$, $\theta_0$ is the initial learning rate, $\theta_1$ is the decay factor, and $Q$ is a constant fixed to 5000.
This schedule is used for SGD with fixed momentum 0.9.
We used mini-batches of size 100.
The full unrolled inner problem consists of $T=5000$ optimization steps, and we used vanilla ES and PES with truncation lengths $K \in \{10, 100\}$, yielding 500 and 50 unrolls per inner problem.
The meta-objective is the sum of training softmax cross-entropy losses over the inner optimization trajectory.
We used Adam as the outer-optimizer, and for each method (ES and PES), we performed a grid search over the outer-learning rates $\{ 0.01, 0.03, 0.1 \}$ to find the most stable and fastest-converging setups.
For both ES and PES, we used perturbation standard deviation $\sigma=0.1$, and 1000 particles (500 antithetic pairs).

\vspace{-0.2cm}
\paragraph{Tuning Many Hyperparameters \& Comparison to Random Search.}

We trained on FashionMNIST with minibatch size $100$ for $T=1000$ inner problem steps, using truncations of length $K=10$, yielding 100 unrolls per inner problem.
For both ES and PES, we used $\sigma=0.3$ and used Adam with learning rate $1e-2$ as the outer optimizer.
We used an MLP with ReLU activations and 5 hidden layers (6 layers including the output layer mapping the final hidden representation to logits).
We tuned separate learning rates and momentum coefficients for SGD with momentum, for each weight matrix and bias vector in the network (this yields 24 hyperparameters, as we have 6 layers each with 2 parameter blocks and 2 hyperparameters tuned).
We also tuned the number of units per hidden layer, by masking the output of each hidden layer, with a deterministic mask that zeros out part of the representation, effectively using only the first $n$ units.
We tune the number of units in each of the 5 hidden layers, yielding 5 discrete hyperparameters, and 29 hyperparameters in total.
Because we are effectively tuning the architecture of the MLP, we apply hidden unit masking at evaluation time in addition to training time.
As the meta-objective, we used the sum of validation losses over the inner optimization trajectory.

To tune the number of hidden units, we used an unconstrained parameterization (in the real numbers) transformed by a sigmoid to the range $(0,1)$ which represents the \textit{fraction of units} that are used, out of the maximum number of units per layer (set to be 100 in our experiments).
The number of units per layer is determined by $\lfloor m_i * \text{sigmoid}(\theta_i)\rfloor$ where $m_i$ is the maximum number of units for hidden layer $i$ and $\theta_i$ is the unconstrained parameterization for the fraction of units to be used.

For random search, we sampled learning rates uniformly at random in log-space, with range $(1\text{e-8}, 1\text{e1})$; we sampled momentum coefficients uniformly at random in logit-space corresponding to the sigmoid-transformed range $(0.01, 0.999)$; and we sampled the number of hidden units per layer from the logit-space corresponding to the sigmoid-transformed range $(0.01, 0.999)$ .
For ES and PES, we initialized each learning rate uniformly at random in log space in the range $(1\text{e-4}, 1\text{e-2})$; we initialized each momentum coefficient uniformly at random in logit-space, to have the sigmoid-transformed range $(0.01, 0.9)$; and we initialized the number of hidden units per layer in logit-space corresponding to the sigmoid-transformed range $(0.2, 0.8)$.
These ranges are slightly smaller than the ones used for random search in order to maintain meta-optimization stability; note from Figure~\ref{fig:many-hparams} that the performance of both ES and PES is initially poor (prior to meta-optimization), indicating that these ranges for random initialization do not increase their performance compared to random search, and thus the improvement for PES is primarily due to its adaptation of the hyperparameters.
For ES and PES, we used perturbation standard deviation $\sigma=0.3$, $N=10$ particles, and Adam with learning rate 0.01 for outer optimization.
We ran each method four times with different random seeds, and plotted the mean performance, with the min and max shown by the shaded regions in Figure~\ref{fig:many-hparams}.
We measured the best meta-objective value achieved so far during meta-optimization, as a function of \textit{total compute}, which takes into account the number of inner iterations performed, as well as the number of parallel workers (or particles); total compute corresponds to the product of inner iterations and the number of workers.

\begin{figure}[H]
    \centering
    \begin{subfigure}[t]{0.33\linewidth}
        \includegraphics[width=\linewidth]{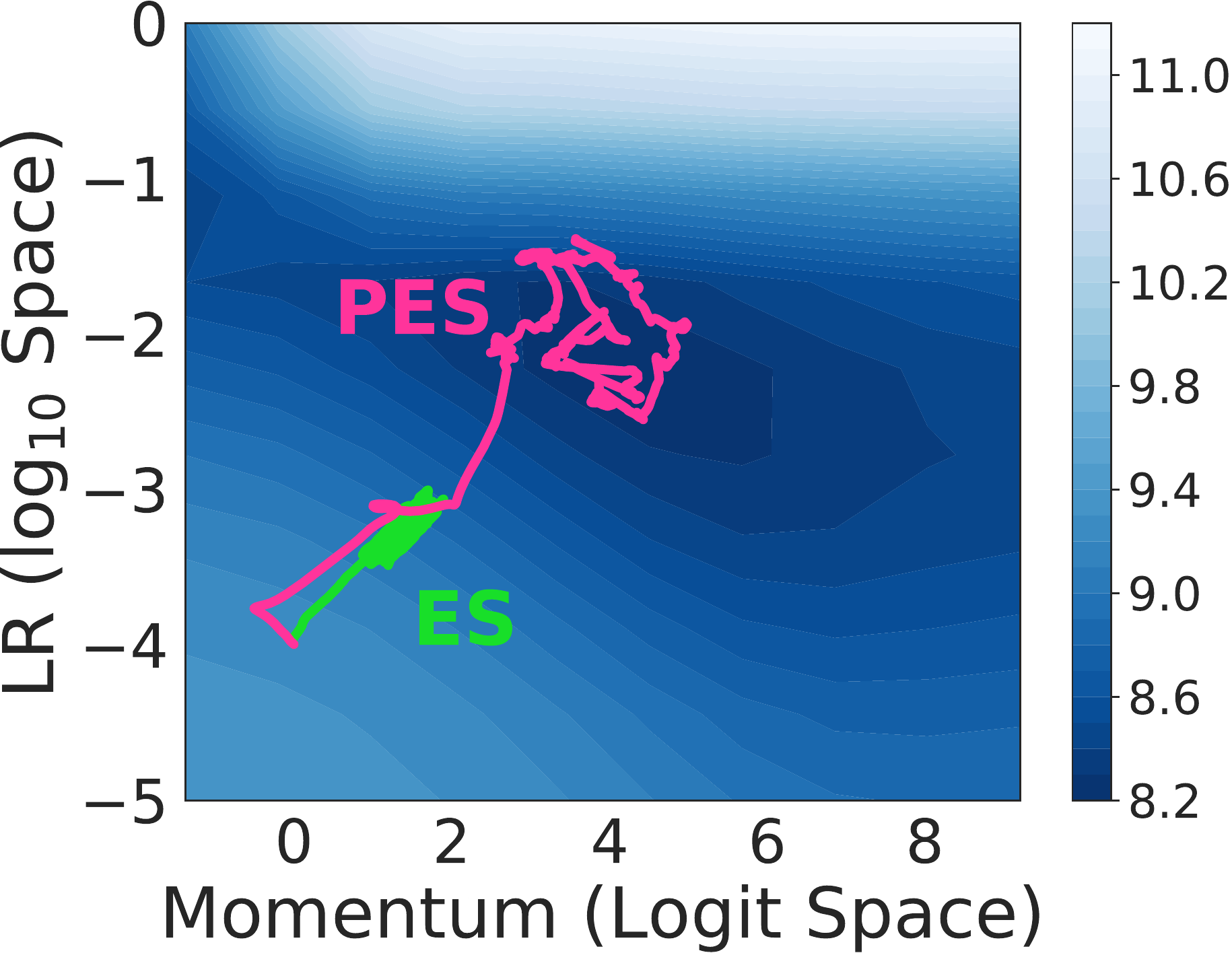}
        \caption{Tuning LR \& momentum for a Myrtle.ai ResNet on CIFAR-10.}
        \label{fig:cifar10-hopt}
    \end{subfigure}
    \qquad
    \begin{subfigure}[t]{0.4\linewidth}
        \includegraphics[width=\linewidth]{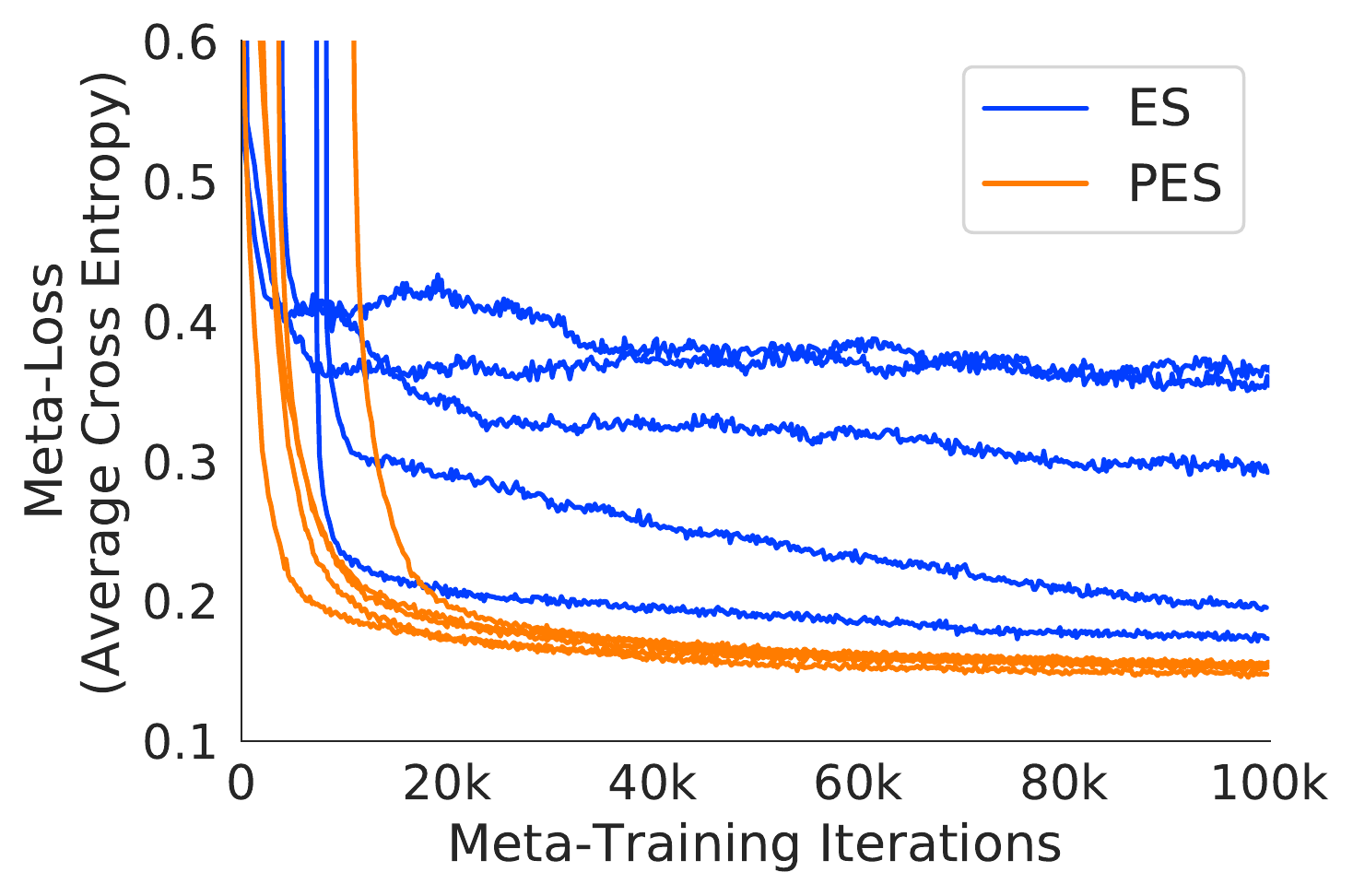}
        \caption{Learned optimizer trained on MNIST.}
        \label{fig:cifar10-lopt}
    \end{subfigure}
    \caption{CIFAR-10 experiment for hyperparameter optimization and MNIST experiment for learned optimizer training.}
    \label{fig:cifar10-exps}
\end{figure}

\paragraph{Additional Hyperparameter Optimization and Learned Optimizer Experiments.}
In Figure~\ref{fig:cifar10-hopt}, we tune hyperparameters for a 1.6M parameter ResNet on CIFAR-10 using ES and PES with $T=5000$, $K=20$, and $N=4$, targeting the sum of validation losses.
In Figure~\ref{fig:cifar10-lopt}, we train a learned optimizer on MNIST (similarly to~\citet{metz2019understanding}).
We use the same configuration as described in Section \ref{sec:learnedopt} but target a 2-hidden layer, 128 unit MLP trained on MNIST.

\paragraph{Tuning Regularization for UCI Regression.}
\begin{figure}[H]
    \centering
    \begin{subfigure}[t]{0.35\linewidth}
        \includegraphics[width=\linewidth]{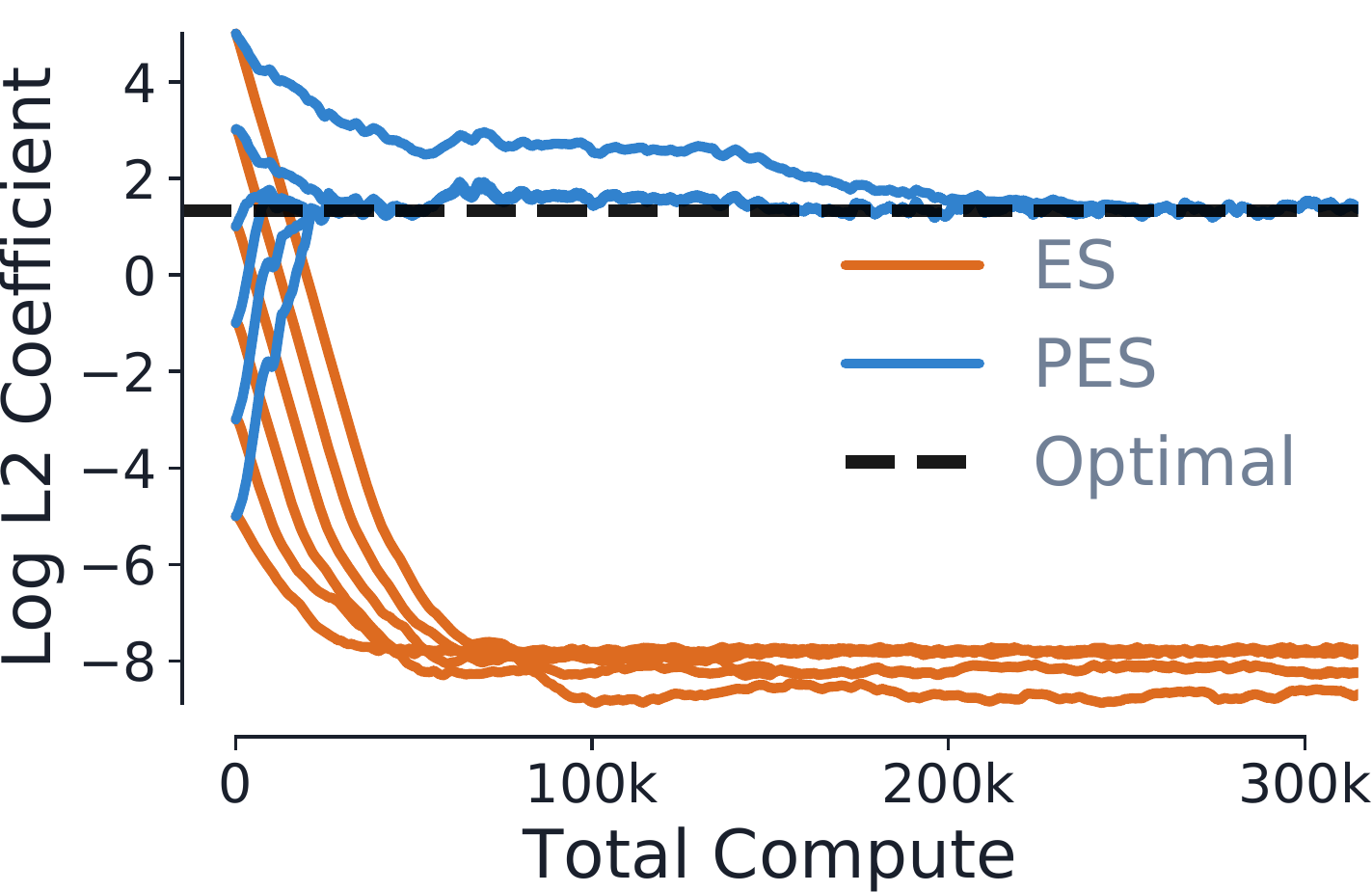}
        \caption{Meta-optimization trajectories using ES and PES from different initial $L_2$ coefficients.}
        \label{fig:uci-trajectories}
    \end{subfigure}
    \qquad
    \begin{subfigure}[t]{0.35\linewidth}
        \includegraphics[width=\linewidth]{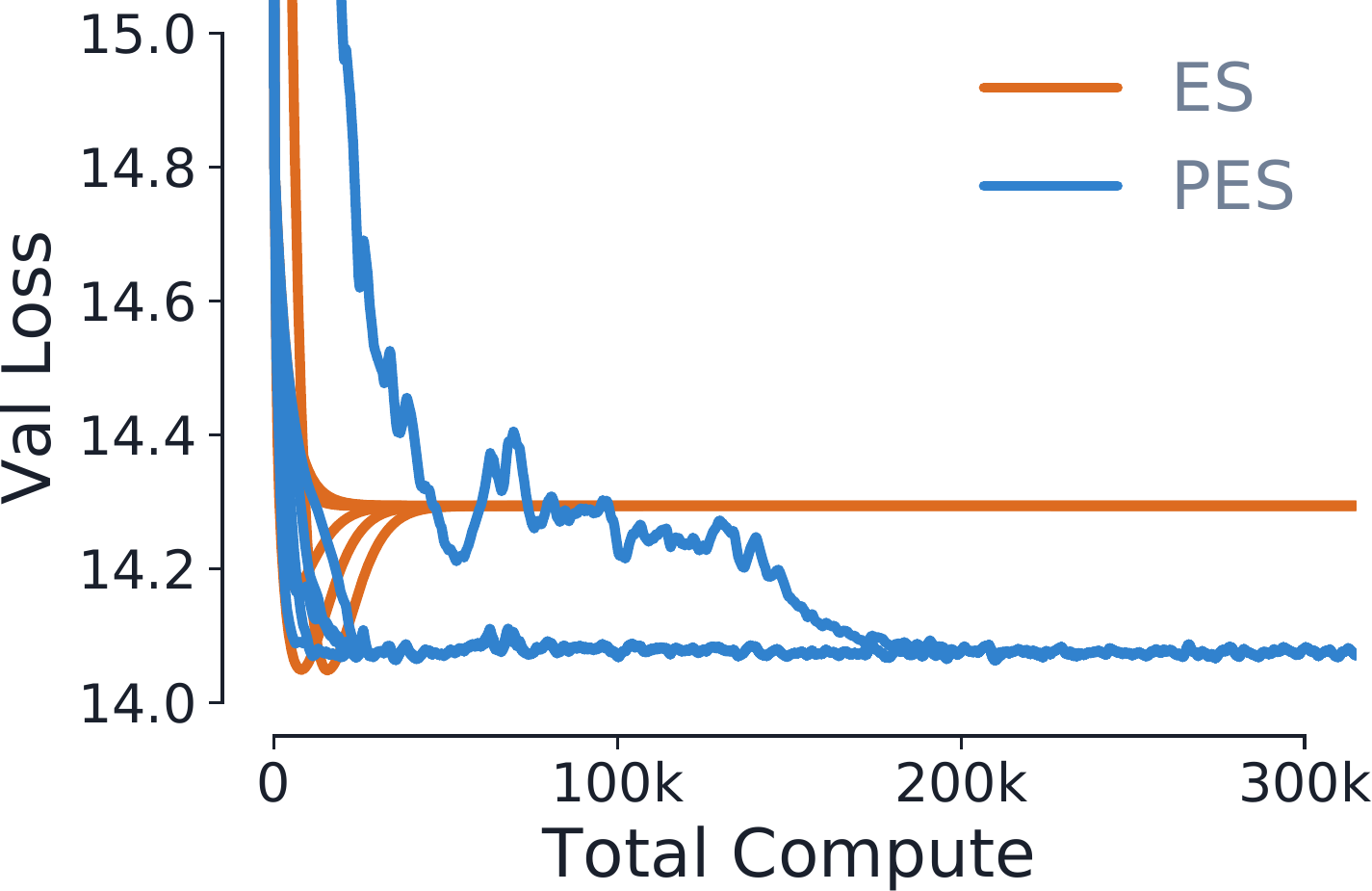}
        \caption{Validation losses using ES and PES for different initial $L_2$ values.}
        \label{fig:uci-val-loss}
    \end{subfigure}
    \caption{Using ES and PES to tune the $L_2$ regularization coefficient for linear regression on the UCI Yacht dataset.}
    \label{fig:uci}
\end{figure}
Here we show that truncation bias can also arise for regularization hyperparameters such as the $L_2$ regularization coefficient.
We tune $L_2$ regularization for linear regression on the Yacht data from the UCI collection~\cite{asuncion2007uci}.
We found the optimal $L_2$ coefficient using a fine-trained grid search.
In Figure~\ref{fig:uci} we compare meta-optimization using ES and PES, starting from different initial $L_2$ coefficients; PES robustly converges to the correct solution in all cases.
We used $\sigma=0.01$, $K=1$, and $N=4$ for both ES and PES.

\subsection{Continuous Control Details}
We used OpenAI Gym\footnote{\url{https://github.com/openai/gym}} to interface with MuJoCo.
In our implementation, each antithetic pair shares a MuJoCo environment state, which is different between different antithetic pairs.
The environment state is reset to the same point before running the partial unrolls of each particle in a pair, to control for randomness (e.g., the antithetic perturbations are evaluated starting from a common state).
As is standard for MuJoCo environments, the length of a full episode is $T=1000$; we ran full-unroll ES with $K=1000$, and we used partial unrolls of length $K=100$ for truncated ES and PES.
We used $10$ antithetic pairs for each of ES and PES.
Following~\citet{mania2018simple}, we used vanilla SGD to optimize the policy parameters.
For each of ES and PES, we performed a grid search over learning rates and perturbation scales, both from the set $\{1.0, 0.3, 0.1, 0.01\}$.
To evaluate policies, we computed the average full-episode reward over 50 random environment seeds.
In Figure~\ref{fig:mujoco}, we show the mean performance of each algorithm over 6 random seeds, with standard deviation shown by the shaded region.
Following~\citet{mania2018simple}, we used a linear policy initialized as all 0s (the linear policy is a single weight matrix with no bias term).
Also following~\citet{mania2018simple}, we divided the rewards by their standard deviation (computed using the aggregated rewards from all antithetic pairs) before computing the ES/PES gradient estimates.
We did not use state normalization, nor did we perform any heuristic selection of a subset of the best sampled perturbation directions (as used in the ARS V2 approach of~\citet{mania2018simple}).

\section{Telescoping Sums}
\label{app:telescoping}

If we wish to target the final loss $L_T$ as the meta-objective, we can define $p_t = L_t - L_{t-1}$, where $L_{-1} \equiv 0$.
This yields the telescoping sum:
\begin{align}
    \sum_{t=0}^T p_t &= (\cancel{L_0} - L_{-1}) + (\cancel{L_1} - \cancel{L_0}) + (\cancel{L_2} - \cancel{L_1}) + \cdots + (\cancel{L_{T-1}} - \cancel{L_{T-2}}) + (L_T - \cancel{L_{T-1}}) = L_T
\end{align}
\begin{wrapfigure}[13]{r}{0.35\linewidth}
    \vspace{-0.5cm}
    \centering
    \includegraphics[width=\linewidth]{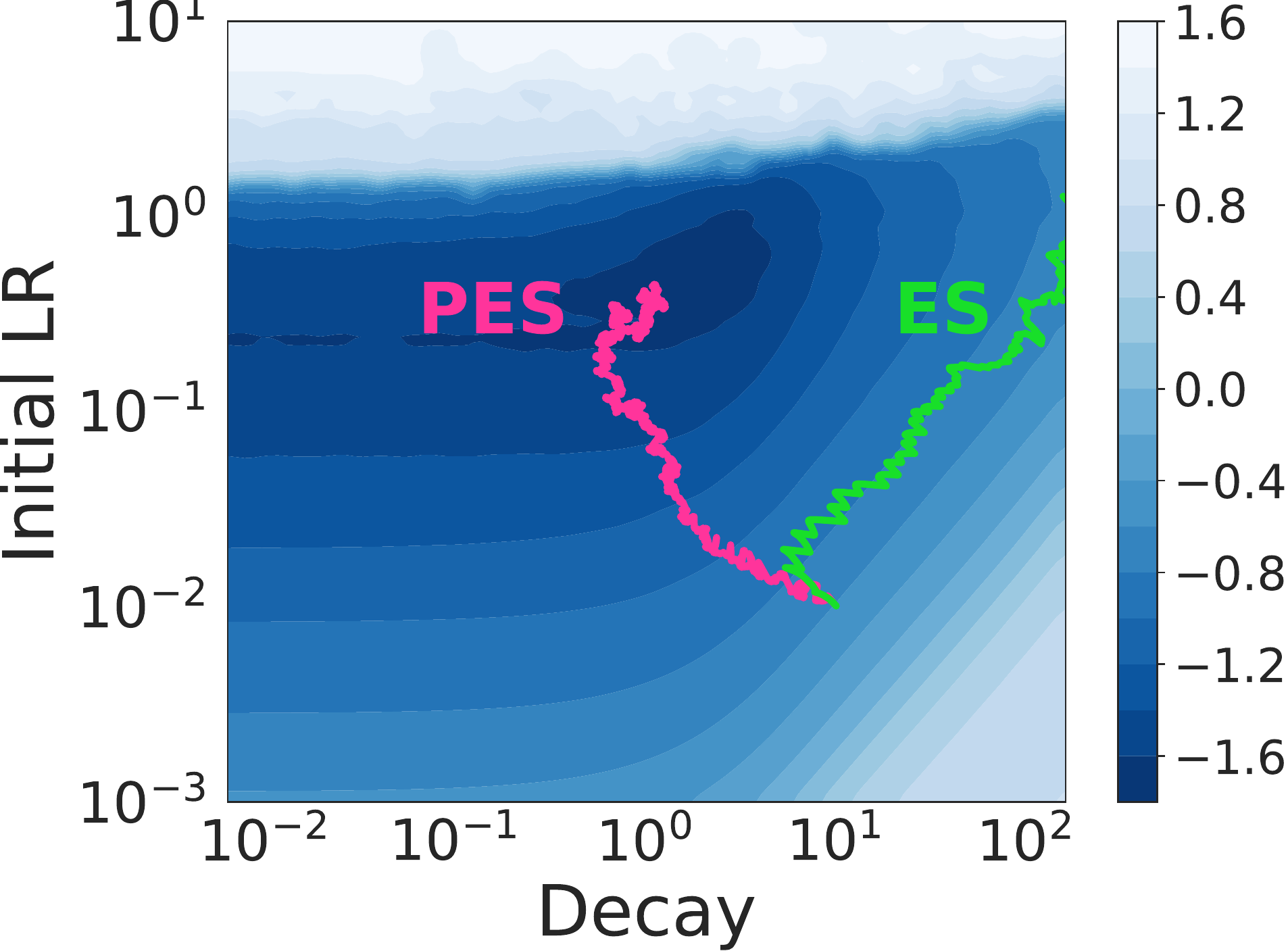}
    \caption{Telescoping sum for FashionMNIST final training loss (colors show log-final-loss).}
    \label{fig:telescoping-fashion-mnist}
\end{wrapfigure}
Targeting the final loss encourages different behavior than targeting the sum or average of the losses.
Targeting the sum of losses encourages fast convergence (small $\sum_t L_t$), but not necessarily the smallest final loss $L_T$, while targeting the final loss encourages finding the smallest $L_T$ potentially at the expense of slower convergence (larger $\sum_t L_t$).

We performed an experiment using telescoping sums to target the final training loss, optimizing an exponential LR schedule for an MLP on FashionMNIST with $T=5000$, $K=20$, $N=100$ (Figure~\ref{fig:telescoping-fashion-mnist}).
Due to the computational expense of evaluating the loss on the full training set to obtain $L_t$ at each partial unroll, we selected a random minibatch at the start of each inner problem, which was kept fixed for the loss evaluations for that inner problem.

\section{Derivation of Persistent Evolution Strategies}
\label{app:pes-derivation}

Here we derive the PES estimator. The derivation here closely follows that in the text body, but shows additional intermediate steps in several places in the derivation. Also see  Appendix~\ref{app:pes-stochastic-computation-graph} for an alternate derivation using stochastic computation graphs~\citep{schulman2015gradient}.

\subsection{Notation}

\begin{figure}[h]
    \centering
    \includegraphics[width=0.75\linewidth]{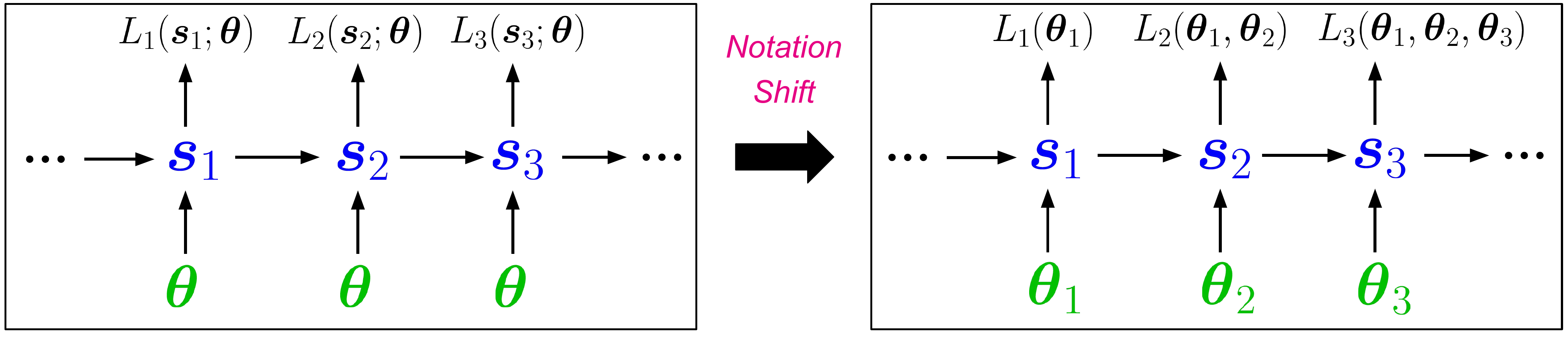}
    \caption{Shift in notation, dropping the dependence on $\bolds_t$ and explicitly including the dependence on each $\boldtheta_t$.}
    \label{fig:notation-shift}
\end{figure}
Unrolled computation graphs (as illustrated in Figure~\ref{fig:unrolled-graph}) depend on shared parameters $\boldtheta$ at every timestep.
In order to account for how these contribute to the overall gradient $\nabla_\boldtheta L(\boldtheta)$, we use subscripts $\boldtheta_t$ to distinguish between applications of $\boldtheta$ at different steps, where $\boldtheta_t = \boldtheta, \forall t$ (see Figure~\ref{fig:notation-shift}). 
We further define $\Theta = (\boldtheta_1, \dots, \boldtheta_T)^\top$, which is a matrix with the per-timestep $\boldtheta_t$ as its rows. 
For notational simplicity in the following derivation, we drop the dependence on $\bolds_t$ and explicitly include the dependence on each $\boldtheta_t$, writing $L_t(\bolds_t ; \boldtheta)$ as either $L_t(\boldtheta_1, \dots, \boldtheta_t)$ or simply $L_t(\Theta)$, with an implicit initial state $\bolds_0$.
Thus, $L(\boldtheta) = \sum_{t=1}^T L_t(\bolds_t ; \boldtheta) = \sum_{t=1}^T L_t(\boldtheta_1, \dots, \boldtheta_t) = \sum_{t=1}^T L_t(\Theta)$.

\subsection{PES is ES Over the Parameters at Each Unroll Step}

We wish to compute the gradient $\nabla_{\boldtheta} L(\boldtheta)$ of the total loss over all unrolls.
We begin by writing this gradient in terms of the full gradient $\pd{L(\Theta)}{\ovec{\Theta}} \in \mathbb{R}^{PT \times 1}$, where $P$ is the number of parameters, and $T$ is the total number of unrolls, and then using ES to approximate $\pd{L(\Theta)}{\ovec{\Theta}}$.
First, note that we can write:
\begin{align*}
\frac{d L(\boldtheta)}{d \boldtheta}
&=
\frac{d L(\Theta)}{d \boldtheta}
=
\frac{\partial L(\Theta)}{\partial \boldtheta_1} \cancelto{1}{\frac{d \boldtheta_1}{d \boldtheta}} + \frac{\partial L(\Theta)}{\partial \boldtheta_2} \cancelto{1}{\frac{d \boldtheta_2}{d \boldtheta}} + \cdots + \frac{\partial L(\Theta)}{\partial \boldtheta_T} \cancelto{1}{\frac{d \boldtheta_T}{d \boldtheta}}
=
\sum_{\tau=1}^T \pd{L\pp{\Theta}}{\boldtheta_\tau}
=
\pp{\mb I \otimes \boldone^\top} \pd{L(\Theta)}{\ovec{\Theta}}
\end{align*}
where $\otimes$ denotes the Kronecker product, $\mb{I}$ has dimension $P \times P$, $\boldone^\top$ has dimension $1 \times T$, and thus $\mb{I} \otimes \boldone^\top$ has dimension $P \times PT$.
Note that because $\pd{L(\Theta)}{\ovec{\Theta}}$ has dimension $PT \times 1$, the product $\pp{\mb I \otimes \boldone^\top} \pd{L(\Theta)}{\ovec{\Theta}}$ will be $P \times 1$.
Next, we will apply ES to approximate the last RHS expression above:
\begin{align*}
\frac{d L(\boldtheta)}{d \boldtheta} \approx
    \boldg^\text{PES} 
    &= \pp{\mb I \otimes \boldone^\top} \expect{\boldepsilon}{
        \frac{1}{\sigma^2} 
        \ovec{\boldepsilon} L\pp{\Theta + \boldepsilon}
        } \\    
    &= \frac{1}{\sigma^2} \expect{\boldepsilon}{
        \pp{\mb I \otimes \boldone^\top} \ovec{\boldepsilon}  L\pp{\Theta + \boldepsilon}
        }\\    
    &= \frac{1}{\sigma^2} \expect{\boldepsilon}{
        \left( \sum_{\tau=1}^T \boldepsilon_\tau \right) L\pp{\Theta + \boldepsilon}
        },
\end{align*}
where $\boldepsilon = \pp{\boldepsilon_1, \dots, \boldepsilon_T}^\top$ is a matrix of perturbations $\boldepsilon_t$ to be added to the $\boldtheta_t$ at each timestep and the expectation is over entries in $\boldepsilon$ drawn from an i.i.d. Gaussian with variance $\sigma^2$. 
This ES approximation is an unbiased estimator of the gradient of the Gaussian-smoothed objective $\mathbb{E}_{\boldepsilon}[ L(\Theta + \boldepsilon)]$.

We next show that $\boldg^\text{PES}$ decomposes into a sum of sequential gradient estimates,
\begin{align}
    \boldg^\text{PES} &= 
        \frac{1}{\sigma^2} \expect{\boldepsilon}{
        \left(\sum_{\tau=1}^T \boldepsilon_\tau \right) L\pp{\Theta + \boldepsilon}
        } \nonumber \\
        &=
        \frac{1}{\sigma^2} \expect{\boldepsilon}{
        \left( \sum_{\tau=1}^T \boldepsilon_\tau \right) \sum_{t=1}^T L_t\pp{\Theta + \boldepsilon}
        } \nonumber \\
        &= 
        \frac{1}{\sigma^2} \expect{\boldepsilon}{
        \sum_{t=1}^T \pp{\sum_{\tau=1}^T \boldepsilon_\tau}  L_t\pp{\Theta + \boldepsilon}
        }  \\
        &= 
        \frac{1}{\sigma^2} \expect{\boldepsilon}{
        \sum_{t=1}^T \pp{\sum_{\tau=1}^t \boldepsilon_\tau}  L_t\pp{\Theta + \boldepsilon}
        } \label{app:eq T to t} \\
        &= 
        \frac{1}{\sigma^2} \expect{\boldepsilon}{
        \sum_{t=1}^T \boldxi_t  L_t\pp{\Theta + \boldepsilon}
        }  \\
    &= \expect{\boldepsilon}{\sum_{t=1}^T \gpes_{t, \boldepsilon} }, \label{app:eq PES expectation} \\
    \gpes_{t, \boldepsilon} &= \frac{1}{\sigma^2} \boldxi_t L_t\pp{\Theta + \boldepsilon} = \frac{1}{\sigma^2} \boldxi_t L_t\pp{\boldtheta_1 + \boldepsilon_1, \dots, \boldtheta_t + \boldepsilon_t}. \label{app:eq pes step contribution}
\end{align}
where $\boldxi_t = \sum_{\tau=1}^t \boldepsilon_\tau$, Equation \ref{app:eq T to t} relies on $L_t\pp{\cdot}$ being independent of $\boldepsilon_\tau$ for $\tau > t$, and Equation \ref{app:eq pes step contribution} similarly relies on $L_t\pp{\cdot}$ only being a function of $\boldtheta_\tau$ for $\tau \leq t$.

The PES estimator consists of Monte Carlo estimates of Equation \ref{app:eq PES expectation},
\begin{align}
    \gpes &= \frac{1}{N} \sum_{i=1}^N \sum_{t=1}^T \gpes_{t, \boldepsilon^{(i)}}    
    \label{eq:pes-appendix}
\end{align} 
where $\boldepsilon^{(i)}$ are samples of $\boldepsilon$, and $N$ is the number of Monte Carlo samples. Gradient estimates at each time step can be evaluated sequentially, and used to perform SGD.

\paragraph{Concrete Example.}
To illustrate how the expressions in the derivation above yield the desired gradient estimate, here we provide a concrete example using two-dimensional $\boldtheta$ with three steps of unrolling.
The matrix $\Theta$ is:
$$
\Theta =
\begin{bmatrix}
 \longdash \boldtheta_1^\top \longdash \\
 \longdash \boldtheta_2^\top \longdash \\
 \longdash \boldtheta_3^\top \longdash
\end{bmatrix}
=
\begin{bmatrix}
\theta_1^{(1)} & \theta_1^{(2)} \\
\theta_2^{(1)} & \theta_2^{(2)} \\
\theta_3^{(1)} & \theta_3^{(2)}
\end{bmatrix}
$$
The vectorized matrix $\ovec{\Theta}$ and gradient $\frac{\partial L(\Theta)}{\partial \ovec{\Theta}}$ are as follows:
$$
\text{vec}(\Theta) = 
\begin{bmatrix}
\theta_1^{(1)} \\[0.5em]
\theta_2^{(1)} \\[0.5em]
\theta_3^{(1)} \\[0.5em]
\theta_1^{(2)} \\[0.5em]
\theta_2^{(2)} \\[0.5em]
\theta_3^{(2)}
\end{bmatrix}
\qquad \qquad
\frac{\partial L(\Theta)}{\partial \text{vec}(\Theta)} =
\begin{bmatrix}
 \frac{\partial L(\Theta)}{\partial \theta_1^{(1)}} \\[1em]
 \frac{\partial L(\Theta)}{\partial \theta_2^{(1)}} \\[1em]
 \frac{\partial L(\Theta)}{\partial \theta_3^{(1)}} \\[1em]
 \frac{\partial L(\Theta)}{\partial \theta_1^{(2)}} \\[1em]
 \frac{\partial L(\Theta)}{\partial \theta_2^{(2)}} \\[1em]
 \frac{\partial L(\Theta)}{\partial \theta_3^{(2)}}
\end{bmatrix}
$$
The Kronecker product is:
$
\mb{I} \otimes \boldone^\top =
\begin{bmatrix}
 1 & 0 \\
 0 & 1
\end{bmatrix}
\otimes
\begin{bmatrix}
 1 & 1 & 1
\end{bmatrix}
=
\begin{bmatrix}
  1 & 1 & 1 & 0 & 0 & 0 \\
  0 & 0 & 0 & 1 & 1 & 1
\end{bmatrix}
$.
Thus, we have:
$$
(\mb{I} \otimes \boldone^\top) \frac{\partial L(\Theta)}{\partial \text{vec}(\Theta)}
=
\begin{bmatrix}
  \frac{\partial L(\Theta)}{\partial \theta_1^{(1)}} + \frac{\partial L(\Theta)}{\partial \theta_2^{(1)}} + \frac{\partial L(\Theta)}{\partial \theta_3^{(1)}} \\[1.5em]
  \frac{\partial L(\Theta)}{\partial \theta_1^{(2)}} + \frac{\partial L(\Theta)}{\partial \theta_2^{(2)}} + \frac{\partial L(\Theta)}{\partial \theta_3^{(2)}}
\end{bmatrix}
=
\underbrace{
\begin{bmatrix}
  \frac{\partial L(\Theta)}{\partial \theta_1^{(1)}} \\[1.5em]
  \frac{\partial L(\Theta)}{\partial \theta_1^{(2)}}
\end{bmatrix}
}_{\frac{\partial L(\Theta)}{\partial \boldtheta_1}}
+
\underbrace{
\begin{bmatrix}
  \frac{\partial L(\Theta)}{\partial \theta_2^{(1)}} \\[1.5em]
  \frac{\partial L(\Theta)}{\partial \theta_2^{(2)}}
\end{bmatrix}
}_{\frac{\partial L(\Theta)}{\partial \boldtheta_2}}
+
\underbrace{
\begin{bmatrix}
  \frac{\partial L(\Theta)}{\partial \theta_3^{(1)}} \\[1.5em]
  \frac{\partial L(\Theta)}{\partial \theta_3^{(2)}}
\end{bmatrix}
}_{\frac{\partial L(\Theta)}{\partial \boldtheta_3}}
=
\sum_{\tau=1}^T \frac{\partial L(\Theta)}{\partial \boldtheta_\tau}
=
\frac{d L(\Theta)}{d \boldtheta}
$$
Similarly, to see how the PES derivation works, consider a matrix of perturbations $\boldepsilon$ and its vectorization $\text{vec}(\boldepsilon)$ as follows:
$$
\boldepsilon
=
\begin{bmatrix}
  \longdash \boldepsilon_1^\top \longdash \\
  \longdash \boldepsilon_2^\top \longdash \\
  \longdash \boldepsilon_3^\top \longdash
\end{bmatrix}
=
\begin{bmatrix}
  \epsilon_1^{(1)} & \epsilon_1^{(2)} \\
  \epsilon_2^{(1)} & \epsilon_2^{(2)} \\
  \epsilon_3^{(1)} & \epsilon_3^{(2)}
\end{bmatrix}
\qquad \qquad
\text{vec}(\boldepsilon)
=
\begin{bmatrix}
  \epsilon_1^{(1)} \\
  \epsilon_2^{(1)} \\
  \epsilon_3^{(1)} \\
  \epsilon_1^{(2)} \\
  \epsilon_2^{(2)} \\
  \epsilon_3^{(2)}
\end{bmatrix}
$$
Then,
$$
(\mb{I} \otimes \boldone^\top) \text{vec}(\boldepsilon)
=
\begin{bmatrix}
  \epsilon_1^{(1)} + \epsilon_2^{(1)} + \epsilon_3^{(1)} \\
  \epsilon_1^{(2)} + \epsilon_2^{(2)} + \epsilon_3^{(2)}
\end{bmatrix}
=
\underbrace{
\begin{bmatrix}
  \epsilon_1^{(1)} \\
  \epsilon_1^{(2)}
\end{bmatrix}
}_{\boldepsilon_1}
+
\underbrace{
\begin{bmatrix}
  \epsilon_2^{(1)} \\
  \epsilon_2^{(2)}
\end{bmatrix}
}_{\boldepsilon_2}
+
\underbrace{
\begin{bmatrix}
  \epsilon_3^{(1)} \\
  \epsilon_3^{(2)}
\end{bmatrix}
}_{\boldepsilon_3}
=
\sum_{\tau=1}^T \boldepsilon_\tau
$$
This shows how the following statements are equivalent in our derivation:
$$
\frac{1}{\sigma^2} \expect{\boldepsilon}{
        \pp{\mb I \otimes \boldone^\top} \ovec{\boldepsilon}  L\pp{\Theta + \boldepsilon}
        }
= \frac{1}{\sigma^2} \expect{\boldepsilon}{
    \left( \sum_{\tau=1}^T \boldepsilon_\tau \right) L\pp{\Theta + \boldepsilon}
    }
$$

\section{Proof that PES is Unbiased}
\label{app:pes-bias}

\begin{statement}
    Let $\boldtheta \in \mathbb{R}^n$ and $L(\boldtheta) = \sum_{t=1}^T L_t(\boldtheta)$.
    Suppose that $\nabla_{\boldtheta} L(\boldtheta)$ exists, and assume that $L$ is quadratic, so that it is equivalent to its second-order Taylor series expansion:
    $$
    L(\Theta + \boldepsilon) = L(\Theta) + \veps^\top \nabla_{\vTheta} L(\Theta) + \frac{1}{2} \veps^\top \nabla^2_{\vTheta} L(\Theta) \veps
    $$
    Consider the PES estimator (using antithetic sampling) below:
    $$
    \gpesanti
    =
    (\mb{I} \otimes \boldone^\top) \mathbb{E}_{\boldepsilon} \left[ \frac{1}{2\sigma^2} \veps (L(\Theta + \boldepsilon) - L(\Theta - \boldepsilon)) \right],
    $$
    where $\epsilon \sim \mathcal{N}(0, I\sigma^2)$.
    Then, $\text{\emph{bias}}(\gpesanti) = \mathbb{E}_{\boldepsilon}[\gpesanti] - \nabla_{\boldtheta} L(\boldtheta) = \boldzero$.
\end{statement}
\begin{proof}
Using the assumption that $L$ is quadratic and due to antithetic sampling, we can simplify this expression $L(\Theta + \boldepsilon) - L(\Theta - \boldepsilon)$ as follows:
\begin{align}
    L(\Theta + \boldepsilon) - L(\Theta - \boldepsilon) =& (\cancel{L(\Theta)} + \veps^\top \nabla_{\vTheta} L(\Theta) + \cancel{\frac{1}{2} \veps^\top \nabla^2_{\vTheta} L(\Theta) \veps)} \\
    &- (\cancel{L(\Theta)} - \veps^\top \nabla_{\vTheta} L(\Theta) + \cancel{\frac{1}{2} \veps^\top \nabla^2_{\vTheta} L(\Theta) \veps)} \\
    =&
    2 \veps^\top \nabla_{\vTheta} L(\Theta)
\end{align}
Thus, we have:
\begin{align}
    \gpesanti
    &= (\mb{I} \otimes \boldone^\top) \mathbb{E}_{\boldepsilon} \left[ \frac{1}{2\sigma^2} 2 \veps \veps^\top \nabla_{\vTheta} L(\Theta) \right] \\
    &= (\mb{I} \otimes \boldone^\top) \frac{1}{\sigma^2} \underbrace{\mathbb{E}_{\boldepsilon} \left[ \veps \veps^\top \right]}_{\sigma^2 I} \nabla_{\vTheta} L(\Theta) \\
    &= (\mb{I} \otimes \boldone^\top) \nabla_{\vTheta} L(\Theta) \\
    &= \nabla_{\boldtheta} L(\boldtheta)
\end{align}

Thus, $\mathbb{E}[\gpesanti] = \nabla_{\boldtheta} L(\boldtheta)$ and $\text{bias}(\gpesanti) = \boldzero$.
\end{proof}

\section{PES Variance}
\label{app:pes-variance}

In this section, we derive the variance of PES.
The antithetic PES estimator assuming quadratic $L$ is as follows:
\begin{align}
    \gpesanti
    &=
    \frac{1}{\sigma^2} \sum_{t=1}^T \boldxi_{t} \ovec{\boldepsilon_{1\dots t}}^\top \nabla_{\ovec{\Theta_{1\dots t}}} L_t(\Theta)
\end{align}
For simplicity in the following derivation, we consider a Monte-Carlo estimate using a single particle pair.
For $N$ particles, the variance will be scaled by a factor of $1 \over N$.
We use the total variance $\tr(\Var(\gpes))$ to quantify the variance of the estimator:
\begin{align}
\tr(\Var(\gpesanti))
&= \tr(\mathbb{E}[\gpesanti \gpesanti^\top] - \mathbb{E}[\gpesanti] {\mathbb{E}[\gpesanti]}^\top), \\
&= \underbrace{\mathbb{E}[\gpesanti^\top \gpesanti]}_{\num{1}} - \underbrace{{\mathbb{E}[\gpesanti]}^\top \mathbb{E}[\gpesanti]}_{\num{2}} \label{eq:total-variance}
\end{align}
Term $\num{2}$ is easy to compute, because the estimator is unbiased, so $\mathbb{E}_{\boldepsilon}[\gpesanti] = \nabla_{\boldtheta} L(\Theta)$.
Thus,
\begin{align}
    \num{2}
    =
    {\mathbb{E}[\gpesanti]}^\top \mathbb{E}[\gpesanti]
    =
    \nabla_{\boldtheta} L(\Theta)^\top \nabla_{\boldtheta} L(\Theta)
    =
    ||\nabla_{\boldtheta} L(\Theta)||^2
\end{align}
To derive term $\num{1}$, we will expand out $\gpesanti^\top \gpesanti$ into a sum of simple sub-expressions and use the linearity of expectation to combine them.
To simplify notation, we use the shorthand $\boldv_t \equiv \ovec{\boldepsilon_{1 \dots t}}$ and $\boldg_t \equiv \nabla_{\ovec{\Theta_{1 \dots t}}} L_t(\Theta)$.
First, note that:
\begin{align}
    \gpesanti^\top \gpesanti
    &=
    \frac{1}{\sigma^4} \pp{\sum_{t=1}^T \boldxi_t \underbrace{\ovec{\boldepsilon_{1 \dots t}}}_{\boldv_t} \underbrace{\nabla_{\ovec{\Theta_{1 \dots t}}} L_t(\Theta)}_{\boldg_t}}^\top \pp{\sum_{t=1}^T \boldxi_t \underbrace{\ovec{\boldepsilon_{1 \dots t}}}_{\boldv_t} \underbrace{\nabla_{\ovec{\Theta_{1 \dots t}}} L_t(\Theta)}_{\boldg_t}}  \label{eq:no-shorthand} \\
    &=
    \frac{1}{\sigma^4} \pp{\boldxi_1 \boldv_1^\top \boldg_1 + \cdots + \boldxi_T \boldv_T^\top \boldg_T}^\top \pp{\boldxi_1 \boldv_1^\top \boldg_1 + \cdots + \boldxi_T \boldv_T^\top \boldg_T} \label{eq:shorthand} \\
    &=
    \frac{1}{\sigma^4} \pp{\underbrace{\boldg_1^\top \boldv_1 \boldxi_1^\top \boldxi_1 \boldv_1^\top \boldg_1}_{\num{a}} + \underbrace{\boldg_1^\top \boldv_1 \boldxi_1^\top \boldxi_2 \boldv_2^\top \boldg_2}_{\num{b}} + \cdots + \boldg_T^\top \boldv_T \boldxi_T^\top \boldxi_T \boldv_T^\top \boldg_T} \label{eq:initial-expansion}
\end{align}

There are two types of terms in Eq.~\ref{eq:initial-expansion}: terms of type $\num{a}$, which have the form $\boldg_i^\top \boldv_i \boldxi_i^\top \boldxi_i \boldv_i^\top \boldg_i$, and terms of type $\num{b}$, which have the form $\boldg_i^\top \boldv_i \boldxi_i^\top \boldxi_j \boldv_j^\top \boldg_j$ where $i \neq j$.
We will derive the expectations of each of these two types of terms separately, and then combine the resulting sub-expressions.

\paragraph{Terms of Type $\num{a}$.}
As the first step in expanding out each term of type $\num{a}$, note that:
\begin{align}
    \boldv_t^\top \boldg_t
    =
    \ovec{\boldepsilon_{1 \dots t}}^\top \nabla_{\ovec{\Theta_{1 \dots t}}} L_t(\Theta)
    =
    \sum_{\tau=1}^t \boldepsilon_\tau^\top \nabla_{\boldtheta_\tau} L_t(\Theta)
\end{align}
Also, note that $\boldxi_i^\top \boldxi_i$ can be expanded as follows:
\begin{align}
    \boldxi_i^\top \boldxi_i
    =
    \pp{\boldepsilon_1 + \cdots + \boldepsilon_i}^\top \pp{\boldepsilon_1 + \cdots + \boldepsilon_i}
    =
    \sum_{m=1}^i \boldepsilon_m^\top \boldepsilon_m + \sum_{m \leq i, n \leq i, m \neq n} \boldepsilon_m^\top \boldepsilon_n
\end{align}

Thus, we have:
\begin{align}
    \boldg_i^\top \boldv_i \boldxi_i^\top \boldxi_i \boldv_i^\top \boldg_i
    &=
    \underbrace{\pp{\sum_{m=1}^i \boldepsilon_m^\top \nabla_{\boldtheta_m} L_i(\Theta)}^\top \pp{\sum_{n=1}^i \boldepsilon_n^\top \boldepsilon_n} \pp{\sum_{m=1}^i \boldepsilon_m^\top \nabla_{\boldtheta_m} L_i(\Theta)}}_{\num{I}} \\
    & \quad + \underbrace{\pp{\sum_{m=1}^i \boldepsilon_m^\top \nabla_{\boldtheta_m} L_i(\Theta)}^\top \pp{\sum_{m \leq i, n \leq i, m \neq n} \boldepsilon_m^\top \boldepsilon_n} \pp{\sum_{m=1}^i \boldepsilon_m^\top \nabla_{\boldtheta_m} L_i(\Theta)}}_{\num{II}}
\end{align}

We see that there are two types of terms in $\num{I}$ with non-zero expectation:
\begin{align}
    \expect{\boldepsilon}{\nabla_{\boldtheta_m} L_i(\Theta)^\top \boldepsilon_m \boldepsilon_m^\top \boldepsilon_m \boldepsilon_m^\top \nabla_{\boldtheta_m} L_i(\Theta)}
    &=
    \nabla_{\boldtheta_m} L_i(\Theta)^\top \underbrace{\expect{\boldepsilon}{\boldepsilon_m \boldepsilon_m^\top \boldepsilon_m \boldepsilon_m^\top}}_{(P+2)\sigma^4 I} \nabla_{\boldtheta_m} L_i(\Theta)  \label{eq:four-repeated-epsilons} \\
    &=
    (P+2)\sigma^4 \norm{\nabla_{\boldtheta_m} L_i(\Theta)}^2
\end{align}

To compute $\expect{\boldepsilon}{\boldepsilon_m \boldepsilon_m^\top \boldepsilon_m \boldepsilon_m^\top}$ in Eq.~\ref{eq:four-repeated-epsilons}, we used the following identity, which was derived in Appendix A.2 of~\cite{maheswaranathan2018guided}.
The $\boldepsilon_m$ are assumed to be drawn from $\mathcal{N}(\boldzero, \Sigma)$, where in our case $\Sigma = \sigma^2 \mb{I}$:
\begin{align}
    \expect{\boldepsilon}{\boldepsilon_m \boldepsilon_m^\top \boldepsilon_m \boldepsilon_m^\top}
    &=
    \tr(\Sigma) \Sigma + 2 \Sigma^2 \\
    &=
    \tr(\sigma^2 \mb{I}) (\sigma^2 \mb{I}) + 2 (\sigma^2 \mb{I})^2 \\
    &=
    P \sigma^4 \mb{I} + 2 \sigma^4 I \\
    &=
    (P + 2) \sigma^4 \mb{I}
\end{align}

There are $i$ terms of this type, that make the following contribution to $\num{I}$:
\begin{align}
    (P+2) \sigma^4 \sum_{m=1}^i \norm{\nabla_{\boldtheta_m} L_i(\Theta)}^2
\end{align}

The second type of term in $\num{I}$ with non-zero expectation has the form $\nabla_{\boldtheta_m} L_i(\Theta)^\top \boldepsilon_m \boldepsilon_n^\top \boldepsilon_n \boldepsilon_m^\top \nabla_{\boldtheta_m} L_i(\Theta)$ where $m \neq n$.
Computing the expectation, we have:
\begin{align}
    \expect{\boldepsilon}{\nabla_{\boldtheta_m} L_i(\Theta)^\top \boldepsilon_m \boldepsilon_n^\top \boldepsilon_n \boldepsilon_m^\top \nabla_{\boldtheta_m} L_i(\Theta)}
    &=
    \nabla_{\boldtheta_m} L_i(\Theta)^\top \underbrace{\expect{\boldepsilon}{\boldepsilon_m \boldepsilon_n^\top \boldepsilon_n \boldepsilon_m^\top}}_{P \sigma^4 I} \nabla_{\boldtheta_m} L_i(\Theta)   \label{eq:two-repeated-epsilons}
\end{align}
Where $\expect{\boldepsilon}{\boldepsilon_m \boldepsilon_n^\top \boldepsilon_n \boldepsilon_m^\top}$ in Eq.~\ref{eq:two-repeated-epsilons} is computed as follows:
\begin{align}
    \expect{\boldepsilon_m, \boldepsilon_n}{\boldepsilon_m \boldepsilon_n^\top \boldepsilon_n \boldepsilon_m^\top}
    &=
    \expect{\boldepsilon_m}{\expect{\boldepsilon_n}{\boldepsilon_m \boldepsilon_n^\top \boldepsilon_n \boldepsilon_m^\top}}
    =
    \mathbb{E}_{\boldepsilon_m}[\boldepsilon_m \underbrace{\expect{\boldepsilon_n}{\boldepsilon_n^\top \boldepsilon_n}}_{P \sigma^2} \boldepsilon_m^\top]
    =
    P \sigma^2 \underbrace{\mathbb{E}_{\boldepsilon_n}[\boldepsilon_n \boldepsilon_n^\top]}_{\sigma^2 \mb{I}}
    =
    P \sigma^4 \mb{I}  \label{eq:two-epsilon-pair-derivation}
\end{align}
In Eq.~\ref{eq:two-epsilon-pair-derivation}, we obtained $\expect{\boldepsilon_m}{\boldepsilon_m^\top \boldepsilon_m} = P \sigma^2$ via:
\begin{align}
    \expect{\boldepsilon_m}{\boldepsilon_m^\top \boldepsilon_m}
    =
    \expect{\boldepsilon_m}{\tr(\boldepsilon_m^\top \boldepsilon_m)}
    =
    \expect{\boldepsilon_m}{\tr(\boldepsilon_m \boldepsilon_m^\top)}
    =
    \tr \pp{\expect{\boldepsilon_m}{\boldepsilon_m \boldepsilon_m^\top}}
    =
    \tr(\sigma^2 \mb{I})
    =
    P \sigma^2
\end{align}

The total contribution of terms of this type is:
\begin{align}
    \sum_{m=1}^i \sum_{n, m \in \{1, \dots, i\}, n \neq m} \nabla_{\boldtheta_m} L_i(\Theta)^\top \expect{\boldepsilon}{\boldepsilon_m \boldepsilon_n^\top \boldepsilon_n \boldepsilon_m^\top} \nabla_{\boldtheta_m} L_i(\Theta)
    &=
    \sum_{m=1}^i \sum_{n,m \in \{1, \dots, i\}, n \neq m} P \sigma^4 \norm{\nabla_{\boldtheta_m} L_i(\Theta)}^2 \\
    &=
    (i-1) P \sigma^4 \sum_{m=1}^i \norm{\nabla_{\boldtheta_m} L_i(\Theta)}^2
\end{align}

So far, we have:
\begin{align}
    \num{I}
    &=
    (P+2) \sigma^4 \sum_{m=1}^i \norm{\nabla_{\boldtheta_m} L_i(\Theta)}^2 + (i-1) P \sigma^4 \sum_{m=1}^i \norm{\nabla_{\boldtheta_m} L_i(\Theta)}^2 \\
    &=
    (i P + 2) \sigma^4 \sum_{m=1}^i \norm{\nabla_{\boldtheta_m} L_i(\Theta)}^2
\end{align}

Next, we need to compute:
$$
\num{II}
=
\pp{\sum_{m=1}^i \boldepsilon_m^\top \nabla_{\boldtheta_m} L_i(\Theta)}^\top \pp{\sum_{m \leq i, n \leq i, m \neq n} \boldepsilon_m^\top \boldepsilon_n} \pp{\sum_{m=1}^i \boldepsilon_m^\top \nabla_{\boldtheta_m} L_i(\Theta)}
$$
Here, the terms with nonzero expectation have the form $\nabla_{\boldtheta_m} L_i(\Theta)^\top \boldepsilon_m \boldepsilon_m^\top \boldepsilon_n \boldepsilon_n^\top \nabla_{\boldtheta_n} L_i(\Theta)$ or $\nabla_{\boldtheta_n} L_i(\Theta)^\top \boldepsilon_n \boldepsilon_m^\top \boldepsilon_n \boldepsilon_m^\top \nabla_{\boldtheta_m} L_i(\Theta)$, both of which have expectation:
\begin{align}
    \expect{\boldepsilon}{\nabla_{\boldtheta_m} L_i(\Theta)^\top \boldepsilon_m \boldepsilon_m^\top \boldepsilon_n \boldepsilon_n^\top \nabla_{\boldtheta_n} L_i(\Theta)}
    &=
    \nabla_{\boldtheta_m} L_i(\Theta)^\top \underbrace{\expect{\boldepsilon}{\boldepsilon_m \boldepsilon_m^\top \boldepsilon_n \boldepsilon_n^\top}}_{\sigma^4 I} \nabla_{\boldtheta_n} L_i(\Theta) \\
    &=
    \sigma^4 \nabla_{\boldtheta_n} L_i(\Theta)^\top \nabla_{\boldtheta_m} L_i(\Theta)
\end{align}

We have the following contribution from terms of this type, where the factor of 2 accounts for the two conditions ($\boldepsilon_n \boldepsilon_m^\top \boldepsilon_n \boldepsilon_m^\top$ and $\boldepsilon_m \boldepsilon_m^\top \boldepsilon_n \boldepsilon_n^\top$):
\begin{align}
    2 \sigma^4 \sum_{m \neq n} \nabla_{\boldtheta_m} L_i(\Theta)^\top \nabla_{\boldtheta_n} L_i(\Theta)
\label{eq needs 2}
\end{align}

Then, we have:
\begin{align}
    \expect{\boldepsilon}{\boldg_i^\top \boldv_i \boldxi_i^\top \boldxi_i \boldv_i^\top \boldg_i}
    =
    (iP + 2) \sigma^4 \sum_{m=1}^i \norm{\nabla_{\boldtheta_m} L_i(\Theta)}^2 + 2 \sigma^4 \sum_{m \leq i, n \leq i, m \neq n} \nabla_{\boldtheta_m} L_i(\Theta)^\top \nabla_{\boldtheta_n} L_i(\Theta)
\end{align}

\paragraph{Terms of Type $\num{b}$.}
Next, we consider terms of type $\num{b}$, which have the form $\boldg_i^\top \boldv_i \boldxi_i^\top \boldxi_j \boldv_j^\top \boldg_j$ where $i \neq j$.
Note that we can expand $\boldxi_i^\top \boldxi_j$ as follows:
\begin{align}
    \boldxi_i^\top \boldxi_j = (\boldepsilon_1 + \cdots + \boldepsilon_i)^\top (\boldepsilon_1 + \cdots + \boldepsilon_j)
    =
    \sum_{m=1}^r \boldepsilon_m^\top \boldepsilon_m + \sum_{m \in \{1, \dots, i \}, n \in \{1, \dots, j \}, m \neq n} \boldepsilon_m^\top \boldepsilon_n
\end{align}
where we define $r = \min(i, j)$.
Plugging in this expansion for $\boldxi_i^\top \boldxi_j$, we have:
\begin{align}
    \boldg_i^\top \boldv_i \boldxi_i^\top \boldxi_j \boldv_j^\top \boldg_j
    &=
    \underbrace{\pp{\sum_{m=1}^i \boldepsilon_m^\top \nabla_{\boldepsilon_m} L_i(\Theta)}^\top \pp{\sum_{m=1}^r \boldepsilon_m^\top \boldepsilon_m} \pp{\sum_{n=1}^j \boldepsilon_n^\top \nabla_{\boldepsilon_n} L_j(\Theta)}}_{\num{I}} \\
    & \quad + \underbrace{\pp{\sum_{m=1}^i \boldepsilon_m^\top \nabla_{\boldtheta_m} L_i(\Theta)}^\top \pp{\sum_{m \leq i, n \leq j, m \neq n} \boldepsilon_m^\top \boldepsilon_n} \pp{\sum_{n=1}^j \boldepsilon_n^\top \nabla_{\boldepsilon_n} L_j(\Theta)}}_{\num{II}}
\end{align}

Expanding $\num{I}$, there are two types of terms of interest: ones of the form $\nabla_{\boldtheta_m} L_i(\Theta)^\top \boldepsilon_m \boldepsilon_m^\top \boldepsilon_m \boldepsilon_m^\top \nabla_{\boldtheta_m} L_j(\Theta)$, and ones of the form $\nabla_{\boldtheta_n} L_i(\Theta)^\top \boldepsilon_n \boldepsilon_m^\top \boldepsilon_m \boldepsilon_n^\top \nabla_{\boldtheta_n} L_j(\Theta)$.
The expectation of the first type of term is:
\begin{align}
    \expect{\boldepsilon}{\nabla_{\boldtheta_m} L_i(\Theta)^\top \boldepsilon_m \boldepsilon_m^\top \boldepsilon_m \boldepsilon_m^\top \nabla_{\boldtheta_m} L_j(\Theta)}
    &=
    \nabla_{\boldtheta_m} L_i(\Theta)^\top \underbrace{\expect{\boldepsilon}{\boldepsilon_m \boldepsilon_m^\top \boldepsilon_m \boldepsilon_m^\top}}_{(P+2) \sigma^4 I} \nabla_{\boldtheta_m} L_j(\Theta) \\
    &=
    (P+2) \sigma^4 \nabla_{\boldtheta_m} L_i(\Theta)^\top \nabla_{\boldtheta_m} L_j(\Theta)
\end{align}
The total contribution from terms like this is:
\begin{align}
    (P+2) \sigma^4 \sum_{m=1}^r \nabla_{\boldtheta_m} L_i(\Theta)^\top \nabla_{\boldtheta_m} L_j(\Theta)
\end{align}
The expectation of the second type of term is:
\begin{align}
    \nabla_{\boldtheta_n} L_i(\Theta)^\top \expect{\boldepsilon}{\boldepsilon_n \boldepsilon_m^\top \boldepsilon_m \boldepsilon_n^\top} \nabla_{\boldtheta_n} L_j(\Theta)
    &=
    \sigma^4 \nabla_{\boldtheta_n} L_i(\Theta)^\top \nabla_{\boldtheta_n} L_j(\Theta)
\end{align}
The total contribution from terms of this type is:
\begin{align}
    \sigma^4 \sum_{m=1}^r \sum_{m \leq r, n \leq r, n \neq m} \nabla_{\boldtheta_n} L_i(\Theta)^\top \nabla_{\boldtheta_n} L_j(\Theta)
\end{align}

Next, we look at terms in expression $\num{II}$.
We have two types of terms that have nonzero expectation:
\begin{align}
    \expect{\boldepsilon}{\nabla_{\boldtheta_m} L_i(\Theta)^\top \boldepsilon_m \boldepsilon_m^\top \boldepsilon_n \boldepsilon_n^\top \nabla_{\boldtheta_n} L_j(\Theta)}
    =
    \sigma^4 \nabla_{\boldtheta_m} L_i(\Theta)^\top \nabla_{\boldtheta_n} L_j(\Theta)
\end{align}
and:
\begin{align}
    \expect{\boldepsilon}{\nabla_{\boldtheta_n} L_i(\Theta)^\top \boldepsilon_n \boldepsilon_m^\top \boldepsilon_n \boldepsilon_m^\top \nabla_{\boldtheta_m} L_j(\Theta)}
    =
    \sigma^4 \nabla_{\boldtheta_n} L_i(\Theta)^\top \nabla_{\boldtheta_m} L_j(\Theta)
\end{align}
The contribution from these terms is:
\begin{align}
    2 \sigma^4 \sum_{m \leq i, n \leq j, m \neq n} \nabla_{\boldtheta_m} L_i(\Theta)^\top \nabla_{\boldtheta_n} L_j(\Theta)
\end{align}

Thus, we have:
\begin{align}
    \expect{\boldepsilon}{\boldg_i^\top \boldv_i \boldxi_i^\top \boldxi_j \boldv_j^\top \boldg_j}
    &=
    (P + 2) \sigma^4 \sum_{m=1}^r \nabla_{\boldtheta_m} L_i(\Theta)^\top \nabla_{\boldtheta_m} L_j(\Theta)
    +
    \sigma^4 \sum_{m=1}^r \sum_{m,n \leq r, n \neq m} \nabla_{\boldtheta_n} L_i(\Theta)^\top \nabla_{\boldtheta_n} L_j(\Theta) \\
    & \quad +
    2 \sigma^4 \sum_{m \leq i, n \leq j, m \neq n} \nabla_{\boldtheta_m} L_i(\Theta)^\top \nabla_{\boldtheta_n} L_j(\Theta)
\end{align}
\paragraph{Combining Terms for $\expect{\boldepsilon}{\gpesanti^\top \gpesanti}$.}
Putting these components together, we have the following overall expression:
\begin{align}
    \expect{\boldepsilon}{\gpesanti^\top \gpesanti}
    &=
    \sum_{i=1}^T \pp{(iP + 2) \sum_{m=1}^i \norm{\nabla_{\boldtheta_m} L_i(\Theta)}^2 + 2 \sum_{m \neq n} \nabla_{\boldtheta_m} L_i(\Theta)^\top \nabla_{\boldtheta_n} L_i(\Theta)} \\
    &+
    \sum_{i \neq j} \Bigg( (P+2) \sum_{m=1}^r \nabla_{\boldtheta_m} L_i(\Theta)^\top \nabla_{\boldtheta_m} L_j(\Theta) + \sum_{m=1}^r \sum_{n \neq m} \nabla_{\boldtheta_n} L_i(\Theta)^\top \nabla_{\boldtheta_n} L_j(\Theta) \\
    &\qquad + 2 \sum_{m \neq n} \nabla_{\boldtheta_m} L_i(\Theta)^\top \nabla_{\boldtheta_n} L_j(\Theta) \Bigg)
\end{align}

To obtain $\tr(\Var(\gpesanti))$, we subtract the following from the expression above:
\begin{align}
    \expect{\boldepsilon}{\gpesanti}^\top \expect{\boldepsilon}{\gpesanti}
    &=
    \nabla_{\boldtheta} L(\Theta)^\top \nabla_{\boldtheta} L(\Theta) \\
    &=
    \pp{\sum_{t=1}^T \sum_{\tau=1}^t \nabla_{\boldtheta_\tau} L_t(\Theta)}^\top \pp{\sum_{t=1}^T \sum_{\tau=1}^t \nabla_{\boldtheta_\tau} L_t(\Theta)}
\end{align}

\subsection{Considering the Dependence on $T$}
The variance depends on the gradients of each loss term $L_t$ with respect to each of the per-timestep parameters $\boldtheta_\tau$.
To gain insight into the structure of these gradients, we can arrange them in a matrix:
\begin{align}
M
=
\begin{bmatrix}
\nabla_{\boldtheta_1} L_1 & \nabla_{\boldtheta_1} L_2 & \nabla_{\boldtheta_1} L_3 & \cdots & \nabla_{\boldtheta_1} L_T \\
\nabla_{\boldtheta_2} L_1 & \nabla_{\boldtheta_2} L_2 & \nabla_{\boldtheta_2} L_3 & \cdots & \nabla_{\boldtheta_2} L_T \\
\nabla_{\boldtheta_3} L_1 & \nabla_{\boldtheta_3} L_2 & \nabla_{\boldtheta_3} L_3 & \cdots & \nabla_{\boldtheta_3} L_T \\
\vdots & \vdots & \vdots & \ddots & \vdots \\
\nabla_{\boldtheta_T} L_1 & \nabla_{\boldtheta_T} L_2 & \nabla_{\boldtheta_T} L_3 & \cdots & \nabla_{\boldtheta_T} L_T \\
\end{bmatrix}
=
\begin{bmatrix}
\nabla_{\boldtheta_1} L_1 & \nabla_{\boldtheta_1} L_2 & \nabla_{\boldtheta_1} L_3 & \cdots & \nabla_{\boldtheta_1} L_T \\
0 & \nabla_{\boldtheta_2} L_2 & \nabla_{\boldtheta_2} L_3 & \cdots & \nabla_{\boldtheta_2} L_T \\
0 & 0 & \nabla_{\boldtheta_3} L_3 & \cdots & \nabla_{\boldtheta_3} L_T \\
\vdots & \vdots & \vdots & \ddots & \vdots \\
0 & 0 & 0 & \cdots & \nabla_{\boldtheta_T} L_T \\
\end{bmatrix}
\label{eq:matrix}
\end{align}
The RHS is upper-triangular due to the fact that $\nabla_{\boldtheta_\tau} L_t = 0$ for all $\tau > t$.
The variance of the PES estimator depends on the \textit{covariance between the gradients} $\nabla_{\boldtheta_\tau} L_t$ in this matrix.

We consider two structures for the matrix: 1) a \textit{diagonal} structure, where the gradients $\nabla_{\boldtheta_i} L_j = 0, \forall i \neq j$; and 2) an \textit{upper-triangular} structure as shown in the RHS of Eq.~\ref{eq:matrix}.
For each of these two matrix structures, we will consider two scenarios for the covariance between gradients: a) all gradients $\nabla_{\boldtheta_i} L_j$ are identical; b) all gradients are i.i.d.

\subsubsection{Diagonal Structure}
We denote the gradient of $L$ by $\boldg = \nabla_{\boldtheta} L(\Theta) = \sum_{t=1}^T \nabla_{\boldtheta} L_t(\Theta) = \sum_{t=1}^T \boldg_t$.
(Note that $\nabla_{\boldtheta} L_t(\Theta) = \cancel{\nabla_{\boldtheta_1} L_t} + \cancel{\nabla_{\boldtheta_2} L_t} + \cdots + \nabla_{\boldtheta_t} L_t = \nabla_{\boldtheta_t} L_t$ due to the diagonal structure.)

In the diagonal case, we have:
\begin{align}
    \expect{\boldepsilon}{\gpesanti^\top \gpesanti}
    &=
    \sum_{i=1}^T \pp{(iP + 2) \sum_{m=1}^i \norm{\nabla_{\boldtheta_m} L_i(\Theta)}^2 + \cancel{2 \sum_{m \leq i, n \leq j, m \neq n} \nabla_{\boldtheta_m} L_i(\Theta)^\top \nabla_{\boldtheta_n} L_i(\Theta)}} \\
    &+
    \sum_{i \neq j} \Bigg( (P+2) \cancel{\sum_{m=1}^r \nabla_{\boldtheta_m} L_i(\Theta)^\top \nabla_{\boldtheta_m} L_j(\Theta)} + \cancel{\sum_{m=1}^r \sum_{n \leq r, n \neq m} \nabla_{\boldtheta_n} L_i(\Theta)^\top \nabla_{\boldtheta_n} L_j(\Theta)} \\
    &\qquad \qquad + 2 \sum_{m \leq i, n \leq j, m \neq n} \nabla_{\boldtheta_m} L_i(\Theta)^\top \nabla_{\boldtheta_n} L_j(\Theta) \Bigg)
\end{align}

Thus, we have:
\begin{align}
    \expect{\boldepsilon}{\gpesanti^\top \gpesanti}
    &=
    \sum_{i=1}^T \pp{(iP + 2) \sum_{m=1}^i \norm{\nabla_{\boldtheta_m} L_i(\Theta)}^2} + 2 \sum_{i \neq j} \sum_{m \leq i, n \leq j, m \neq n}\nabla_{\boldtheta_m} L_i(\Theta)^\top \nabla_{\boldtheta_n} L_j(\Theta)  \label{eq:diag-simple1} \\
    &=
    \sum_{i=1}^T (iP + 2) \norm{\nabla_{\boldtheta_i} L_i(\Theta)}^2 + 2 \sum_{i \neq j} \nabla_{\boldtheta_i} L_i(\Theta)^\top \nabla_{\boldtheta_j} L_j(\Theta)  \label{eq:diag-simple2}
\end{align}
To go from Eq.~\ref{eq:diag-simple1} to Eq.~\ref{eq:diag-simple2}, we use the fact that $\nabla_{\boldtheta_m} L_i(\Theta) = 0$ for $m \neq i$ and $\nabla_{\boldtheta_n} L_j(\Theta) = 0$ for $n \neq j$.
Next, note that when $M$ is diagonal, we have:
\begin{align}
    \mathbb{E}[\gpesanti]^\top \mathbb{E}[\gpesanti]
    &=
    \pp{\sum_{t=1}^T \sum_{\tau=1}^t \nabla_{\boldtheta_\tau} L_t(\Theta)}^\top \pp{\sum_{t=1}^T \sum_{\tau=1}^t \nabla_{\boldtheta_\tau} L_t(\Theta)} \\
    &=
    \pp{\sum_{t=1}^T \nabla_{\boldtheta_t} L_t(\Theta)}^\top \pp{\sum_{t=1}^T \nabla_{\boldtheta_t} L_t(\Theta)} \\
    &=
    \sum_{t=1}^T \norm{\nabla_{\boldtheta_t} L_t(\Theta)}^2 + \sum_{i \neq j} \nabla_{\boldtheta_i} L_i(\Theta)^\top \nabla_{\boldtheta_j} L_j(\Theta)
\end{align}
Thus, the total variance is:
\begin{align}
    \tr(\Var(\gpesanti))
    &=
    \expect{\boldepsilon}{\gpesanti^\top \gpesanti} - \mathbb{E}[\gpesanti]^\top \mathbb{E}[\gpesanti] \\
    &=
    \sum_{i=1}^T (iP + 2) \norm{\nabla_{\boldtheta_i} L_i(\Theta)}^2 + 2 \sum_{i \neq j} \nabla_{\boldtheta_i} L_i(\Theta)^\top \nabla_{\boldtheta_j} L_j(\Theta) \\
    &\qquad -
    \sum_{i=1}^T \norm{\nabla_{\boldtheta_t} L_t(\Theta)}^2 - \sum_{i \neq j} \nabla_{\boldtheta_i} L_i(\Theta)^\top \nabla_{\boldtheta_j} L_j(\Theta) \\
    &=
    \boxed{\sum_{i=1}^T (iP + 1) \norm{\nabla_{\boldtheta_i} L_i(\Theta)}^2 + \sum_{i \neq j} \nabla_{\boldtheta_i} L_i(\Theta)^\top \nabla_{\boldtheta_j} L_j(\Theta)}
\end{align}

\paragraph{Scenario 1: All the $\nabla_{\boldtheta_i} L_i$ are equal.}
Recall our notation for the total gradient, $\boldg = \nabla_{\boldtheta} L(\Theta) = \sum_{t=1}^T \nabla_{\boldtheta} L_t(\Theta) = \sum_{t=1}^T \boldg_t$.
If we assume that the gradients for each unroll are identical to each other, then:
\begin{align}
    \boldg_t &= \frac{1}{T} \boldg \\
    \norm{\boldg_t}^2 &= \norm{\frac{1}{T} \boldg}^2 = \frac{1}{T^2} \norm{\boldg}^2
\end{align}
So,
\begin{align}
    \sum_{t=1}^T \norm{\boldg_t}^2 (tP + 1) + \sum_{i \neq j} \boldg_i^\top \boldg_j
    &=
    \sum_{t=1}^T \frac{1}{T^2} \norm{\boldg}^2 (tP + 1) + \sum_{i \leq T, j \leq T, i \neq j} \frac{1}{T^2} \norm{\boldg}^2 \\
    &=
    \frac{1}{T^2} \norm{\boldg}^2 \pp{T + P \frac{T (T+1)}{2}} + \frac{1}{T^2} \norm{\boldg}^2 (T^2 - T) \\
    &=
    \frac{1}{T^2} \norm{\boldg}^2 \pp{T^2 + \frac{PT^2 + PT}{2}} \\
    &=
    \norm{\boldg}^2 \pp{\frac{P}{2T} + \frac{P}{2} + 1}
\end{align}

\paragraph{Scenario 2: All the $\nabla_{\boldtheta_i} L_j$ are i.i.d.}
If we assume that the gradients for each unroll are i.i.d., then:
\begin{align}
    \label{eq variance unroll}
    \expect{}{\norm{\boldg}^2} &= T\, \expect{}{\norm{\boldg_t}^2}  \\
    \expect{}{\norm{\boldg_t}^2} &= \frac{1}{T}\, \expect{}{\norm{\boldg}^2} \\
    \expect{}{\norm{\boldg_t}^2} &= \frac{1}{T} \expect{}{\norm{\boldg}^2}
\end{align}
Thus,
\begin{align}
    \sum_{t=1}^T \norm{\boldg_t}^2 (tP + 1) + \sum_{i \neq j} \boldg_i^\top \boldg_j
    &=
    \sum_{t=1}^T \frac{1}{T} \norm{\boldg}^2 (tP + 1) + \sum_{i \neq j} \frac{1}{T} \norm{\boldg}^2 \\
    &=
    \frac{1}{T} \norm{\boldg}^2 \pp{T + P \frac{T (T+1)}{2}} + \frac{1}{T} \norm{\boldg}^2 \pp{T^2 - T} \\
    &=
    \frac{1}{T} \norm{\boldg}^2 \pp{T + T^2 - T + \frac{PT (T+1)}{2}} \\
    &=
    \frac{1}{T} \norm{\boldg}^2 \pp{T^2 + \frac{PT (T+1)}{2}} \\
    &=
    \norm{\boldg}^2 \pp{T + \frac{P (T+1)}{2}} \\
    &=
    \norm{\boldg}^2 \pp{\frac{PT}{2} + \frac{P}{2} + T}
\end{align}

\subsubsection{Upper-Triangular Structure}

\paragraph{Scenario 1: All the $\nabla_{\boldtheta_i} L_j$ are equal.}
Suppose all the terms in the matrix are equal, e.g., $\nabla_{\boldtheta_i} L_j = \boldh, \forall i,j$.
The total gradient $\boldg = \nabla_{\boldtheta} L(\Theta)$ is equal to the sum of the gradients in the upper-triangular matrix.
Thus, $\boldg = \frac{T (T+1)}{2} \boldh$, so we can write:
\begin{align}
    \boldh = \frac{2}{T (T+1)} \boldg
\end{align}

We have:
\begin{align}
    \expect{\boldepsilon}{\gpesanti^\top \gpesanti}
    &=
    \underbrace{\sum_{i=1}^T \pp{(iP +2) \sum_{m=1}^i \norm{\boldh}^2 + 2 \sum_{m \leq i, n \leq i, m \neq n} \norm{\boldh}^2}}_{\num{I}} \\
    &+
    \underbrace{\sum_{i \neq j} \pp{(P+2) \sum_{m=1}^r \norm{\boldh}^2 + \sum_{m=1}^r \sum_{m \leq r, n \leq r, n \neq m} \norm{\boldh}^2 + 2 \sum_{m \leq i, n \leq j, m \neq n} \norm{\boldh}^2}}_{\num{II}}
\end{align}

Term $\num{I}$ is as follows:
\begin{align}
    \num{I}
    &=
    \sum_{i=1}^T \pp{(iP + 2) i \norm{\boldh}^2 + 2 (i^2 - i) \norm{\boldh}^2} \\
    &=
    \norm{\boldh}^2 \pp{\sum_{i=1}^T (P + 2) i^2} \\
    &=
    \norm{\boldh}^2 (P + 2) \frac{T (T+1) (2T+1)}{6}
\end{align}

Term $\num{II}$ is as follows:
\begin{align}
    \num{II}
    &=
    \sum_{i \neq j} \Big( \underbrace{(P+2) \sum_{m=1}^r \norm{\boldh}^2}_{(P+2) r \norm{\boldh}^2} + \underbrace{\sum_{m=1}^r \sum_{n \neq m} \norm{\boldh}^2}_{r (r-1) \norm{\boldh}^2} + \underbrace{2 \sum_{m \neq n} \norm{\boldh}^2}_{2 (ij - r) \norm{\boldh}^2} \Big) \\
    &=
    \norm{\boldh}^2 \sum_{i \leq T, j \leq T, i \neq j} (Pr + r^2 - r + 2 ij) \\
    &=
    2 \norm{\boldh}^2 \sum_{i=1}^T \sum_{j=1}^{i-1} \pp{(P - 1) j + j^2 + 2 ij} \\
    &=
    2 \norm{\boldh}^2 \pp{\underbrace{\sum_{i=1}^T \sum_{j=1}^{i-1} (P - 1) j}_{\num{a}} + \underbrace{\sum_{i=1}^T \sum_{j=1}^{i-1} j^2}_{\num{b}} + \underbrace{\sum_{i=1}^T \sum_{j=1}^{i-1} 2 ij}_{\num{c}}}
    \label{eq:upper-triangular-terms}
\end{align}

Next we derive each of the terms that arise in Eq.~\ref{eq:upper-triangular-terms}.

\begin{align}
    \num{a}
    =
    \sum_{i=1}^T \sum_{j=1}^{i-1} (P-1) j
    &=
    (P-1) \sum_{i=1}^T \sum_{j=1}^{i-1} j \\
    &=
    (P-1) \sum_{i=1}^T \frac{i (i-1)}{2} \\
    &=
    \frac{(P-1)}{2} \pp{\frac{T (T+1) (2T+1)}{6} - \frac{T (T+1)}{2}}
\end{align}

\begin{align}
    \num{b}
    =
    \sum_{i=1}^T \sum_{j=1}^{i-1} j^2
    &=
    \sum_{i=1}^T \frac{i (i-1) (2i-1)}{6} \\
    &=
    \sum_{i=1}^T \frac{1}{6} \pp{2i^3 - 3i^2 + i} \\
    &=
    \frac{1}{3} \sum_{i=1}^T i^3 - \frac{1}{2} \sum_{i=1}^T i^2 + \frac{1}{6} \sum_{i=1}^T i \\
    &=
    \frac{1}{3} \frac{T^2 (T+1)^2}{4} - \frac{1}{2} \frac{T (T+1) (2T+1)}{6} + \frac{1}{6} \frac{T (T+1)}{2} \\
    &=
    \frac{1}{12} \pp{T^2 (T+1)^2 - T(T+1)(2T+1) + T(T+1)}
\end{align}

\begin{align}
    \num{c}
    =
    \sum_{i=1}^T \sum_{j=1}^{i-1} 2 ij
    &=
    \sum_{i=1}^T i \sum_{j=1}^{i-1} j \\
    &=
    \sum_{i=1}^T i \pp{\frac{i (i-1)}{2}} \\
    &=
    \sum_{i=1}^T \frac{1}{2} i (i^2 - i) \\
    &=
    \frac{1}{2} \sum_{i=1}^T i^3 - i^2 \\
    &=
    \frac{1}{2} \pp{\frac{T^2 (T+1)^2}{4} - \frac{T (T+1) (2T+1)}{6}}
\end{align}

Combining all these terms, we obtain the following expression for the total variance:
\begin{align}
    \norm{\boldh}^2 \pp{\frac{PT (T + 1) (2T + 1)}{3} - \frac{PT (T+1)}{2} + \frac{5T^2 (T+1)^2}{12} - \frac{T (T + 1) (2T + 1)}{6} + \frac{2T (T + 1)}{3}}
\end{align}
Combining terms, we obtain:
\begin{align}
    \norm{\boldh}^2 \pp{\frac{5}{12} T^4 + \frac{2}{3} PT^3 + \frac{1}{2} PT^2 - \frac{1}{6} PT + \frac{1}{2} T^3 + \frac{7}{12} T^2 + \frac{1}{2} T}
\end{align}

We are interested in the scaling behavior as a function of the total gradient norm $\norm{\boldg}^2$, where
\begin{align}
\norm{\boldh}^2 = \pp{\frac{2}{T (T+1)}}^2 \norm{\boldg}^2  \label{eq:h-and-g-relation}
\end{align}
Because the denominator in Eq.~\ref{eq:h-and-g-relation} is $\mathcal{O}(T^4)$, we will have terms in the total variance that scale as:
\begin{align}
\norm{\boldg}^2 \pp{\mathcal{O} \pp{1} + \mathcal{O}\pp{\frac{P}{T}} + \mathcal{O} \pp{\frac{P}{T^2}} - \mathcal{O} \pp{\frac{P}{T^3}} + \mathcal{O} \pp{\frac{1}{T}} + \mathcal{O} \pp{\frac{1}{T^2}} + \mathcal{O} \pp{\frac{1}{T^3}}}
\end{align}

\paragraph{Scenario 2: All the $\nabla_{\boldtheta_i} L_j$ are i.i.d.}
In this case, by direct analogy to Equation \ref{eq variance unroll}, we have:
\begin{align}
    \expect{\boldepsilon}{\norm{\boldh}^2} = \frac{2}{T (T+1)} \expect{\boldepsilon}{\norm{\boldg}^2}
\end{align}

Here, the denominator is of order $O(T^2)$, while the numerator is of order $O(T^4)$, yielding variance that scales as $\mathcal{O}(T^2)$:
\begin{align}
\norm{\boldg}^2 \pp{\mathcal{O} \pp{T^2} + \mathcal{O}\pp{PT} + \mathcal{O} \pp{P} - \mathcal{O} \pp{\frac{P}{T}} + \mathcal{O} \pp{T} + \mathcal{O} \pp{1} + \mathcal{O} \pp{\frac{1}{T}}}
\end{align}

\begin{figure}[H]
     \centering
     \begin{subfigure}[b]{0.31\textwidth}
         \centering
        \includegraphics[width=\textwidth]{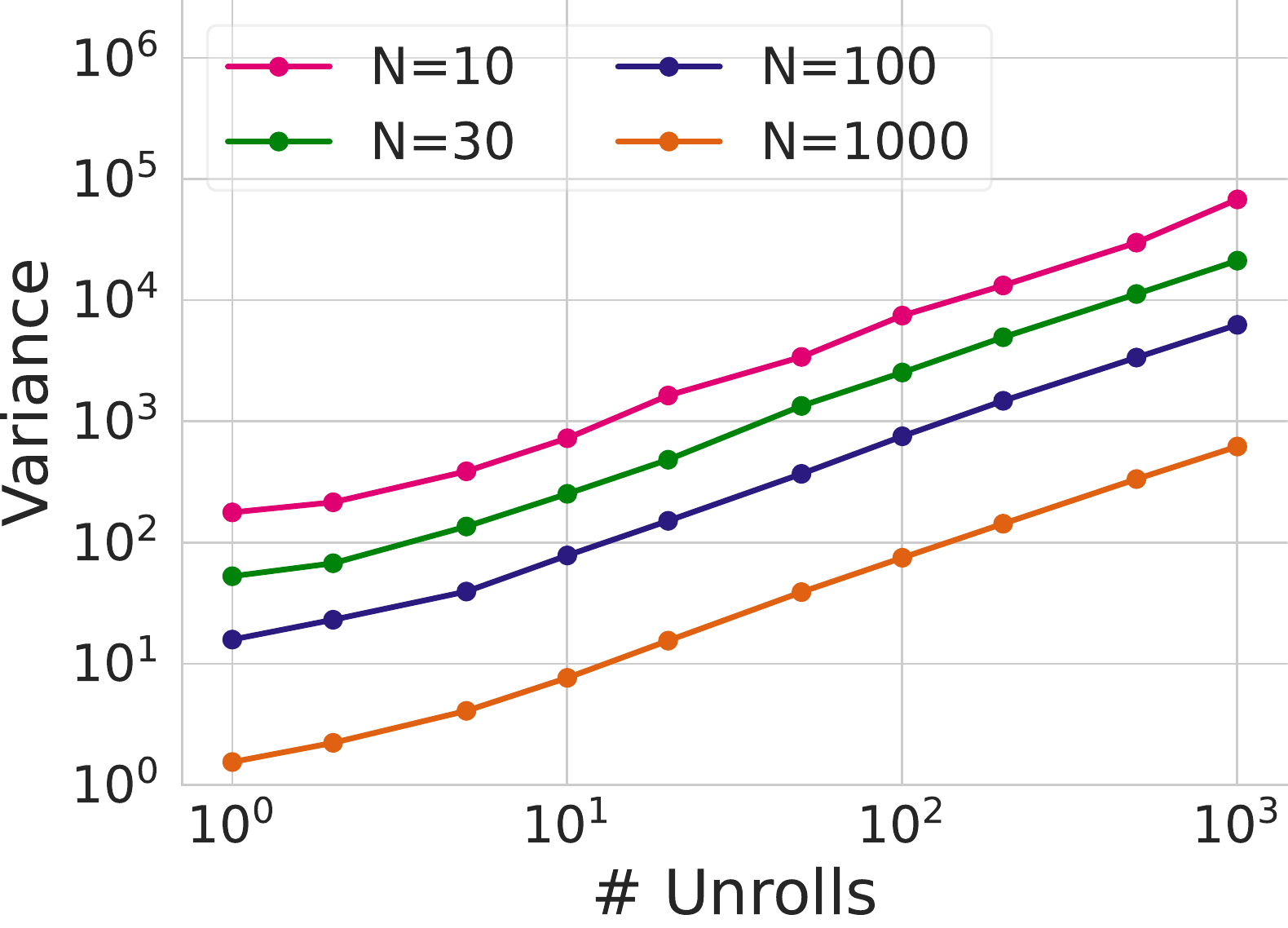}
         \caption{Random sequence}
         \label{fig:lstm-variance-random-seq}
     \end{subfigure}
     \hfill
     \begin{subfigure}[b]{0.31\textwidth}
         \centering
        \includegraphics[width=\textwidth]{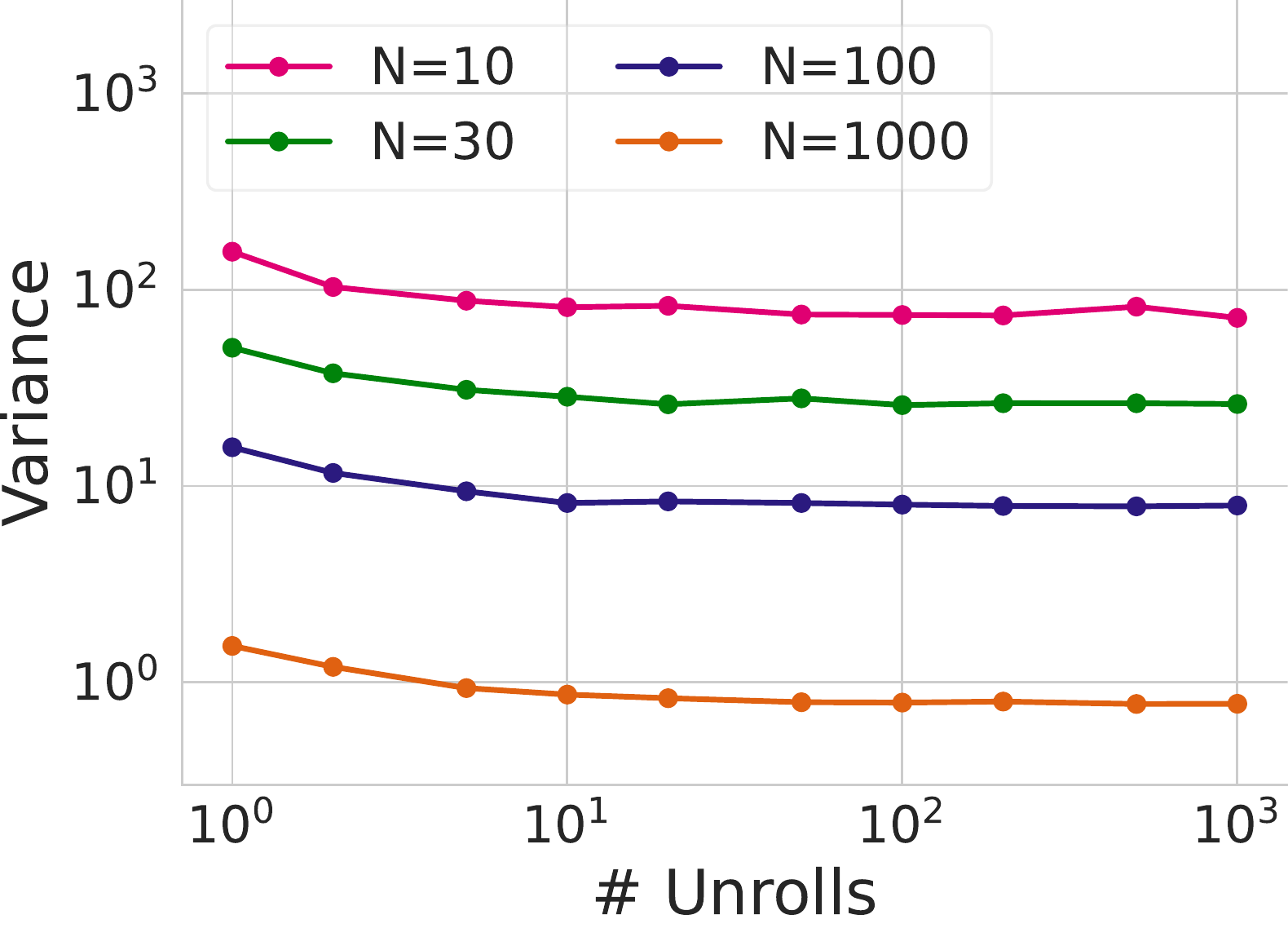}
         \caption{Single character repeated}
         \label{fig:lstm-variance-same-char}
     \end{subfigure}
     \hfill
     \begin{subfigure}[b]{0.31\textwidth}
         \centering
        \includegraphics[width=\textwidth]{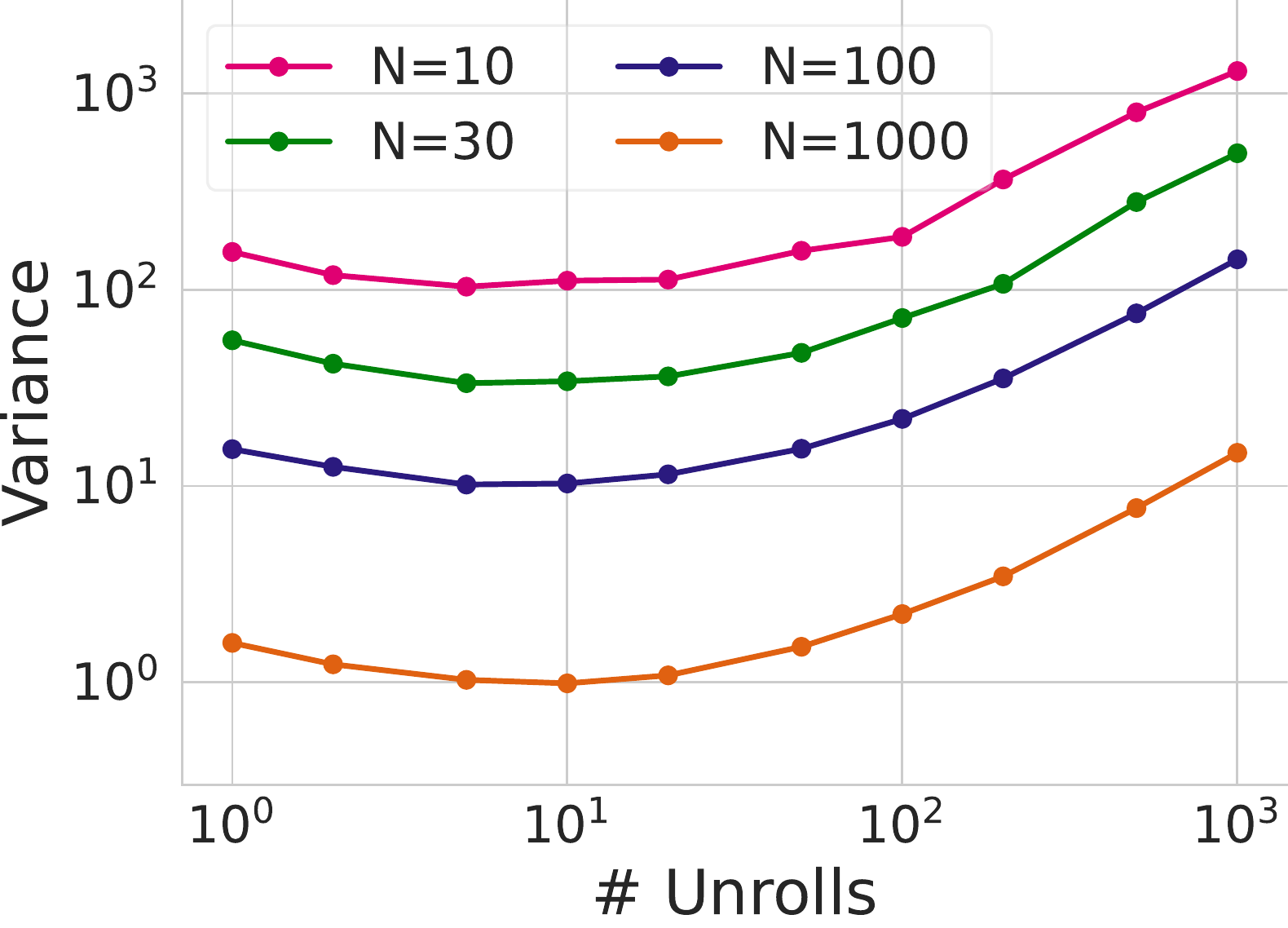}
         \caption{Real PTB sequence}
         \label{fig:lstm-variance-real-data}
     \end{subfigure}
        \vspace{-0.3cm}
        \caption{\textbf{Empirical variance measurements for three scenarios.}}
        \label{fig:lstm-variance-scenarios}
\end{figure}

Figure~\ref{fig:lstm-variance-scenarios} shows the empirical variance for several potential scenarios.
We performed an analysis similar to that in Section~\ref{sec:pes}, measuring the variance of the PES gradient with respect to the number of unrolls for a small LSTM on the Penn TreeBank (PTB) dataset.
We constructed synthetic data sequences to illustrate different scenarios: in Figure~\ref{fig:lstm-variance-random-seq} we used a $10^3$ length sequence consisting of characters sampled uniformly at random from the PTB vocabulary, simulating the first scenario; in Figure~\ref{fig:lstm-variance-same-char} we used a $10^3$ length sequence consisting of a single repeated character, simulating the second scenario;
Figure~\ref{fig:lstm-variance-real-data} shows the variance for real data---the first $10^3$ characters of PTB---which exhibits characteristics of both synthetic scenarios.

\section{Reducing Variance by Incorporating the Analytic Gradient}
\label{app:analytic-gradient}

\begin{figure}[H]
     \centering
     \begin{subfigure}[b]{0.31\textwidth}
         \centering
         \includegraphics[width=\textwidth]{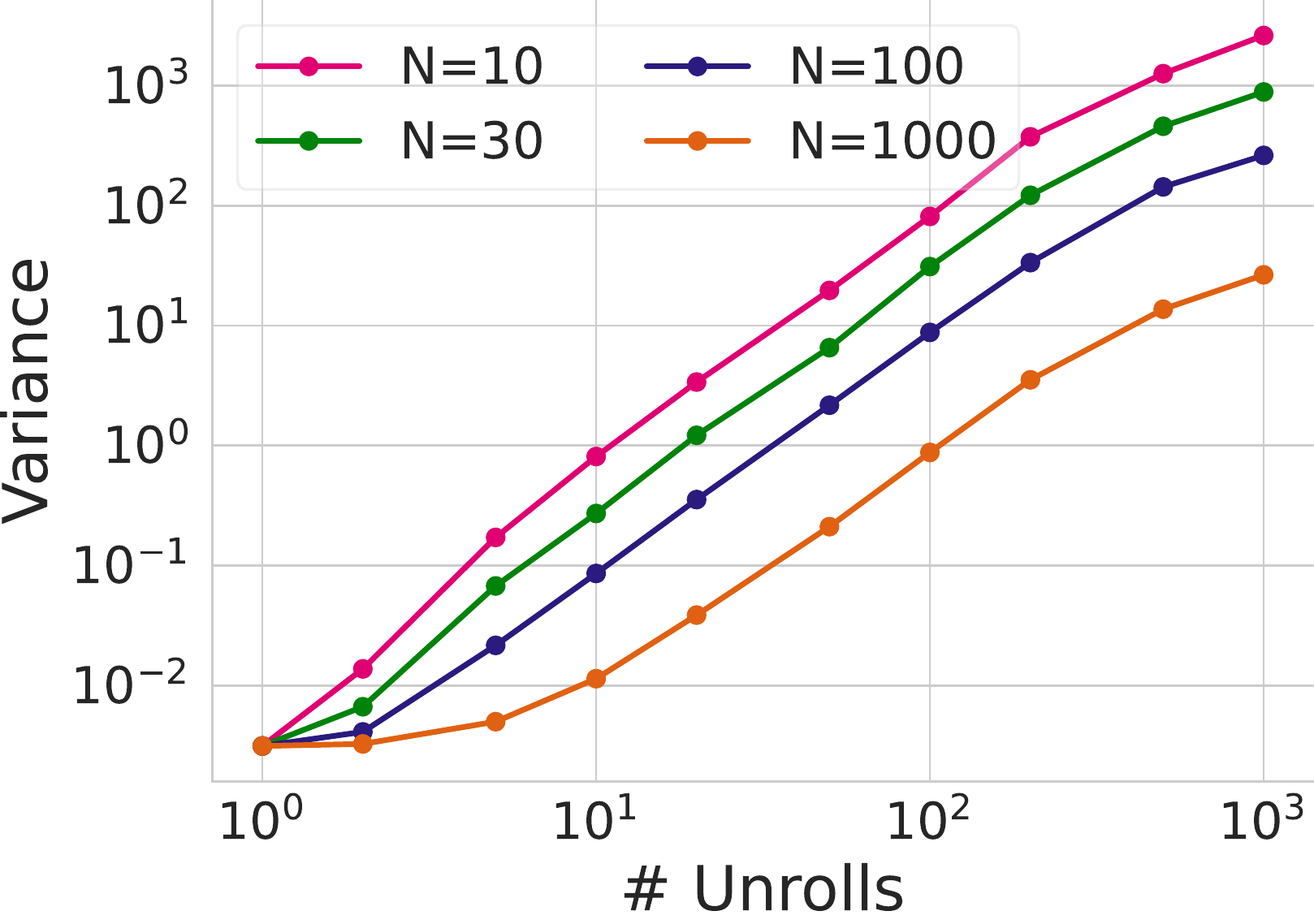}
         \caption{Random sequence}
         \label{fig:analytic-lstm-variance-random-seq}
     \end{subfigure}
     \hfill
     \begin{subfigure}[b]{0.31\textwidth}
         \centering
         \includegraphics[width=\textwidth]{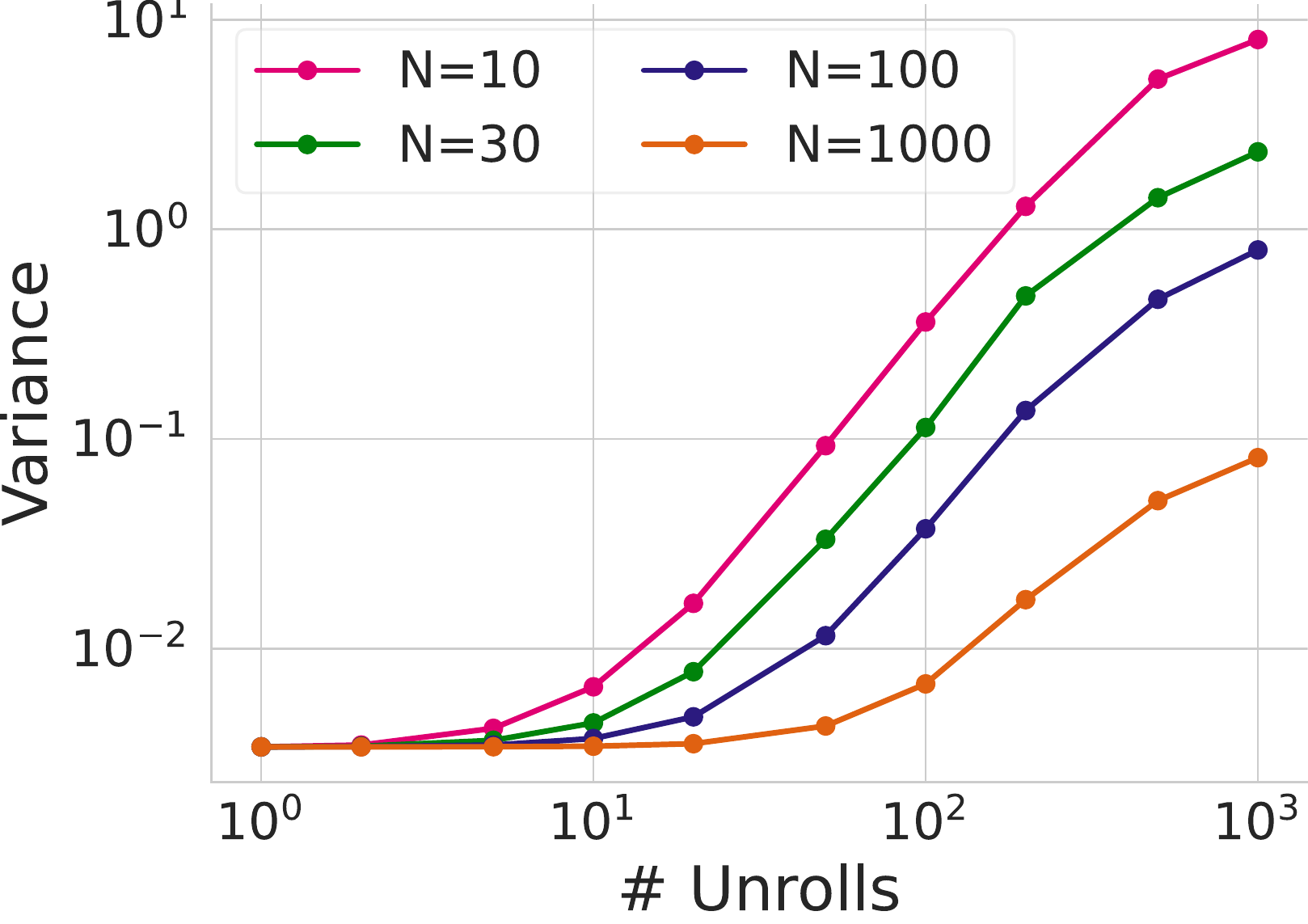}
         \caption{Single character repeated}
         \label{fig:analytic-lstm-variance-same-char}
     \end{subfigure}
     \hfill
     \begin{subfigure}[b]{0.31\textwidth}
         \centering
         \includegraphics[width=\textwidth]{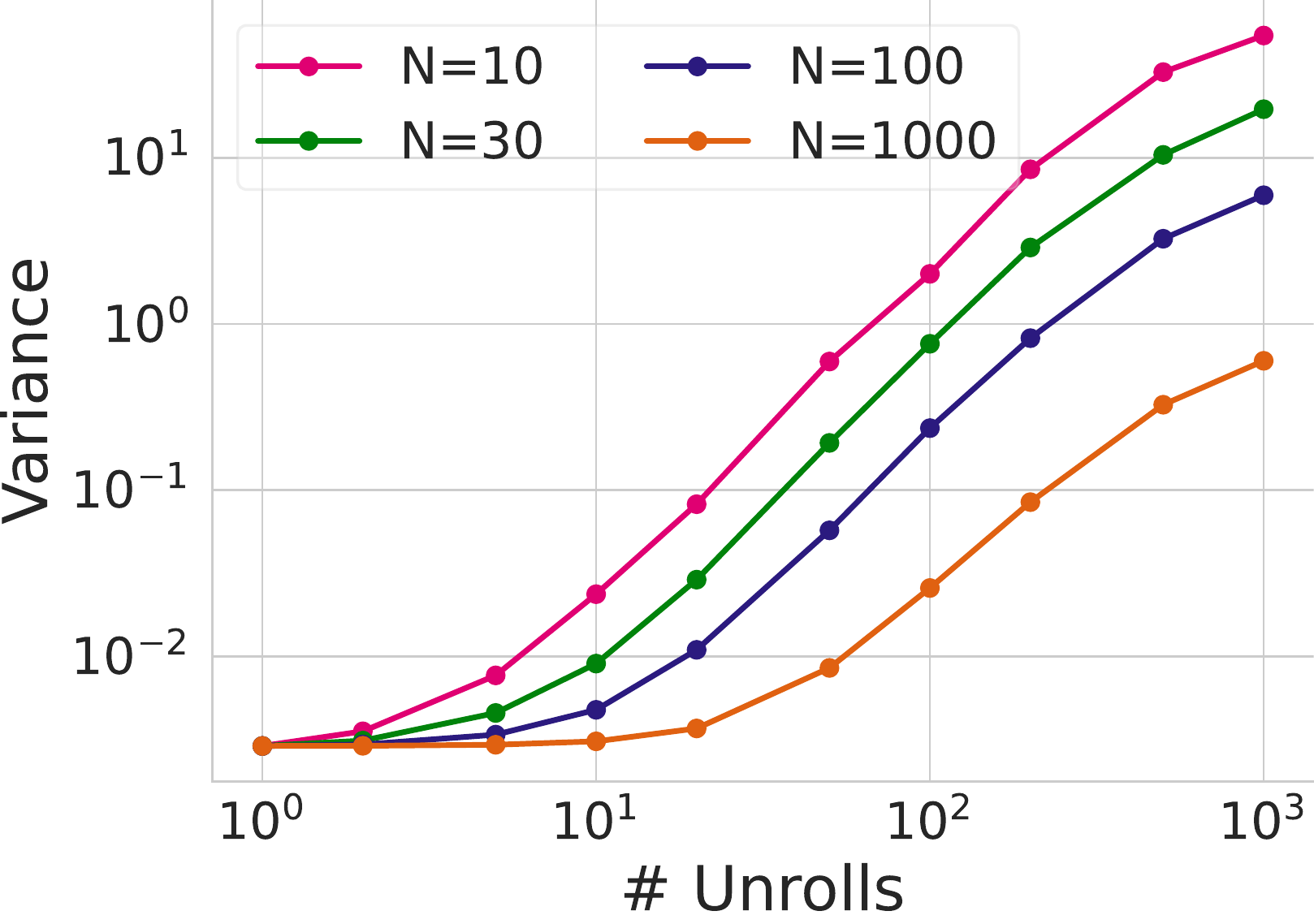}
         \caption{Real PTB sequence}
         \label{fig:analytic-lstm-variance-real-data}
     \end{subfigure}
        \vspace{-0.3cm}
        \caption{\textbf{Empirical variance measurements for three scenarios, incorporating the analytic gradient from the most recent unroll to reduce variance.}}
        \label{fig:analytic-lstm-variance-scenarios}
\end{figure}

For functions $L$ that are differentiable, we can use the analytic gradient from the most recent partial unroll (e.g., backpropagating through the last $K$-step unroll) to reduce the variance of the PES gradient estimates.
Below, we show how we can incorporate the analytic gradient in the ES estimate for $\frac{\partial L_t(\Theta)}{\partial \boldtheta}$:
\begin{align}
    \frac{\partial L_t(\Theta)}{\partial \boldtheta}
    &\approx \frac{1}{\sigma^2} \mathbb{E}_{\boldepsilon}\left[ \left( \sum_{\tau \leq t} \boldepsilon_\tau \right) L_t(\Theta + \boldepsilon) \right] \\
    &= \frac{1}{\sigma^2} \mathbb{E}_{\boldepsilon} \left[ \left( \sum_{\tau < t} \boldepsilon_\tau \right) L_t(\Theta + \boldepsilon) \right] + \frac{1}{\sigma^2} \mathbb{E}_{\boldepsilon} [\boldepsilon_t L_t(\Theta + \boldepsilon)] \\
    &= \frac{1}{\sigma^2} \mathbb{E}_{\boldepsilon} \left[ \left( \sum_{\tau < t} \boldepsilon_\tau \right) L_t(\Theta + \boldepsilon) \right] + \underbrace{\frac{\partial L_t(\Theta)}{\partial \boldtheta_t}}_{\equiv \boldp_t} \\
    &= \frac{1}{\sigma^2} \mathbb{E}_{\boldepsilon} \left[ \left( \sum_{\tau < t} \boldepsilon_\tau \right) L_t(\Theta + \boldepsilon) \right] + \boldp_t - \underbrace{\frac{1}{\sigma^2} \mathbb{E}_{\boldepsilon} \left[ \left( \sum_{\tau < t} \boldepsilon_\tau \right) \boldepsilon_t^\top \boldp_t \right]}_{= 0} \\
    &= \frac{1}{\sigma^2} \mathbb{E}_{\boldepsilon} \left[ \left( \sum_{\tau < t} \boldepsilon_\tau \right) (L_t(\Theta + \boldepsilon) - \boldepsilon_t^\top \boldp_t) \right] + \boldp_t
\end{align}
We call the resulting estimator PES+Analytic.
Algorithm~\ref{alg:pes-analytic} describes the implementation of this estimator, which requires a few simple changes from the standard PES estimator.
We repeated the empirical variance measurement described in Section~\ref{sec:pes} and Appendix~\ref{app:pes-variance} using the PES+Analytic estimator, for each of the three scenarios from Appendix~\ref{app:pes-variance}, shown in Figure~\ref{fig:analytic-lstm-variance-scenarios}.
Similarly to the other variance measurements, we report variance normalized by the squared norm of the true gradient.
We found that variance increases with the number of unrolls, but the PES+Analytic variance is 1-2 orders of magnitude smaller than the standard PES variance.

\begin{figure*}[h]
\label{fig-algs-analytic}
\begin{minipage}{0.48\textwidth}
\begin{algorithm}[H]
  \caption{Original persistent evolution strategies (PES) estimator, identical to Section~\ref{sec:pes}.}
  \label{alg:pes-restated}
\begin{algorithmic}
    \State \textbf{Input:} $\bolds_0$, initial state
    \State \hspace{2.7em} $K$, truncation length for partial unrolls
    \State \hspace{2.7em} $N$, number of particles
    \State \hspace{2.9em} $\sigma$, standard deviation of perturbations
    \State \hspace{2.9em} $\alpha$, learning rate for PES optimization
    \State {\color{white}Initialize $\bolds = \bolds_0$}
    \State Initialize $\bolds^{(i)} = \bolds_0$ for $i \in \{1, \dots, N\}$
    \State Initialize $\boldxi^{(i)} \gets \boldzero$ for $i \in \{ 1, \dots, N \}$
    \While {true}
        \State {\color{white} $\bolds, L \gets \text{unroll}(\bolds, \boldtheta, K)$}
        \State {\color{white} $\boldp \gets \nabla_{\boldtheta} L$}
        \State $\gpes \gets \boldzero$
        \For{$i=1,\dots, N$}
            \State $\boldepsilon^{(i)} \gets 
                \left\{\begin{array}{lcl}
                	\text{draw from } \mathcal{N}(0, \sigma^2 I) &  & i \text{ odd} \\
                	-\boldepsilon^{(i-1)} &  & i \text{ even}
                \end{array}\right.$            
            \State $\bolds^{(i)}$, $\hat{L}_K^{(i)} \gets \text{unroll}(\bolds^{(i)}, \boldtheta + \boldepsilon^{(i)}, K)$  
            \State $\boldxi^{(i)} \gets \boldxi^{(i)} + \boldepsilon^{(i)}$
            \State $\gpes \gets \gpes + \boldxi^{(i)} \hat{L}_K^{(i)}$
        \EndFor
        \State $\gpes \gets \frac{1}{N \sigma^2} \gpes$
        \State $\boldtheta \gets \boldtheta - \alpha \gpes$
    \EndWhile
\end{algorithmic}
\end{algorithm}
\end{minipage}
\hfill
\begin{minipage}{0.48\textwidth}
\begin{algorithm}[H]
  \caption{PES + analytic gradient. Differences from PES are {\color{purple} highlighted in purple.}}
  \label{alg:pes-analytic}
\begin{algorithmic}
    \State \textbf{Input:} $\bolds_0$, initial state
    \State \hspace{2.7em} $K$, truncation length for partial unrolls
    \State \hspace{2.7em} $N$, number of particles
    \State \hspace{2.9em} $\sigma$, standard deviation of perturbations
    \State \hspace{2.9em} $\alpha$, learning rate for PES optimization
    \State {\color{purple}Initialize $\bolds = \bolds_0$}
    \State Initialize $\bolds^{(i)} = \bolds_0$ for $i \in \{1, \dots, N\}$
    \State Initialize $\boldxi^{(i)} \gets \boldzero$ for $i \in \{ 1, \dots, N \}$
    \While {true}
        \State {\color{purple} $\bolds, L \gets \text{unroll}(\bolds, \boldtheta, K)$}
        \State {\color{purple} $\boldp \gets \nabla_{\boldtheta} L$}
        \State $\gpes \gets \boldzero$
        \For{$i=1,\dots, N$}
            \State $\epsilon^{(i)} \gets 
                \left\{\begin{array}{lcl}
                	\text{draw from } \mathcal{N}(0, \sigma^2 I) &  & i \text{ odd} \\
                	-\boldepsilon^{(i-1)} &  & i \text{ even}
                \end{array}\right.$            
            \State $\bolds^{(i)}$, $\hat{L}_K^{(i)} \gets \text{unroll}(\bolds^{(i)}, \boldtheta + \boldepsilon^{(i)}, K)$  
            
            \State {\color{purple} $\gpes \gets \gpes + \boldxi^{(i)} (\hat{L}_K^{(i)} - {\boldepsilon^{(i)}}^\top \boldp$)}
            \State {\color{purple} $\boldxi^{(i)} \gets \boldxi^{(i)} + \boldepsilon^{(i)}$}
        \EndFor
        \State {\color{purple} $\gpes \gets \frac{1}{N \sigma^2} \gpes + \boldp$}
        \State $\boldtheta \gets \boldtheta - \alpha \gpes$
    \EndWhile
\end{algorithmic}
\end{algorithm}
\end{minipage}
\vspace{-0.1cm}
\caption{\textbf{A comparison of the PES and PES+Analytic gradient estimators}, applied to partial unrolls of a computation graph.
The conditional statement for $\boldepsilon^{(i)}$ is used to implement antithetic sampling.
For clarity, we describe the meta-optimization updates to $\boldtheta$ using SGD, but we typically use Adam in practice.
}
\vspace{-0.1cm}
\end{figure*}

\section{Connection to Gradient Estimation in Stochastic Computation Graphs}
\label{app:pes-stochastic-computation-graph}

\begin{figure}[h]
    \centering
    \includegraphics[width=0.3\linewidth]{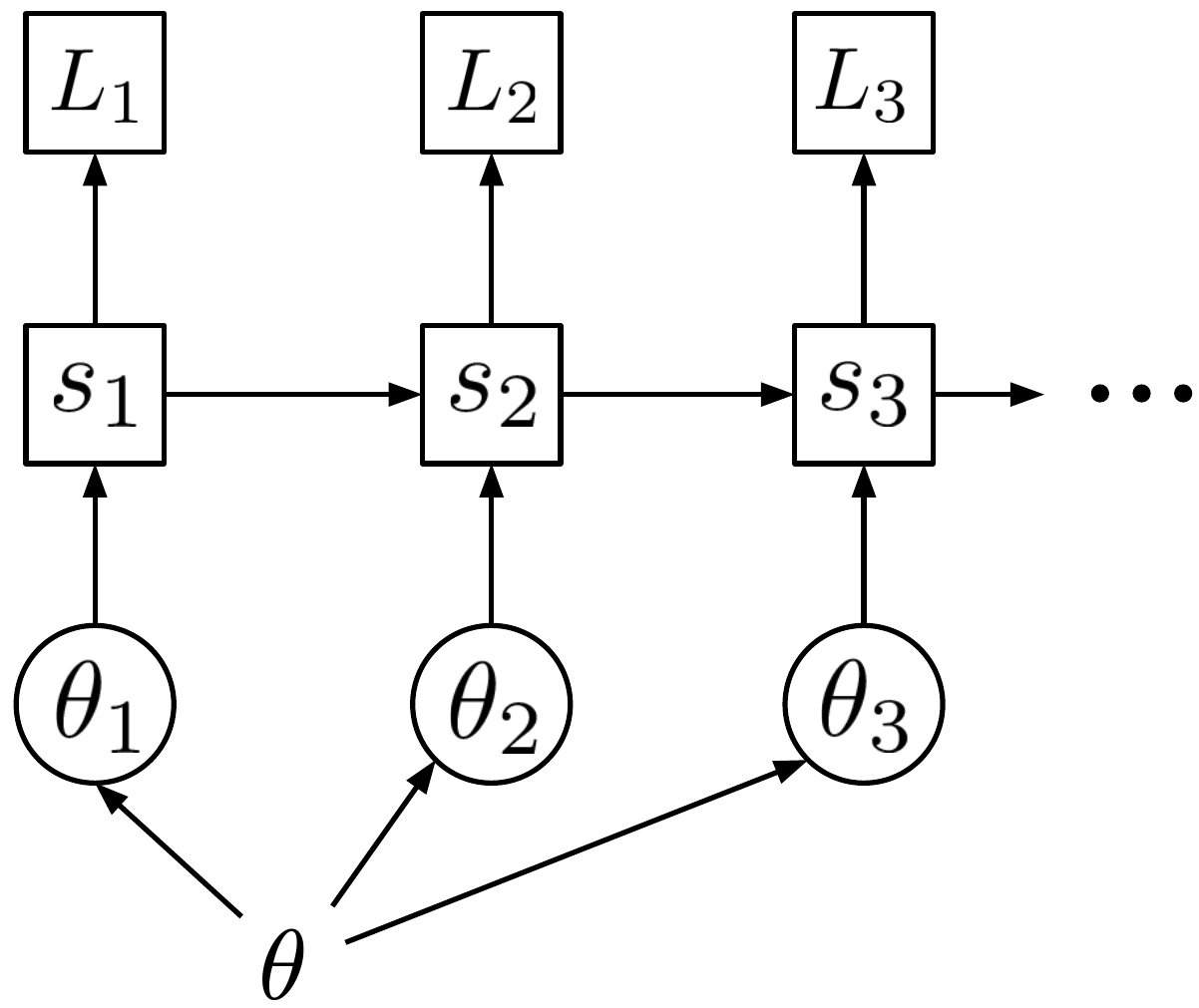}
    \caption{\textbf{Unrolled stochastic computation graph for PES}, in the notation of~\cite{schulman2015gradient}.
    }
    \label{fig:schulman-style-diagram}
\end{figure}

In this section, we show how PES can be derived using the framework for gradient estimation in stochastic computation graphs introduced in~\cite{schulman2015gradient}.
We follow their notation for this exposition: in Figure~\ref{fig:schulman-style-diagram}, squares represent deterministic nodes, which are functions of their parents; circles represent stochastic nodes which are distributed conditionally on their parents, and nodes not in squares or circles represent inputs.
For notational simplicity, in the following exposition we consider 1-dimensional $\theta$.
We represent the unrolled computation graph in terms of an input node $\theta$, that gives rise to a stochastic variable $\theta_t$ at each time step; the sampled $\theta_t$ is used to compute the state $s_t$, which is a deterministic function of the previous state $s_{t-1}$ and the current parameters $\theta_t$.
The losses $L_t$ are designated as cost nodes, and our objective is $L = \sum_t L_t$.

Theorem 1 from~\cite{schulman2015gradient} gives the following general form for the gradient of the sum of cost nodes in such a stochastic computation graph.
Here, $\mathcal{C}$ is the set of cost nodes; $\mathcal{S}$ is the set of stochastic nodes; $\text{DEPS}_w$ denotes the set of nodes that $w$ \textit{depends on}; $a \prec^D b$ indicates that node $a$ depends deterministically on node $b$ (note that this relationship holds as long as there are no stochastic nodes along a path from $a$ to $b$; in our case, $\theta \prec^D \theta_t$ holds for all $t$); and $\hat{Q}_w$ is the sum of cost nodes downstream from node $w$.
\begin{equation} \label{eq:schulman-general}
    \frac{\partial}{\partial \theta} \mathbb{E} \left[ \sum_{c \in \mathcal{C}} c \right] = \mathbb{E} \left[ \sum_{w \in \mathcal{S},\\ \theta \prec^D w} \left( \frac{\partial}{\partial \theta} \log p(w | \text{DEPS}_w) \right) \hat{Q}_w + \sum_{c \in \mathcal{C}, \theta \prec^D c} \frac{\partial}{\partial \theta} c(\text{DEPS}_c) \right]
\end{equation}

For the computation graph in Figure~\ref{fig:schulman-style-diagram}, $\theta$ does not deterministically influence any of the cost nodes $L_t$, so the second term in the expectation in Eq.~\ref{eq:schulman-general} will be 0.
In addition, each stochastic node $\theta_t$, depends only on $\theta$, e.g. $\text{DEPS}_{\theta_t} = \{ \theta \}, \forall t$.
Thus, our gradient estimate is:
\begin{equation} \label{eq:our-schulman-estimator}
    \frac{\partial}{\partial \theta} \mathbb{E} \left[ \sum_{t=1}^T L_t \right] = \mathbb{E} \left[ \sum_{t=1}^T \left( \frac{\partial}{\partial \theta} \log p(\theta_t | \theta) \right) \hat{Q}_{\theta_t} \right]
\end{equation}

$\hat{Q}_{\theta_t}$ is the sum of cost nodes downstream of $\theta_t$, thus $\hat{Q}_{\theta_t} = \sum_{i=t}^T L_i$.
Now, each $\theta_t \sim \mathcal{N}(\theta, \sigma^2)$, so we have:
\begin{equation}
    \log p(\theta_t \mid \theta) = \log \mathcal{N}(\theta_t \mid \theta, \sigma^2) = \log \frac{1}{\sqrt{2 \pi} \sigma} - \frac{1}{2 \sigma^2} (\theta_t - \theta)^2
\end{equation}
Then,
\begin{align}
    \frac{\partial}{\partial \theta} \log p(\theta_t \mid \theta)
    &= - \frac{1}{2 \sigma^2} \cdot 2 (\theta_t - \theta) \cdot (-1) \\
    &= \frac{1}{\sigma^2} (\theta_t - \theta) \\
    &= \frac{1}{\sigma^2} (\theta + \epsilon_t - \theta) \\
    &= \frac{1}{\sigma^2} \epsilon_t
\end{align}
where we used the reparameterization $\theta_t = \theta + \epsilon_t$ with $\epsilon_t \sim \mathcal{N}(0, \sigma^2)$.
Plugging this into Eq.~\ref{eq:our-schulman-estimator}, we have:
\begin{align}
    \frac{\partial}{\partial \theta} \mathbb{E} \left[ \sum_{t=1}^T L_t \right]
    &= \mathbb{E} \left[ \sum_{t=1}^T \frac{1}{\sigma^2} \epsilon_t \hat{Q}_{\theta_t} \right] \\
    &= \frac{1}{\sigma^2} \mathbb{E}\left[ \epsilon_1 (L_1 + L_2 + \cdots + L_T) + \epsilon_2 (L_2 + L_3 + \cdots + L_T) + \cdots + \epsilon_T L_T \right] \\
    &= \frac{1}{\sigma^2} \mathbb{E}[\epsilon_1 L_1 + (\epsilon_1 + \epsilon_2) L_2 + (\epsilon_1 + \epsilon_2 + \epsilon_3) L_3 + \cdots + (\epsilon_1 + \cdots + \epsilon_T) L_T] \\
    &= \frac{1}{\sigma^2} \mathbb{E} \left[ \sum_{t=1}^T \left( \sum_{\tau=1}^t \epsilon_\tau \right) L_t \right] \label{eq:pes-from-schulman}
\end{align}
Eq.~\ref{eq:pes-from-schulman} recovers the PES estimator.

\section{Derivations and Compute/Memory Costs}
\label{app:deriv-RTRLmemory}

\paragraph{BPTT, TBPTT, ARTBP.}
Backpropagating through a full unroll of $T$ steps requires $T$ forward and backward passes, yielding compute $T(F + B)$; all $T$ states must be stored in memory to be available for gradient computation during backprop, yielding memory cost $TS$.
Similarly, because TBPTT unrolls the computation graph for $K$ steps, it requires $K$ forward and backward passes, yielding computation $K (F + B)$, and requires storing $K$ states in memory, yielding memory cost $KS$.
ARTBP is identical to TBPTT except that it randomly samples the truncation length in a theoretically-justified way to reduce or eliminate truncation bias.
In theory, the sampled truncation lengths must allow for maximum length $T$, yielding worst-case compute $T (F+B)$ and memory cost $TS$.
However, in practice this is often intractable, so truncation lengths may be sampled within a restricted range centered around $K$---this is no longer unbiased, but yields average case compute $K(F+B)$ and memory cost $KS$ (which is reported in Table~\ref{table:computation-comparison}).

\paragraph{RTRL.}
We begin by deriving RTRL, which simply corresponds to forward-mode differentiation.
Let the state be $\bolds_t \in \mathbb{R}^S$ and the parameters be $\boldtheta \in \mathbb{R}^P$.
We have a dynamical system defined by:
\begin{equation} \label{eq:app-dynamical}
    \bolds_{t} = f(\bolds_{t-1}, \boldx_t, \boldtheta)
\end{equation}
and our objective is $L = \sum_{t=1}^T L_t$.
In order to optimize this objective, we need the gradient $\nabla_{\boldtheta} L = \sum_{t=1}^T \frac{d L_t}{d \boldtheta}$.
The loss at step $t$ is a function of $\bolds_t$, so we have:
\begin{equation}
    \frac{d L_t(\bolds_t)}{d \boldtheta} = \frac{\partial L_t}{\partial \bolds_t} \frac{d \bolds_t}{d \boldtheta}
\end{equation}
Using Eq.~\ref{eq:app-dynamical} and the chain rule, we have:
\begin{align}
    \frac{d \bolds_t}{d \boldtheta} &= \frac{d f(\bolds_{t-1}, \boldx_t, \boldtheta)}{d \boldtheta} \\
    &= \frac{\partial \bolds_t}{\partial \bolds_{t-1}} \frac{d \bolds_{t-1}}{d \boldtheta} + \frac{\partial \bolds_t}{\partial \boldx_t} \cancelto{0}{\frac{d \boldx_t}{d \boldtheta}} + \frac{\partial \bolds_t}{\partial \boldtheta} \cancelto{1}{\frac{d \boldtheta}{d \boldtheta}} \\
    &= \frac{\partial \bolds_t}{\partial \bolds_{t-1}} \frac{d \bolds_{t-1}}{d \boldtheta} + \frac{\partial \bolds_t}{\partial \boldtheta}
\end{align}

Thus, we have the recurrence relation:
\begin{equation}
    \underbrace{\frac{d \bolds_t}{d \boldtheta}}_{G_t} = \underbrace{\frac{\partial \bolds_t}{\partial \bolds_{t-1}}}_{H_t} \underbrace{\frac{d \bolds_{t-1}}{d \boldtheta}}_{G_{t-1}} + \underbrace{\frac{\partial \bolds_t}{\partial \boldtheta}}_{F_t}
\end{equation}
Here, $G_t$ is $S \times P$, $H_t$ is $S \times S$, and $F_t$ is $S \times P$.
RTRL maintains the Jacobian $G_t$, which requires memory $SP$; furthermore, instantiating the matrices $H_t$ and $F_t$ requires memory $S^2$ and $SP$, respectively, so the total memory cost of RTRL is $2SP + S^2$.
The matrix multiplication $H_t G_{t-1}$ has computational complexity $S^2 P$.
The cost of computing the Jacobian $F_t$ is approximately $\text{min} \{ S(F+B), P(F+B) \}$, depending on which of $S$ or $P$ is smaller-dimensional (and correspondingly whether we use forward-mode or reverse-mode automatic differentiation to compute the rows/columns of the Jacobian).
Similarly, the cost of computing the Jacobian $H_t$ is approximately $S(F+B)$ (using either forward or reverse mode autodiff).
Thus, the total computational cost of RTRL is: $S^2 P + S(F+B) + \text{min} \{ S(F+B) + P(F+B) \}$.

Note that, in general, it matters which of $\bolds_t$ or $\boldtheta$ is higher dimensional.
In the case of unrolled optimization, $S$ is usually larger than $P$, causing RTRL to be particularly memory-intensive due to the $S \times S$ Jacobian $H_t$.
The computation and memory costs we have derived here are expressed in a general form for state and parameter dimensions $S$ and $P$, respectively.
In the case of RNN training, most prior work (such as ~\cite{tallec2017unbiased,mujika2018approximating,benzing2019optimal}) assumes that the RNN parameters are of dimensionality $S^2$, where $S$ is the size of the hidden state.
\footnote{This is a simplification of the parameter count for RNNs, assuming that it is dominated by the hidden-to-hidden weight matrix.}

\paragraph{UORO.}
Unbiased Online Recurrent Optimization (UORO)~\citep{tallec2017unbiased} approximates RTRL by maintaining a rank-1 estimate of the Jacobian $G_t$ as:
\begin{equation}
    G_t \approx \tilde{\bolds}_t \tilde{\boldtheta}_t^\top
\end{equation}
where $\tilde{\bolds}_t$ and $\tilde{\boldtheta}_t$ are vectors of dimensions $S$ and $P$, respectively.
Ultimately, we are interested in the gradient $\frac{\partial L_t}{\partial \boldtheta}$.
Using the UORO approximation to $G_t$, we can write the gradient as follows:
\begin{align}
    \frac{\partial L_t}{\partial \boldtheta}
    &= \frac{\partial L_t}{\partial \bolds_t} \frac{d \bolds_t}{d \boldtheta} \\
    &= \frac{\partial L_t}{\partial \bolds_t} G_t \\
    &= \frac{\partial L_t}{\partial \bolds_t} (H_t G_{t-1} + F_t) \\
    &= \frac{\partial L_t}{\partial \bolds_t} (H_t (\tilde{\bolds}_t \tilde{\boldtheta}_t^\top) + F_t) \\
    &= \frac{\partial L_t}{\partial \bolds_t} (H_t (\tilde{\bolds}_t \tilde{\boldtheta}_t^\top)) + \frac{\partial L_t}{\partial \bolds_t} F_t \\
    &= \frac{\partial L_t}{\partial \bolds_t} (H_t (\tilde{\bolds}_t \tilde{\boldtheta}_t^\top)) + \frac{\partial L_t}{\partial \bolds_t} \frac{\partial \bolds_t}{\partial \boldtheta} \\
    &= \frac{\partial L_t}{\partial \bolds_t} (H_t (\tilde{\bolds}_t \tilde{\boldtheta}_t^\top)) + \frac{\partial L_t}{\partial \boldtheta} \\
    &= \underbrace{\left(\frac{\partial L_t}{\partial \bolds_t} H_t \tilde{\bolds}_t \right)}_{1 \times 1} \tilde{\boldtheta}_t^\top + \underbrace{\frac{\partial L_t}{\partial \boldtheta}}_{1 \times P}
\end{align}
Here, $\frac{\partial L_t}{\partial \bolds_t}$ is $1 \times S$, $H_t$ is $S \times S$, $\tilde{\bolds}_t$ is $S \times 1$, $\tilde{\boldtheta}_t$ is $1 \times P$, and $F_t$ is $S \times P$.

This leads to a total computation cost of $F + B + S^2 + P$.
We require one pass of backprop to compute the partial derivative, a vector-matrix product size $S$ by $S \times S$ ($S^2$), then element-wise operations on the full parameter space ($P$).
The memory cost of storing both $\tilde{\bolds}_t$ and $\tilde{\boldtheta}_t$ is $S + P$.

\paragraph{Reparameterization.}
The reparameterization gradient estimator is $\hat{g}^{\text{reparam}} = \frac{1}{N} \sum_{i=1}^N \nabla_\boldtheta L(\boldtheta + \sigma \boldepsilon^{(i)})$, where $\boldepsilon^{(i)} \sim \mathcal{N}(0, I)$.
With respect to computational complexity, this is equivalent to BPTT: its compute cost is $T(F+B)$ and its memory cost is $TS$.

\paragraph{ES.}
ES applied to an unroll of length $K$ requires performing $K$ forward passes---it does not require any backward passes, since ES is not gradient-based (e.g., it is a zeroth-order optimization algoritm).
Because ES does not require backprop, it does not need to store the intermediate states in memory, only the most recent state, yielding memory cost $S$ that is independent of the unroll length.
Using ES with $N$ particles yields total compute and memory costs $NKF$ and $NS$, respectively.

\paragraph{PES.}
As PES is an evolutionary strategies-based method, it also does not require backward passes; applied to unrolls of length $K$, PES has compute cost $KF$.
In addition to storing the current state of size $S$ as in ES, PES also maintains a perturbation accumulator for each particle; thus, the memory cost of a single PES chain is $S + P$.
Using PES with $N$ particles yields total compute and memory costs $NKF$ and $N (S + P)$, respectively.

\paragraph{PES+Analytic.}
Similarly to standard PES, we need to maintain a collection of $N$ states, each of size $S$, and $N$ perturbation accumulators, each of size $P$, yielding memory cost $N(S+P)$; unrolling each state for $K$ steps requires computational cost $NKF$.
To incorporate the analytic gradient, we need to maintain one additional particle that is unrolled using the mean $\boldtheta$ rather than a perturbed version $\boldtheta + \boldepsilon$; this adds memory cost $S$.
The main computational and memory overhead comes from the gradient computation through the partial unroll of length $K$: similarly to TBPTT, this requires storing $K$ intermediate states, yielding memory cost $KS$, and requires $K$ forward and $K$ backward operations, yielding computational cost $K (F+B)$.
Combined with the memory and computational cost of standard PES, we have total compute cost $NKF + K(F+B)$ and total memory cost $N(S+P) + (K+1)S$.

\section{Diagrammatic Representation of Algorithms}
\label{app:diagram-alg}

\begin{figure*}[h]
    \hspace{2cm}{\large \textbf{ES}}\hspace{9.5cm}{\large \textbf{PES}} \\ \\
    \includegraphics[width=0.24\linewidth]{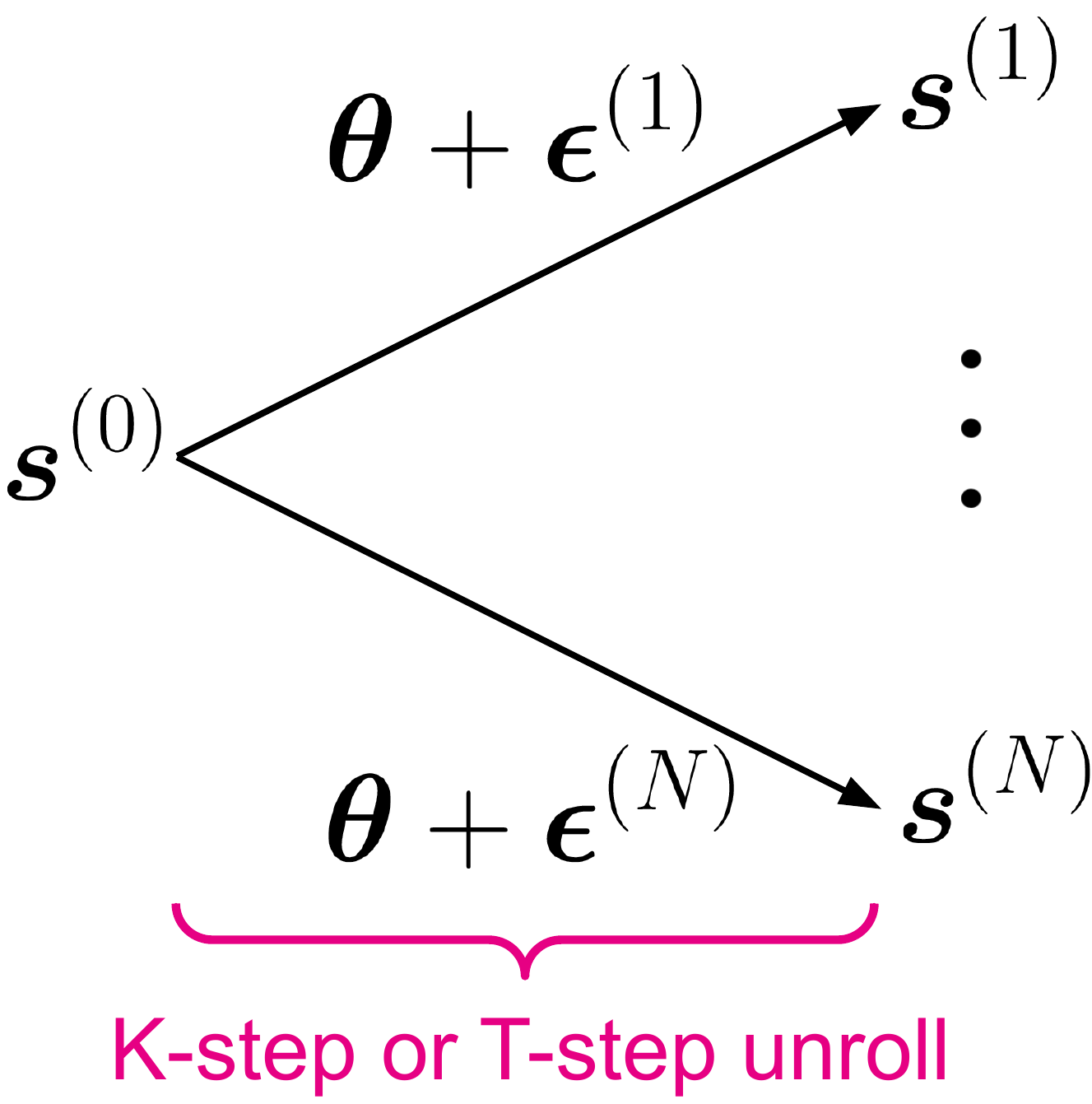}
    \hfill
    \includegraphics[width=0.65\linewidth]{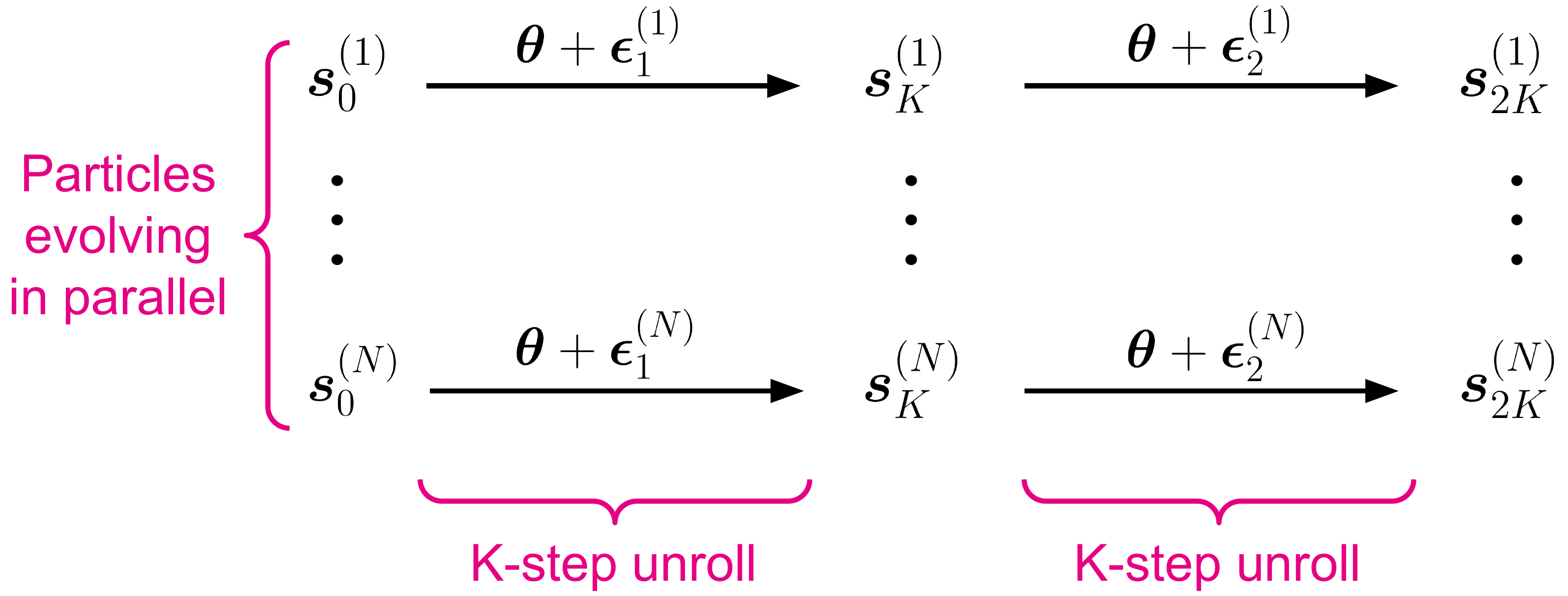}
    \caption{\textbf{Left:} Evolution strategies (ES). \textbf{Right:} Persistent evolution strategies (PES).}
    \label{fig:es-pes-diagram}
\end{figure*}

Figure~\ref{fig:es-pes-diagram} provides diagrammatic representations of ES and PES.
For each partial unroll, vanilla ES starts from a shared initial state $\bolds^{(0)}$ that is evolved in parallel using perturbed parameters $\boldtheta + \boldepsilon^{(i)}$.
After each truncated unroll, the mean parameters $\boldtheta$ are used to update the state, which then becomes the initial state for the next truncated unroll; no information is passed between truncated unrolls for vanilla ES.
In contrast, PES maintains a set of states $\bolds^{(i)}$ that are evolved in parallel, each according to a different perturbation of the parameters $\boldtheta$ in each truncated unroll.
Intuitively, these states maintain their history between truncated unrolls, since we accumulate the perturbations experienced by each state over the course of meta-optimization; when we reach the end of an inner problem, the states are reset to the same initialization, and the perturbation accumulators are reset to $\boldzero$.

\section{Ablation Studies}
\label{app:sensitivity}

\begin{figure*}[h]
    \centering
    \includegraphics[width=0.4\linewidth]{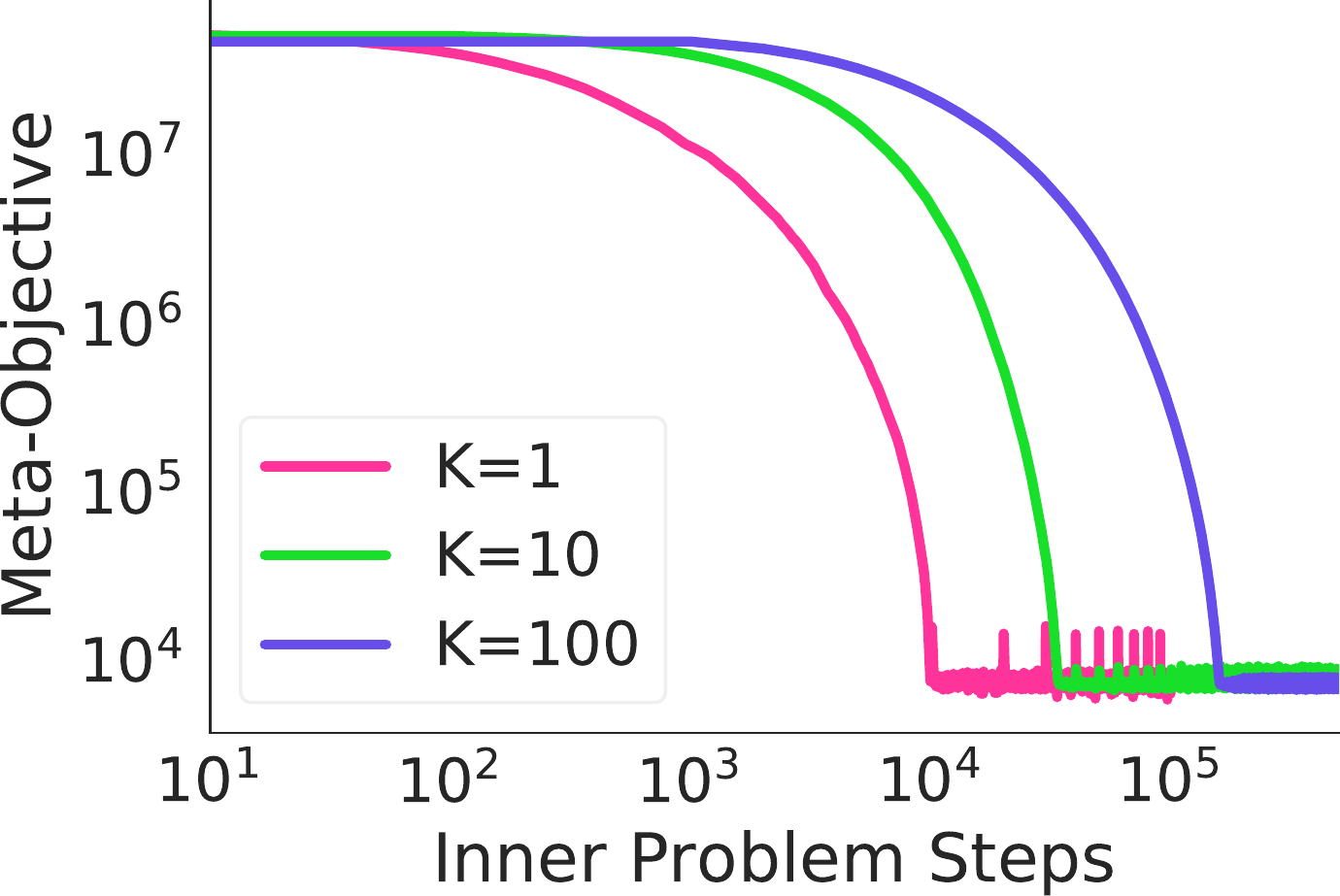}
    \hspace{1cm}
    \includegraphics[width=0.4\linewidth]{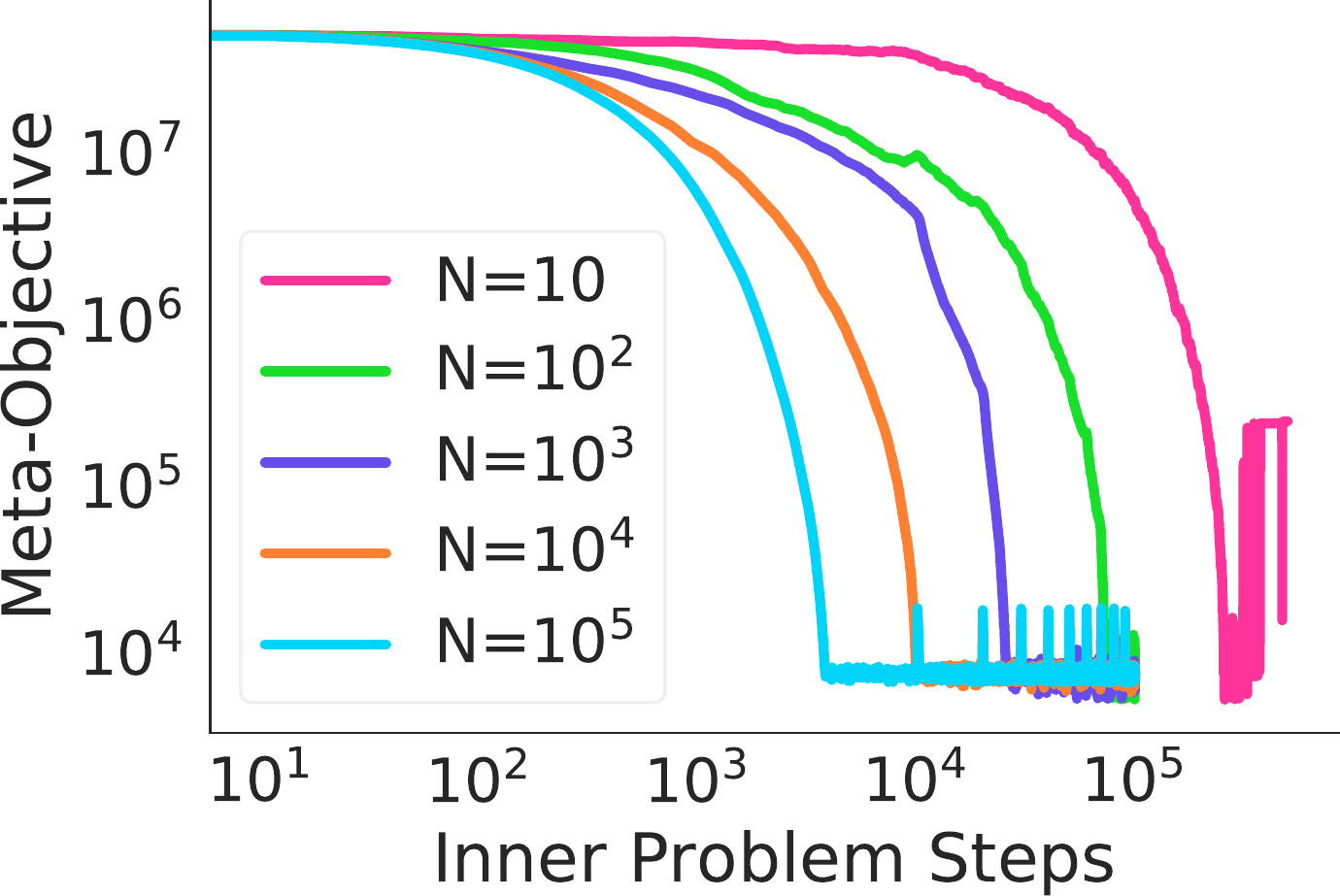} \\
    \hspace*{1cm}\textbf{(a)} \hspace{7.8cm} \textbf{(b)} \\
    \caption{Ablation over meta-parameters for PES applied to the toy 2D regression task with total number of inner steps $T=10,000$.
    Here we vary the truncation length $K$ and number of particles $N$; all other meta-parameters are fixed: we used Adam with learning rate 3e-2 for meta-optimization, and perturbation standard deviation 0.1.
    \textbf{(a)} Decreasing $K$ yields shorter truncations, which allow for more frequent meta-updates, improving performance compared to longer truncations. For these runs, $N=10^4$. \textbf{(b)} As in standard ES, increasing the particle count for PES reduces variance and can yield substantial improvements in terms of inner iterations performed, or wall-clock time. For these runs, $K=1$.} 
    \label{fig:toy2d-sensitivity}
\end{figure*}

In this section, we show an ablation study over the the number of particles $N$, and the truncation length $K$ (which controls the number of unrolls per inner-problem).
In Figure~\ref{fig:toy2d-sensitivity} we show the sensitivity of PES to these meta-parameters for a version of the 2D regression problem (from Section~\ref{sec:2d-regression}) with total inner problem length $T=10,000$.

\pagebreak

\section{Implementation}
\label{app:code}

Code Listing~\ref{lst:code} presents a simple JAX implementation of the toy 2D regression meta-learning problem from Section~\ref{sec:2d-regression}, in a self-contained, runnable example.
PES is easy to implement efficiently in JAX by making use of the construct \texttt{jax.vmap} (or \texttt{jax.pmap} in settings with multiple workers) to parallelize the unrolling computations over $N$ particles.

\begin{lstlisting}[language=Python, caption={Simplified PES implementation in JAX, for the 2D regression problem from Section~\ref{sec:2d-regression}.}, label={lst:code}]
from functools import partial
import jax
import jax.numpy as jnp

def loss(x):
    """Inner loss."""
    return jnp.sqrt(x[0]**2 + 5) - jnp.sqrt(5) + jnp.sin(x[1])**2 * \
           jnp.exp(-5*x[0]**2) + 0.25*jnp.abs(x[1] - 100)

# Gradient of inner loss
loss_grad = jax.grad(loss)

def update(state, i):
    """Performs a single inner problem update, e.g., a single unroll step.
    """
    (L, x, theta, t_curr, T, K) = state
    lr = jnp.exp(theta[0]) * (T - t_curr) / T + jnp.exp(theta[1]) * t_curr / T
    x = x - lr * loss_grad(x)
    L += loss(x) * (t_curr < T)
    t_curr += 1
    return (L, x, theta, t_curr, T, K), x

@partial(jax.jit, static_argnums=(3,4))
def unroll(x_init, theta, t0, T, K):
    """Unroll the inner problem for K steps.

    Args:
      x_init: the initial state for the unroll
      theta: a 2-dimensional array of outer parameters (log_init_lr, log_final_lr)
      t0: initial time step to unroll from
      T: maximum number of steps for the inner problem
      K: number of steps to unroll
    
    Returns:
      L: the loss resulting from the unroll
      x_curr: the updated state at the end of the unroll
    """
    L = 0.0
    initial_state = (L, x_init, theta, t0, T, K)
    state, outputs = jax.lax.scan(update, initial_state, None, length=K)
    (L, x_curr, theta, t_curr, T, K) = state
    return L, x_curr

@partial(jax.jit, static_argnums=(5,6,7,8))
def pes_grad(key, xs, pert_accum, theta, t0, T, K, sigma, N):
    """Compute PES gradient estimate.

    Args:
      key: JAX PRNG key
      xs: Nx2 array of particles/states to be updated
      pert_accum: Nx2 array of accumlated perturbations for each particle
      theta: a 2-dimensional array of outer parameters (log_init_lr, log_final_lr)
      t0: initial time step for the current unroll
      T: maximum number of steps for the inner problem
      K: truncation length for the unroll
      sigma: standard deviation of the Gaussian perturbations
      N: number of perturbations (as N//2 antithetic pairs)
    
    Returns:
      theta_grad: PES gradient estimate
      xs: Nx2 array of updates particles/states
      pert_accum: Nx2 array of updated perturbations for each particle
    """
    # Generate antithetic perturbations
    pos_perts = jax.random.normal(key, (N//2, theta.shape[0])) * sigma  # Antithetic positives
    neg_perts = -pos_perts  # Antithetic negatives
    perts = jnp.concatenate([pos_perts, neg_perts], axis=0)

    # Unroll the inner problem for K steps using the antithetic perturbations of theta
    L, xs = jax.vmap(unroll, in_axes=(0,0,None,None,None))(xs, theta + perts, t0, T, K)
    # Add the perturbations from this unroll to the perturbation accumulators
    pert_accum = pert_accum + perts
    # Compute the PES gradient estimate
    theta_grad = jnp.mean(pert_accum * L.reshape(-1, 1) / (sigma**2), axis=0)
    return theta_grad, xs, pert_accum

opt_params = { 'lr': 1e-2, 'b1': 0.99, 'b2': 0.999, 'eps': 1e-8,
                 'm': jnp.zeros(2),
                 'v': jnp.zeros(2) }

def outer_optimizer_step(params, grads, opt_params, t):
    lr = opt_params['lr']
    b1 = opt_params['b1']
    b2 = opt_params['b2']
    eps = opt_params['eps']
    opt_params['m'] = (1 - b1) * grads + b1 * opt_params['m']
    opt_params['v'] = (1 - b2) * (grads**2) + b2 * opt_params['v']
    mhat = opt_params['m'] / (1 - b1**(t+1))
    vhat = opt_params['v'] / (1 - b2**(t+1))
    updated_params = params - lr * mhat / (jnp.sqrt(vhat) + eps)
    return updated_params, opt_params

T = 100      # Total inner problem length
K = 10       # Truncation length for partial unrolls
N = 100      # Number of particles in total (N//2 antithetic pairs)
sigma = 0.1  # Standard deviation of perturbations

t = 0
theta = jnp.log(jnp.array([0.01, 0.01]))
x = jnp.array([1.0, 1.0])
xs = jnp.ones((N, 2)) * jnp.array([1.0, 1.0])
pert_accum = jnp.zeros((N, theta.shape[0]))

key = jax.random.PRNGKey(3)
for i in range(10000):
    key, skey = jax.random.split(key)
    if t >= T:
        # Reset the inner problem: the inner iteration, inner parameters, and perturbation accumulator
        t = 0
        xs = jnp.ones((N, 2)) * jnp.array([1.0, 1.0])
        x = jnp.array([1.0, 1.0])
        pert_accum = jnp.zeros((N, theta.shape[0]))

    theta_grad, xs, pert_accum = pes_grad(skey, xs, pert_accum, theta, t, T, K, sigma, N)
    theta, opt_params = outer_optimizer_step(theta, theta_grad, opt_params, i)
    t += K

    if i % 100 == 0:
        L, _ = unroll(jnp.array([1.0, 1.0]), theta, 0, T, T)  # Run a full unroll to get the cost
        print(i, jnp.exp(theta), theta_grad, L)
\end{lstlisting}

\end{document}